\documentclass[times,twocolumn,final]{elsarticle}

\usepackage{medima}
\usepackage{framed,multirow}

\usepackage{amssymb}
\usepackage{latexsym}

\usepackage{url}
\usepackage{xcolor}

\definecolor{newcolor}{rgb}{.8,.349,.1}

\journal{Medical Image Analysis}

\usepackage{graphicx}
\usepackage{amsfonts}
\usepackage{amsmath}
\usepackage[separate-uncertainty=true]{siunitx}
\usepackage{newunicodechar}
\usepackage[english]{babel}
\usepackage{multirow}
\usepackage{svg}
\usepackage{subcaption}
\usepackage[export]{adjustbox}
\usepackage{acro}
\usepackage{placeins}
\usepackage{tablefootnote}
\usepackage[hyperfootnotes=true]{hyperref}	
\usepackage[capitalise,noabbrev]{cleveref}
\usepackage[normalem]{ulem}

\newcommand\todo[1]{}
\newcommand\rev[1]{#1}
\newcommand\revTwo[1]{#1}
\newcommand\etal{\textit{et~al.}}
\newunicodechar{‑}{\babelhyphen{nobreak}}

\DeclareMathOperator*{\argmin}{argmin}

\DeclareAcronym{dof}{
  short=DoF,
  long=degree of freedom,
}
\newcommand\dof{\acs{dof} }

\DeclareAcronym{hl}{
  short=HL 2,
  long=HoloLens 2,
}
\DeclareAcronym{ak}{
  short=Azure Kinect,
  long=Azure Kinect,
}
\DeclareAcronym{pv}{
  short=PV,
  long=photo-video,
}
\DeclareAcronym{pnp}{
  short=PnP,
  long=perspective-n-point,
}
\DeclareAcronym{hmd}{
  short=HMD,
  long=head-mounted device,
}
\DeclareAcronym{psp}{
  short=PSP,
  long=pedicle screw placement,
}
\DeclareAcronym{drr}{
  short=DRR,
  long=digitally reconstructed radiograph,
}
\DeclareAcronym{fov}{
  short=FOV,
  long=field of view,
}
\DeclareAcronym{ir}{
  short=IR,
  long=infrared,
}
\DeclareAcronym{ar}{
  short=AR,
  long=augmented reality,
}
\DeclareAcronym{fps}{
  short=fps,
  long=frames per second,
}
\DeclareAcronym{bop}{
  short=BOP,
  long=benchmark for 6D object pose estimation,
}

\begin{document}

\verso{Jonas Hein \etal}

\begin{frontmatter}

\title{Next-generation Surgical Navigation: Marker-less Multi-view 6\dof Pose Estimation of Surgical Instruments}%

\author[1,2]{Jonas \snm{Hein}}
\ead{jonas.hein@inf.ethz.ch}
\author[1]{Nicola \snm{Cavalcanti}}
\author[3]{Daniel \snm{Suter}}
\author[3]{Lukas \snm{Zingg}}
\author[1,4]{Fabio \snm{Carrillo}}
\author[4]{Lilian \snm{Calvet}}
\author[3]{Mazda \snm{Farshad}}
\author[5]{Nassir \snm{Navab}}
\author[2]{Marc \snm{Pollefeys}}
\author[1,4]{Philipp \snm{Fürnstahl}}

\address[1]{Research in Orthopedic Computer Science, Balgrist University Hospital, University of Zurich, Zurich, Switzerland}
\address[2]{Computer Vision and Geometry Group, ETH Zurich, Zurich, Switzerland}
\address[3]{Balgrist University Hospital, University of Zurich, Zurich, Switzerland}
\address[4]{OR-X Translational Center for Surgery, Balgrist University Hospital, University of Zurich, Zurich, Switzerland}
\address[5]{Computer Aided Medical Procedures, Technical University Munich, Munich, Germany}


\begin{abstract}
State-of-the-art research of traditional computer vision is increasingly leveraged in the surgical domain.
A particular focus in computer-assisted surgery is to replace marker-based tracking systems for instrument localization with pure image-based 6\dof pose estimation using deep-learning methods. 
However, state-of-the-art single-view pose estimation methods do not yet meet the accuracy required for surgical navigation.
In this context, we investigate the benefits of multi-view setups for highly accurate and occlusion-robust 6\dof pose estimation of surgical instruments and derive recommendations for an ideal camera system that addresses the challenges in the operating room.

\rev{Our contributions are threefold. 
First, we present a multi-view RGB-D video dataset of ex-vivo spine surgeries, captured with static and head-mounted cameras and including rich annotations for surgeon, instruments, and patient anatomy. 
Second, we perform an extensive evaluation of three state-of-the-art single-view and multi-view pose estimation methods, analyzing the impact of camera quantities and positioning, limited real-world data, and static, hybrid, or fully mobile camera setups on the pose accuracy, occlusion robustness, and generalizability.
Third, we design a multi-camera system for marker-less surgical instrument tracking, achieving an average position error of \SI{1.01}{\milli\meter} and orientation error of \SI{0.89}{\degree} for a surgical drill, and \SI{2.79}{\milli\meter} and \SI{3.33}{\degree} for a screwdriver under optimal conditions.}
Our results demonstrate that marker-less tracking of surgical instruments is becoming a feasible alternative to existing marker-based systems.
\end{abstract}


\end{frontmatter}


\section{Introduction}

\begin{figure*}[t]
\hspace*{0pt}
\hfill
\includegraphics[height=5cm, width=0.245\linewidth, keepaspectratio]{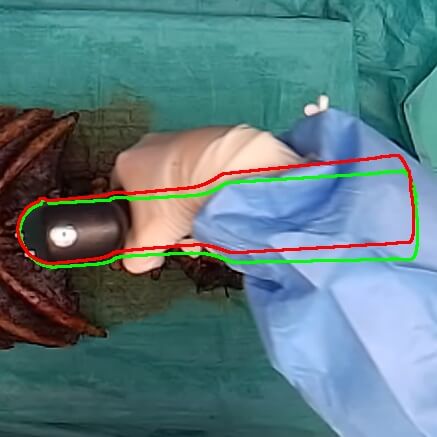}
\hfill
\includegraphics[height=5cm, width=0.245\linewidth, keepaspectratio]{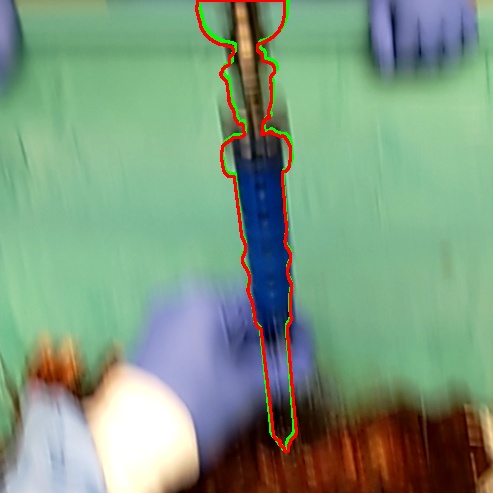}
\hfill 
\includegraphics[height=5cm, width=0.245\linewidth, keepaspectratio]{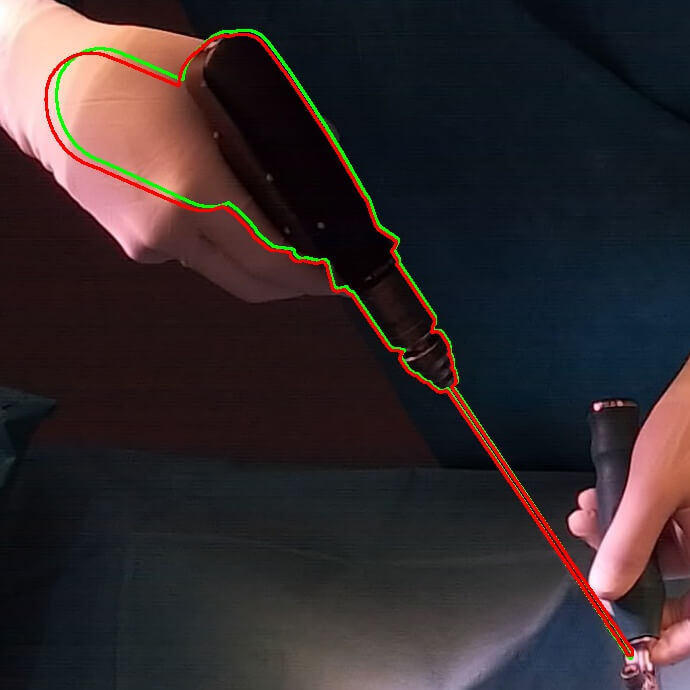}
\hfill
\includegraphics[height=5cm, width=0.245\linewidth, keepaspectratio]{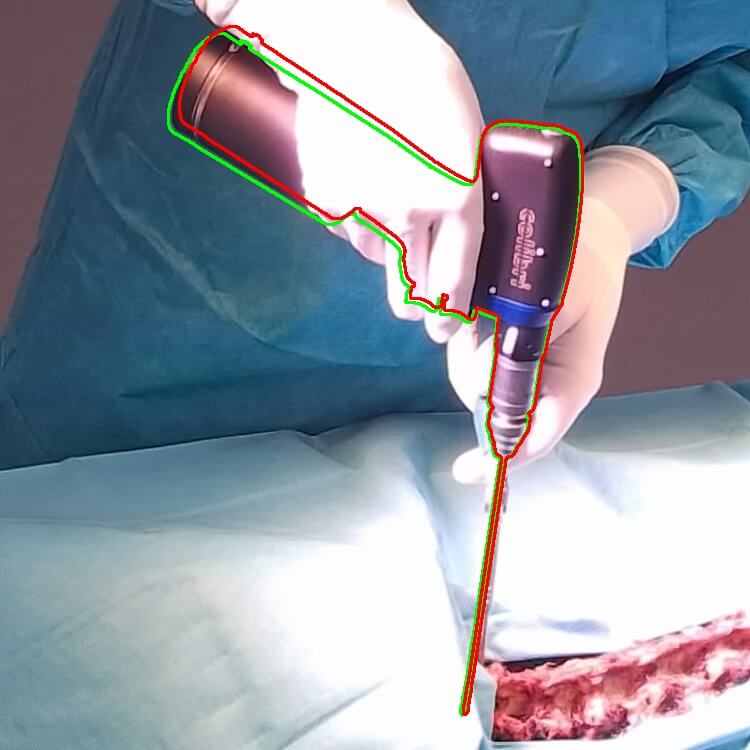}
\hfill
\hspace*{0pt}
\caption{Excerpt of our test set in a surgical wet lab (left) and an operating room (right).
6\dof pose estimation of surgical instruments is a complex task due to challenging lighting conditions, frequent occlusions, as well as motion blur in ego-centric perspectives.
The superimposed outlines indicate the ground truth pose (green) and the pose estimate of our multi-view baseline (red).
}
\label{fig:teaser}
\end{figure*}

Computer-assisted interventions have benefited significantly from advances in computer vision \citep{mascagni2022computer} to increase autonomy, accuracy, and usability for tasks such as navigation, surgical robotics, surgical phase recognition, or automated performance assessment \citep{farshad2021first,doughty2022hmd,haidegger2022robot,garrow2021machine,lam2022machine}.
While most methods are currently being studied in isolation for specific use cases, the intention is to integrate them holistically in a new generation of operating rooms optimized for the utilization of computer vision \citep{feussner2017surgery, maier2022surgical, ozsoy2023holistic}.
Hereby, the data streams are utilized to support the surgical staff in all relevant aspects of a surgery ranging from clinical process optimization to precision surgery \citep{Ozsoy2022_4D_OR}.

\begin{figure*}[t]
\centering
\hspace*{0pt}
\includegraphics[height=33.5mm, keepaspectratio]{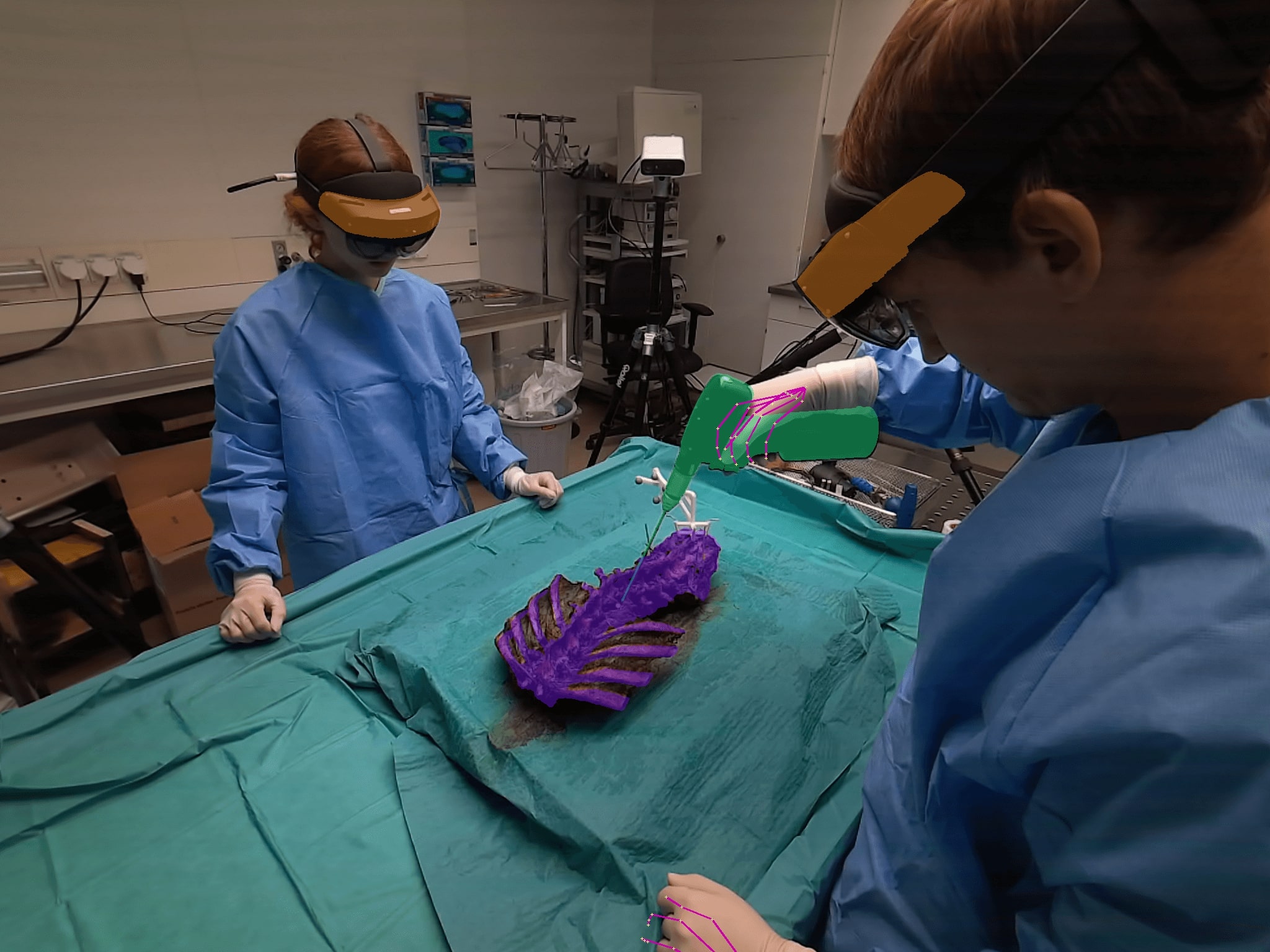}
\hfill
\includegraphics[height=33.5mm, keepaspectratio]{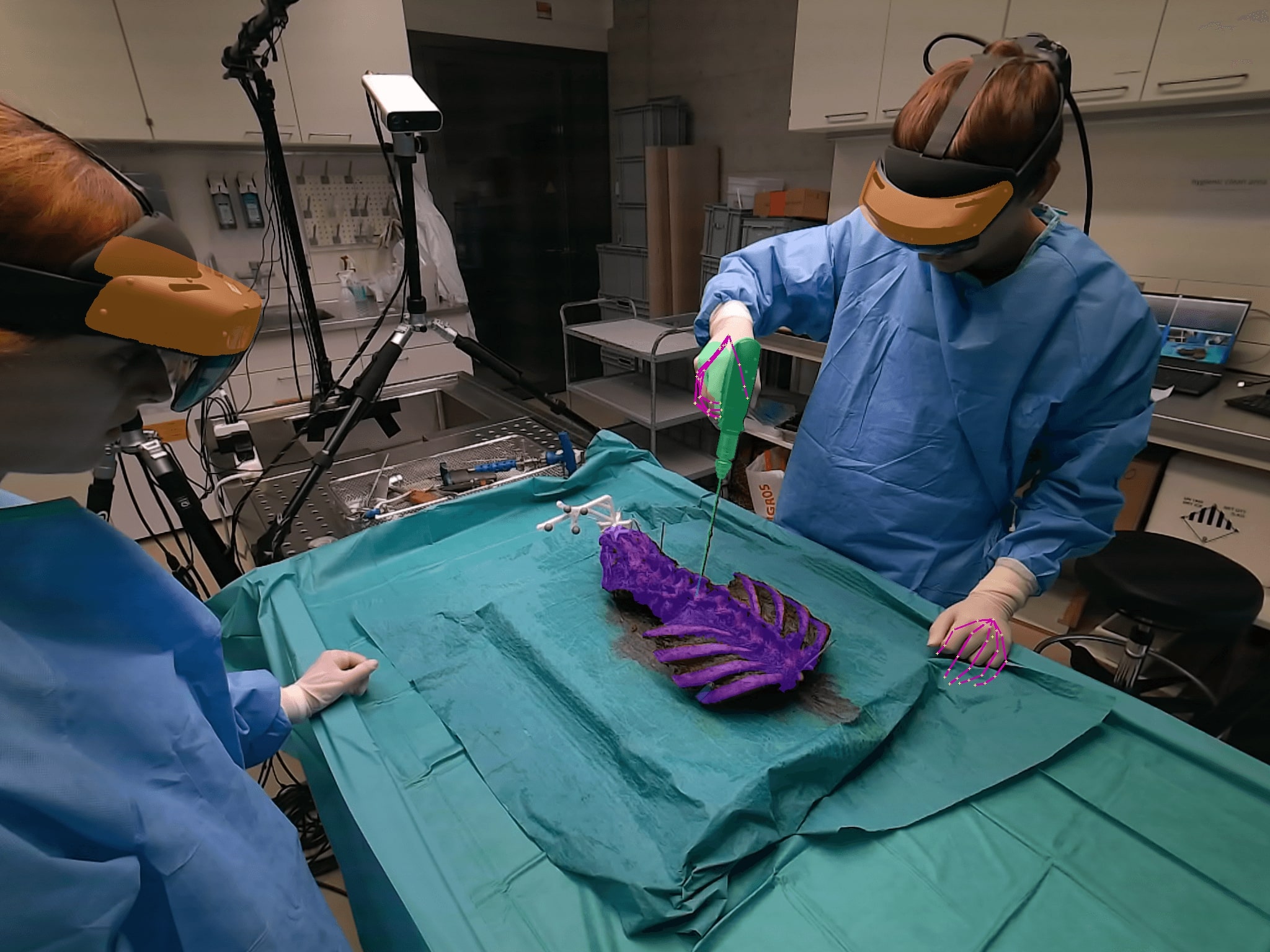}
\hfill
\includegraphics[height=33.5mm, keepaspectratio]{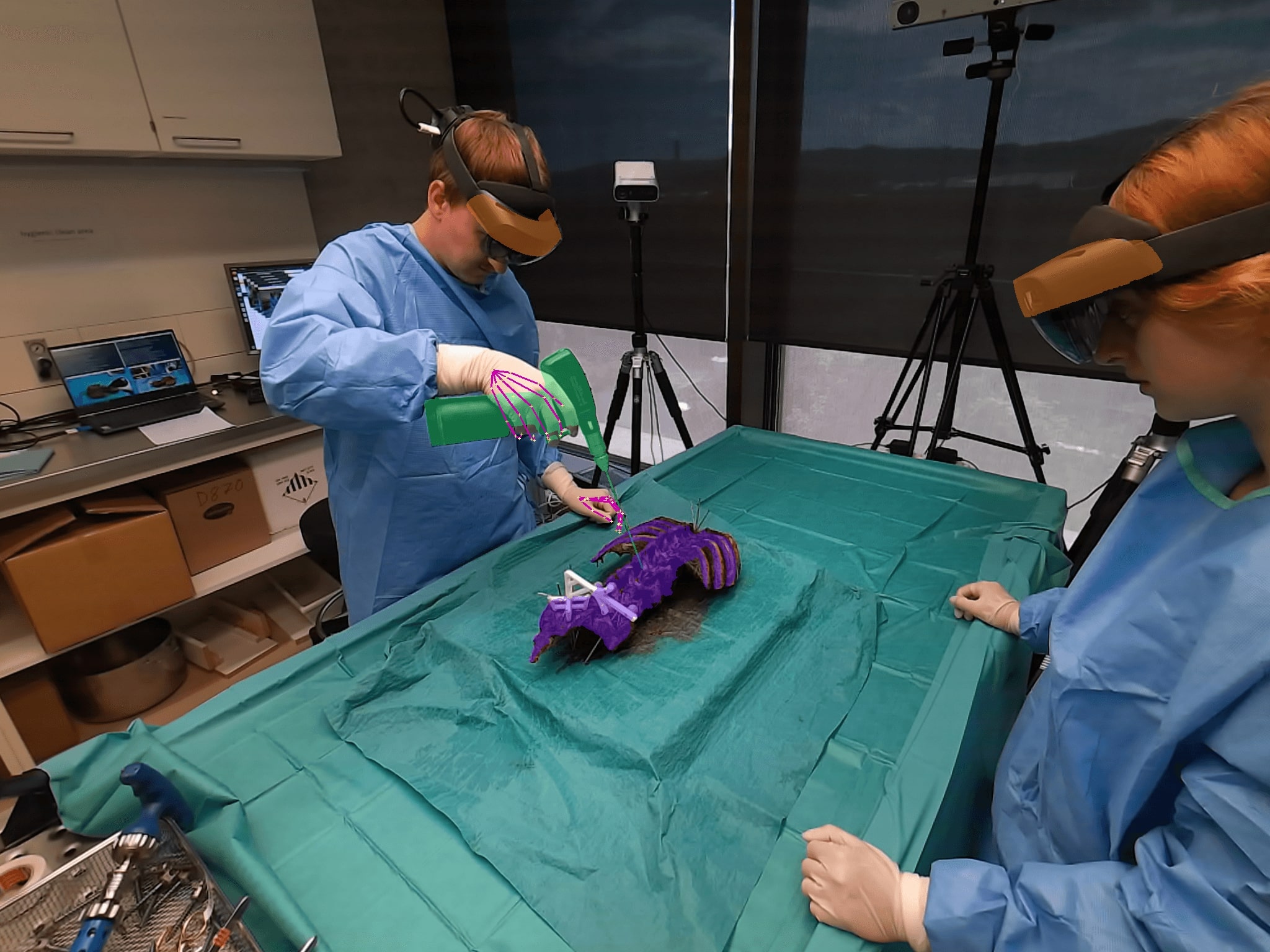}
\hfill
\includegraphics[height=33.5mm, keepaspectratio]{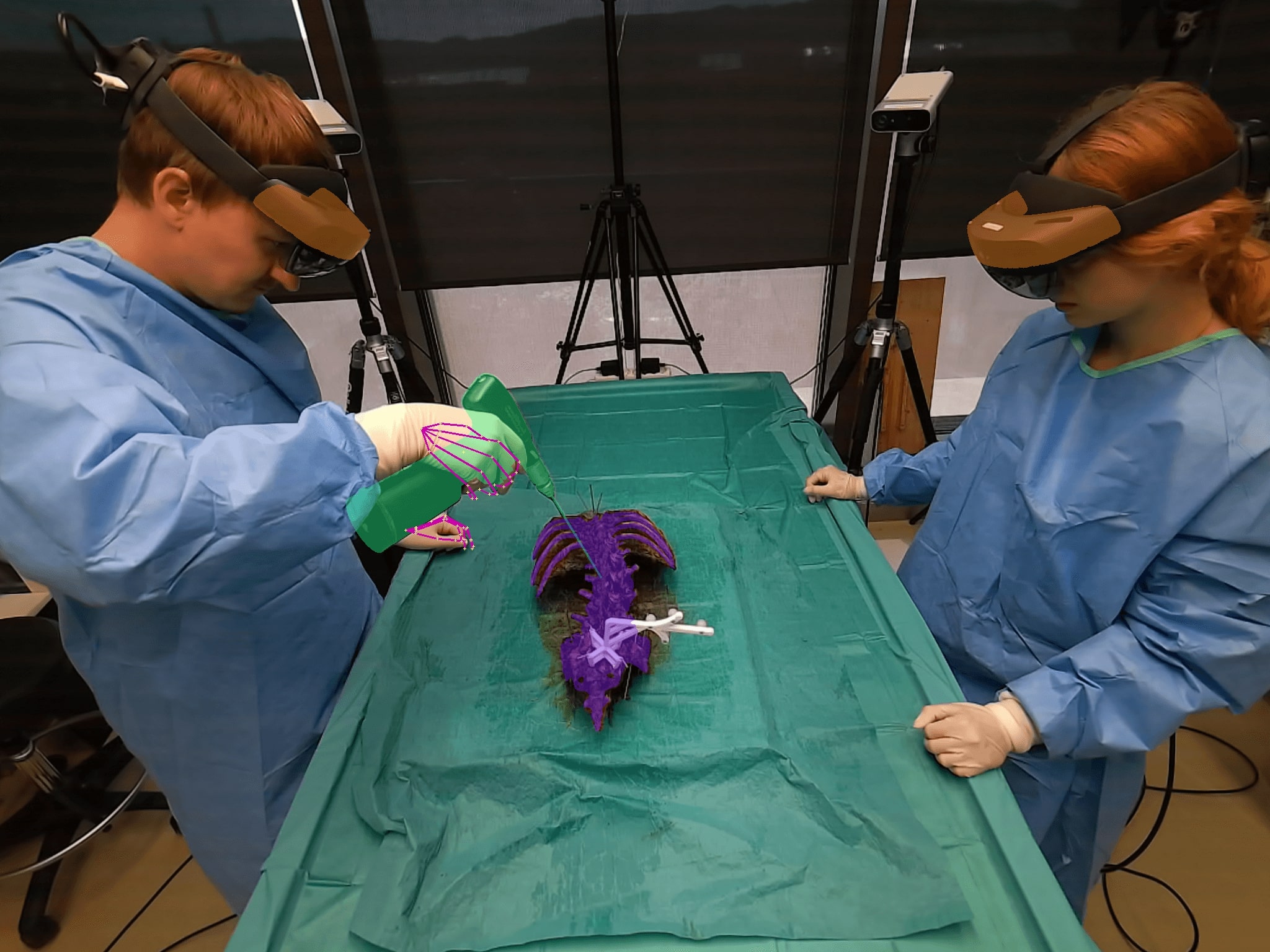}
\hspace*{0pt}
\\ \vspace*{.5mm}
\hspace*{0pt}
\includegraphics[height=36.5mm, keepaspectratio]{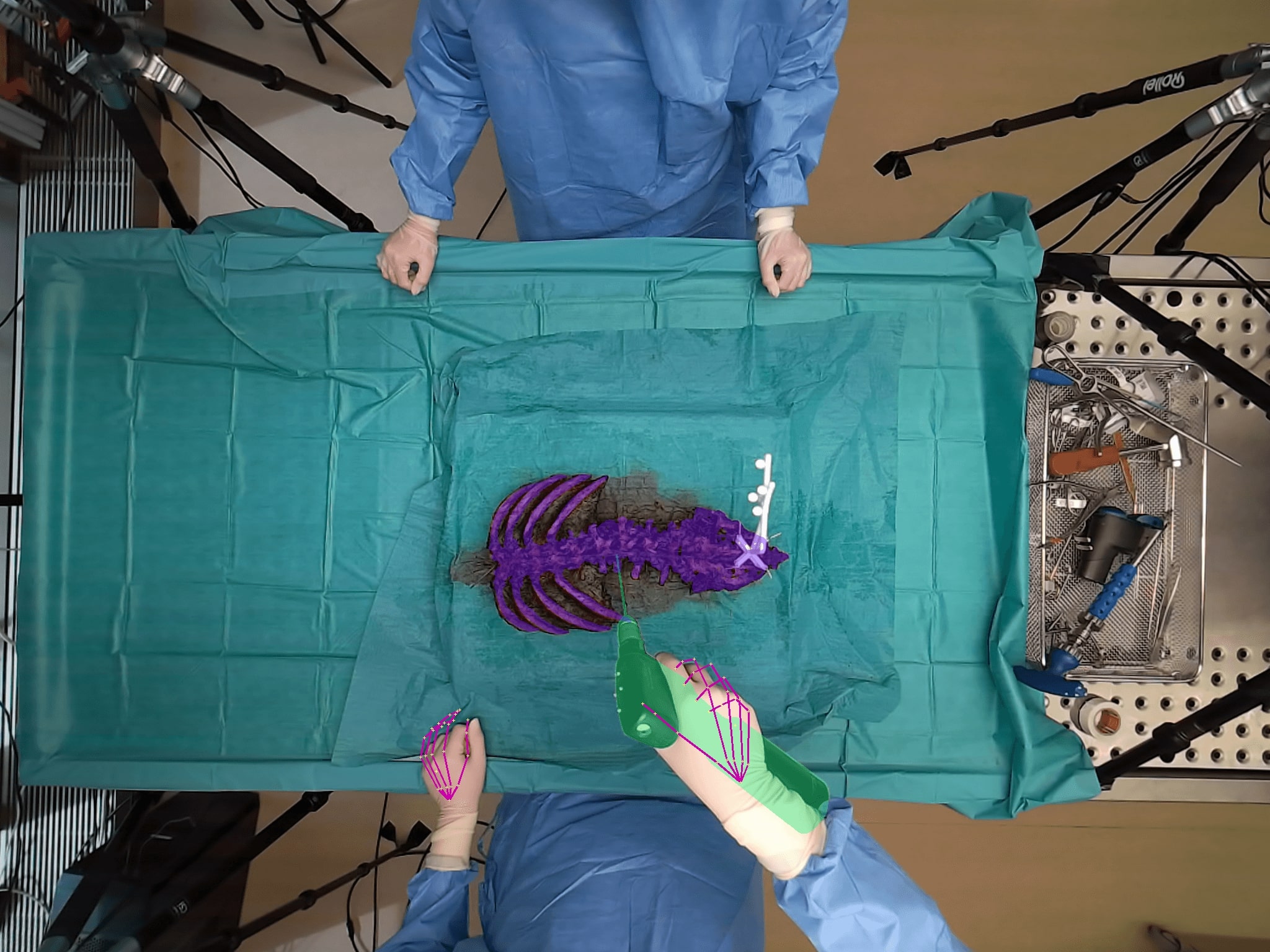}
\hfill
\includegraphics[height=36.5mm, keepaspectratio]{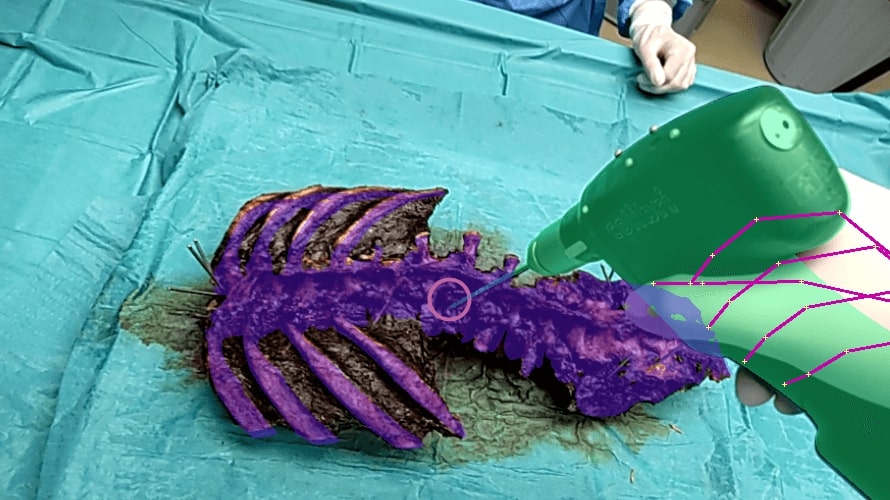}
\hfill
\includegraphics[height=36.5mm, keepaspectratio]{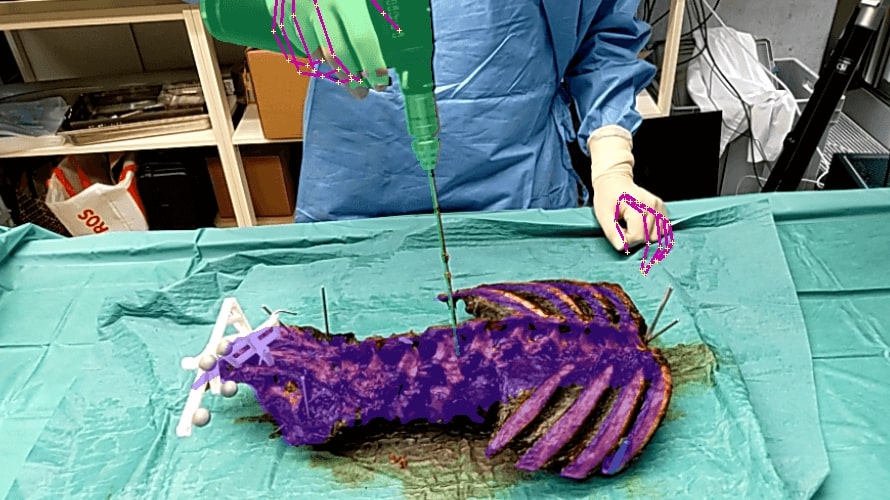}
\hspace*{0pt}
\caption{Overview of the camera views in the surgical wet lab setup, with ground truth pose overlays of the drill, anatomy, hand tracking, and eye gaze. 
The shown cameras are (top-to-bottom, left-to-right) left (L), opposite left (OL), opposite right (OR), right (R), ceiling (C), surgeon (S), and assistant (A).
}
\label{fig:mvpsp-example-frames}
\end{figure*}

\begin{figure}[t]
\centering
\hspace*{0pt}
\hfill
\includegraphics[height=4cm, width=0.49\linewidth, keepaspectratio]{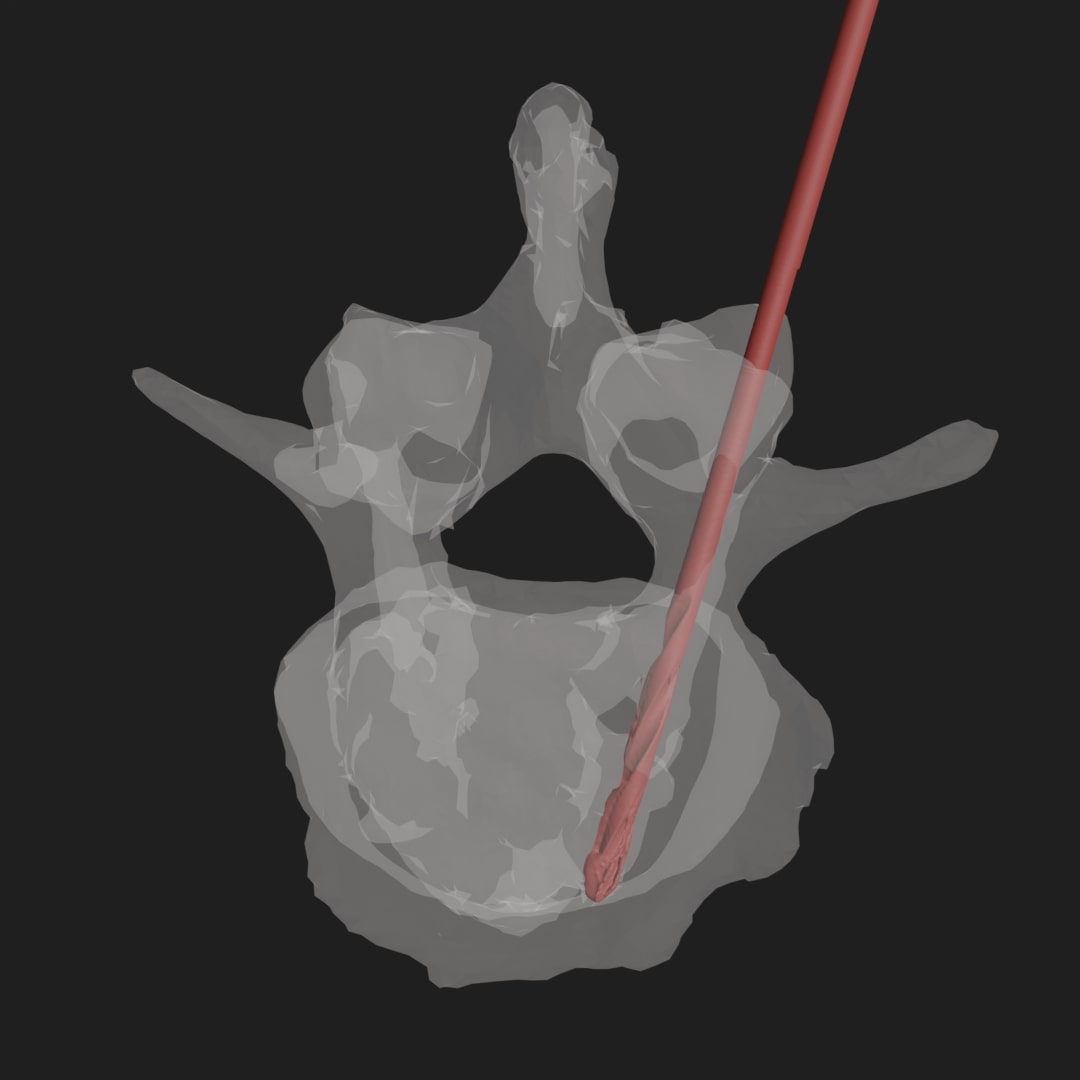}
\hfill
\adjincludegraphics[height=4cm, width=0.49\linewidth, trim={{.0\width} {0.05\height} {.05\width} 0}, clip, keepaspectratio]{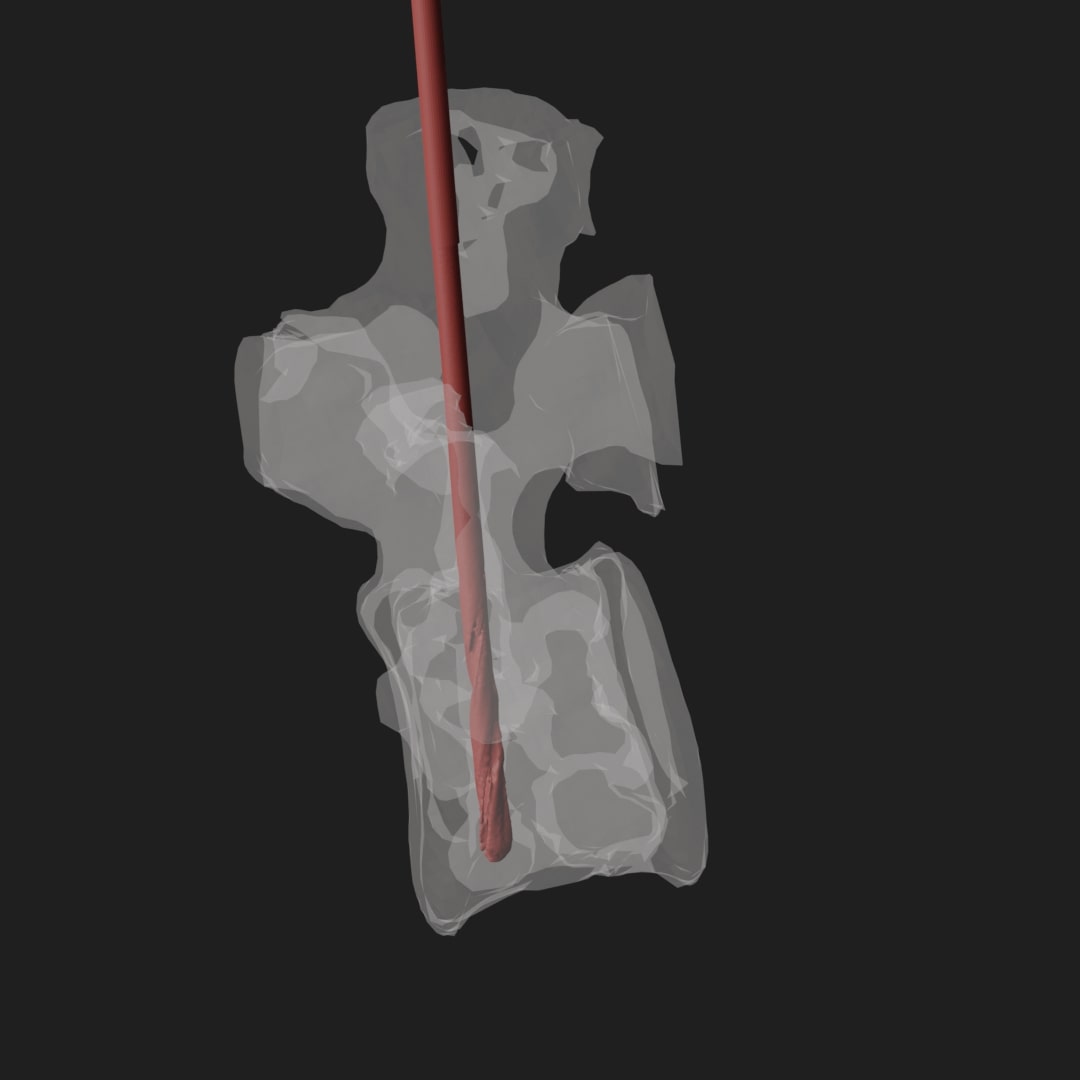}
\hfill
\hspace*{0pt}
\caption{Superior and lateral view of an exemplary drill trajectory inside the L4 vertebra.}
\label{fig:drill_trajectory}
\end{figure}

In precision surgery, a particular significance is attributed to surgical navigation, which improves the safety and efficiency of interactions between the surgeon, instruments, and the patient \citep{virk2019navigation}.
Marker-based navigation systems have been available for \rev{more than} two decades and have been shown to increase accuracy and reduce revision rates \citep{girardi1999placement, luther2015comparison, perdomo2019accuracy}.
\rev{However, their limited applicability and inherent technical restrictions such as line-of-sight issues, extensive calibration requirements, and the impracticality of large tracking markers complicate their integration into existing workflows and limit acceptance and dissemination \citep{hartl2013worldwide, joskowicz2016computer}.
In contrast, marker-less approaches have significant potential to seamlessly integrate into the surgical workflow and considerably reduce logistics and calibration overhead.}

As a fundamental computer vision problem, marker-less object pose estimation remains an active research focus with a continuously improving state of the art.
\rev{Outside of the medical domain, most proposed methods operate on single RGB frames due to their broad applicability \citep{hinterstoisser2012model,xiang2018posecnn,Wang_2021_GDRN}, however, their accuracy is \rev{constrained} by depth ambiguities.
Other works \rev{address this limitation by incorporating} RGB-D sensors \citep{labbe2020cosypose,haugaard2022surfemb} or multiple cameras \citep{labbe2020cosypose,shugurov2021dpodv2,haugaard2023multi}.
In particular, multi-view methods show potential for high pose accuracy and occlusion robustness due to the redundancy of multiple viewpoints and the robust triangulation in wide baseline camera setups.}
Such state-of-the-art object pose estimation methods have been successfully applied in various fields like robotic grasping \citep{wang2019densefusion}, augmented reality \citep{liu2022gen6d}, or outer space \citep{hu2021wide}.
\rev{
However, a systematic evaluation of the feasibility and requirements of these methods in surgery is still lacking, primarily due to the absence of publicly available datasets for training and evaluation. This lack of suitable benchmarks has been recognized as a key challenge in translating state-of-the-art methods to the surgical domain \citep{bouget2017vision,mascagni2022computer}.
}

\vspace{5mm}

\rev{Several works have investigated marker-less approaches for pose estimation and tracking of surgical instruments, however, the proposed approaches are often based on strong assumptions about the instrument shape \citep{hasan2021detection,chiu2022markerless} or image appearance \citep{allan2015image}.
These assumptions restrict their generalization and applicability to a broader range of instruments and use cases.
Other works propose registration-based methods with depth sensors \citep{lee2017multi}, or exploit correlations between the hand and hand-held instrument for pose estimation \citep{hein2021towards,doughty2022hmd}.
Still, these monocular methods fail to achieve sufficient accuracy due to their limited robustness to occlusions and noisy depth measurements. 
Despite the evident potential of multi-view methods, no such approach has yet been proposed for surgical instrument pose estimation or tracking.
}

\rev{Dedicated multi-view datasets can support the development of multi-view approaches, however, such datasets remain scarce in both quality and quantity.
In the surgical domain, most existing datasets provide 2D annotations such as bounding boxes, tool tip positions, or segmentation masks \citep{sarikaya2017detection, allan20202018}, but lack 6\dof pose annotations due to the added complexity during data acquisition.
To address this challenge, some datasets automatically annotate 6\dof instrument poses based on the surgeon's hand pose and grasp information \citep{hein2021towards,wang2023pov}.
However, the accuracy of the estimated instrument pose is often insufficient for clinical applications due to accumulating errors in the hand pose and grasp estimation.}
\rev{A notable exception is datasets collected on the Da Vinci robotic platform \citep{allan2015image,speidel_endoscopic_2023}.
While these datasets include accurate 6\dof pose annotations, they are inherently limited to minimally invasive surgery and the specific robotic instruments used with the Da Vinci system. 
Complementary to real-world data collection, some works generate synthetic images of hand-held surgical instruments \citep{hein2021towards, birlo_hup-3d_2024} to support the training process. 
Nevertheless, real-world data remains essential for evaluating a method’s accuracy under realistic conditions.
To the best of our knowledge, no publicly available benchmark exists that enables a systematic evaluation of state-of-the-art single-view and multi-view approaches, based on RGB or RGB-D data, for surgical instrument tracking.}

\rev{In this work, we address the existing limitations in surgical instrument tracking through three key contributions. 
First, we introduce the first public and comprehensive multi-modal and multi-camera spine surgery dataset to overcome the lack of benchmarks. 
This dataset includes 23 recordings of surgical procedures on human ex-vivo anatomy performed by five operators using two distinct instruments. 
The data capture setup comprises RGB-D video streams from seven cameras, including static and head-mounted configurations, collected in both a surgical wet lab and a mock operating room. 
A marker-based tracking system with sub-millimeter accuracy provides precise pose annotations for the surgical instruments, patient anatomy, and \acp{hmd}. 
This dataset establishes a robust benchmark for advancing research on pose estimation and tracking of surgical instruments.
Moreover, the rich annotations and modalities broaden the dataset's applicability to several related tasks such as hand or joint hand-object pose estimation and tracking \citep{hein2021towards, wang2023pov}, reconstruction \citep{leng2023dynamic}, or novel view synthesis \citep{mildenhall2021nerf, truong2023sparf}.
In the clinical context, our dataset can serve as the basis for surgical behavioral and interaction models based on the provided instrument-, hand- and anatomy poses and eye gaze information displayed in \cref{fig:mvpsp-example-frames,fig:drill_trajectory}.
Moreover, the instrument and anatomy information can be used to render \acp{drr} of realistic instrument trajectories, enabling the training of pose estimation and phase detection models in the x-ray domain \citep{kugler2020i3posnet, killeen2023pelphix}.
}

\rev{Second, we conduct an extensive evaluation of pose estimation methods to assess the feasibility of marker-less surgical instrument tracking. 
This evaluation benchmarks three state-of-the-art single-view and multi-view methods, examining the influence of camera quantity and placement, ego-centric perspectives from HMDs, and varying camera configurations, including static, hybrid, and fully mobile setups. 
Furthermore, we analyze how different training strategies and limited real-world training data impact pose accuracy, occlusion robustness, and generalizability.}

\rev{Third, we propose a 6\dof instrument tracking system and training strategy based on the results of our evaluation.
The system integrates multiple off-the-shelf cameras with state-of-the-art pose estimation methods to address the challenges in the operating room.
We demonstrate that marker-less tracking is becoming a viable alternative to existing marker-based navigation systems.
The dataset \revTwo{is} publicly available on our project page \hyperlink{https://jonashein.github.io/mvpsp/}{https://jonashein.github.io/mvpsp/}.}

\section{Methodology}\label{sec:methodology}

\rev{Our 6\dof marker-less tracking approach is specifically designed for open surgery procedures, with spinal surgery serving as a representative application.}
Our objective is to track the 3D position and orientation of two commonly used surgical instruments: a surgical drill and a screwdriver.

\rev{We choose spinal surgery as a representative use case due to its high prevalence and stringent accuracy requirements.
Ex-vivo validation studies for surgical navigation systems generally target a screw placement accuracy of \SI{2}{\milli\meter} and \SI{2}{\degree}. 
In clinical practice, the primary criterion is the complete embedding of the screw within the bone \citep{gertzbein1990accuracy}. 
Breach severity is typically categorized into classes ranging from \SIrange{2}{6}{\milli\meter} based on the screw edge's distance from the pedicle cortex \citep{nevzati2014accuracy} or relative to the screw diameter \citep{mahesh2020acceptable}. 
A theoretical derivation of the tolerable position and orientation errors can be found in the work of \cite{rampersaud2001accuracy}. 
}

The subsequent sections of this chapter are organized as follows: In \cref{sec:camera_setup,sec:calibration}, we introduce the multi-camera acquisition setup as well as the joint calibration and synchronization method developed for our study.
Next, we present two datasets captured in a surgical wet lab and a real operating room, which are suitable for the evaluation of pose estimation and tracking methods, as well as for the training of learning-based models. 
These datasets are presented in \cref{sec:datasets}.
Last, we describe the integration of the state-of-the-art pose estimation baselines into our tracking system in \cref{sec:baselines}.

\subsection{Camera Setup} \label{sec:camera_setup}

\begin{figure}[t]
\centering
\includegraphics[width=.7\linewidth, keepaspectratio]{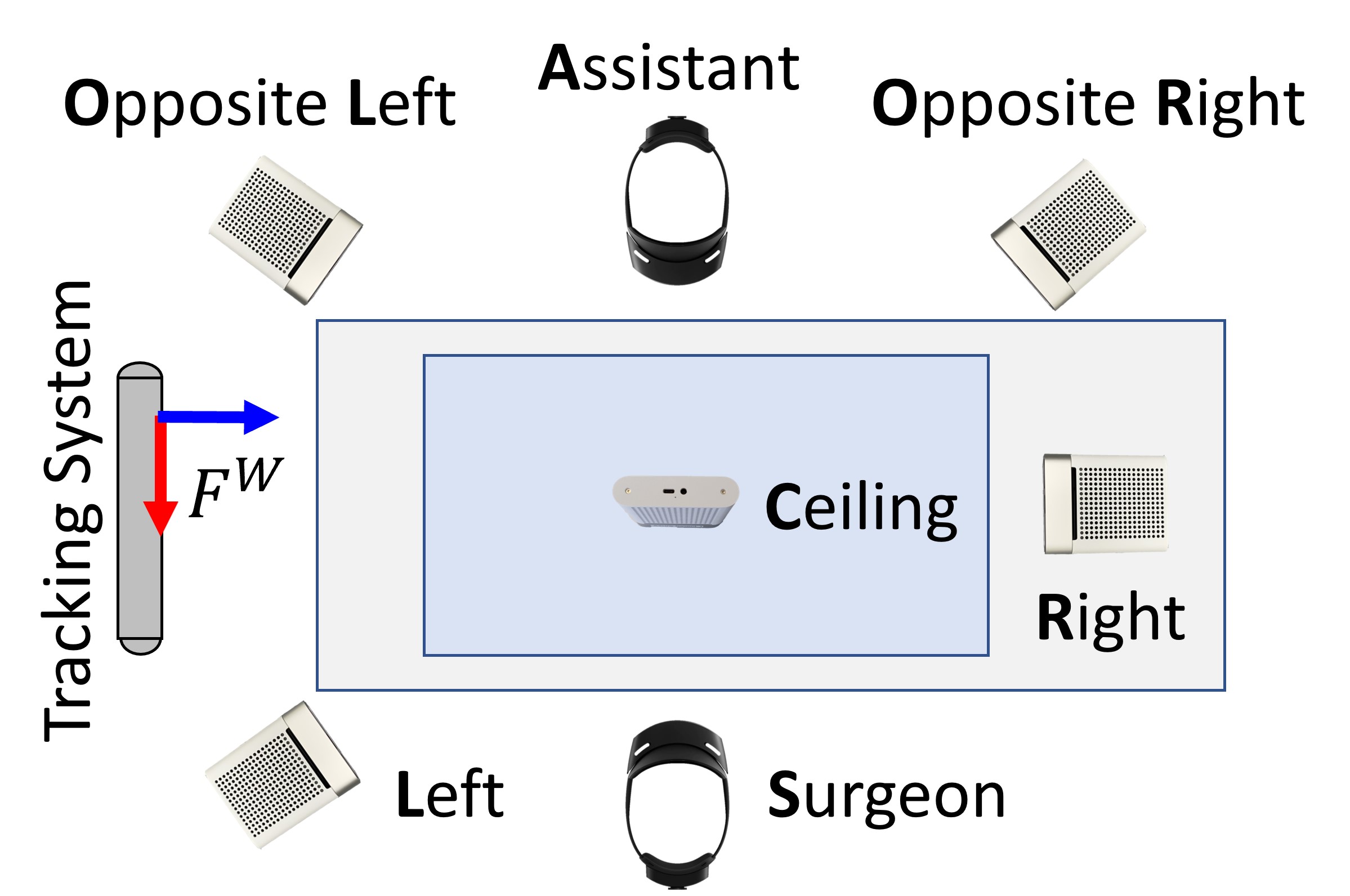}
\\ \vspace*{.5mm}
\adjincludegraphics[height=4cm, width=.9\linewidth, trim={{.04\width} 0 {.08\width} 0}, clip, keepaspectratio]{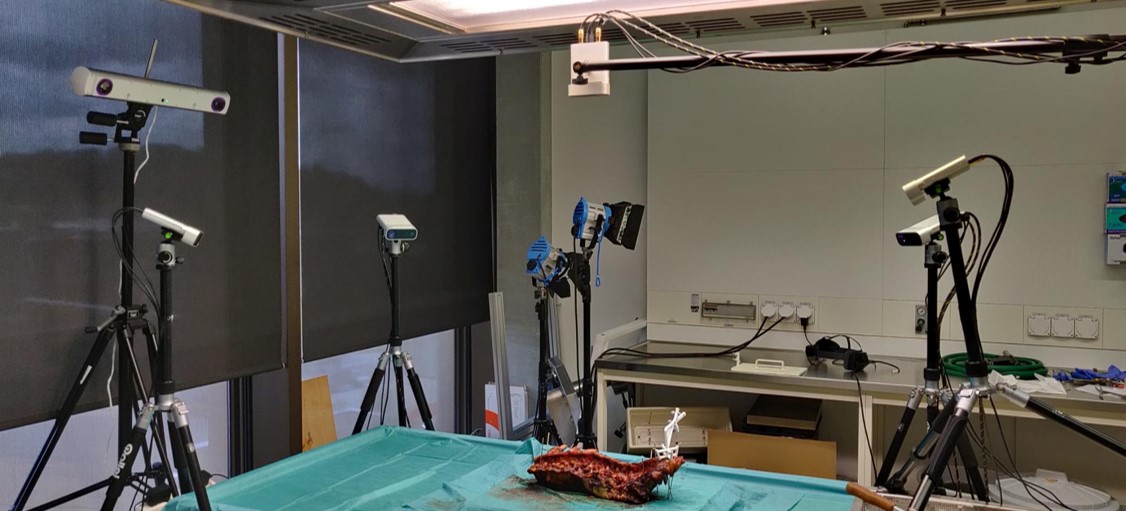}
\caption{Overview of the multi-camera acquisition setup in the surgical wet lab. Multiple static RGB-D cameras are placed around the operating field and on the ceiling. The surgeon and assistant are equipped with \acp{hmd}. All cameras are calibrated beforehand. To obtain accurate ground truth data, all instruments and \acp{hmd} are tracked with a marker-based tracking system.
}
\label{fig:camera_setup}
\end{figure}

Our envisioned camera setup for a next-generation operating room (as shown in \cref{fig:mvpsp-example-frames,fig:camera_setup}) consists of multiple static and mobile cameras, the latter in the form of \acl{ar} \acp{hmd} that are worn by the surgeons.
We place four \acl{ak} cameras (Microsoft Corporation, Redmond, WA, USA) around the surgical site, while a fifth \acl{ak} camera captures a bird-eye-view of the operating table, similar to the perspective of a camera integrated into overhead OR lights. 
In addition, two \acl{hl} (\acs{hl}, Microsoft Corporation, Redmond, WA, USA) devices capture the egocentric perspectives of the operating surgeon and an assistant, and provide hand pose and eye gaze information.

\paragraph{Ground Truth Generation}

\begin{figure}[t]
\centering
\hspace*{0pt}
\hfill
\includegraphics[height=4cm, width=.48\linewidth, keepaspectratio]{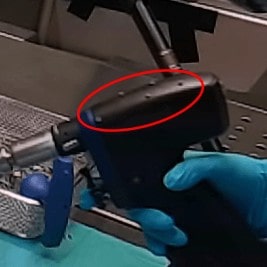}
\includegraphics[height=4cm, width=.48\linewidth, keepaspectratio]{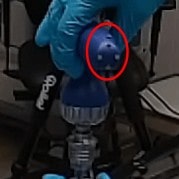}
\hfill
\hspace*{0pt} 
\\ \vspace*{.5mm}
\hspace*{0pt}
\hfill
\includegraphics[height=4cm, width=.48\linewidth, keepaspectratio]{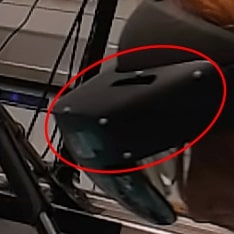}
\includegraphics[height=4cm, width=.48\linewidth, keepaspectratio]{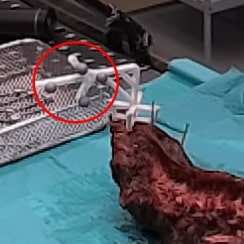}
\hfill
\hspace*{0pt}
\caption{Both instruments, the \acp{hmd}, and the anatomy are tracked with a marker-based tracking system.
We use hemispherical fiducials with \SI{3}{\milli\meter} diameter on instruments and \acp{hmd}, and spherical fiducials with \SI{12}{\milli\meter} diameter for the anatomy.
}
\label{fig:marker_arrays}
\end{figure}

\begin{figure}[t]
    \centering
    \includegraphics[height=6cm, width=\linewidth, keepaspectratio]{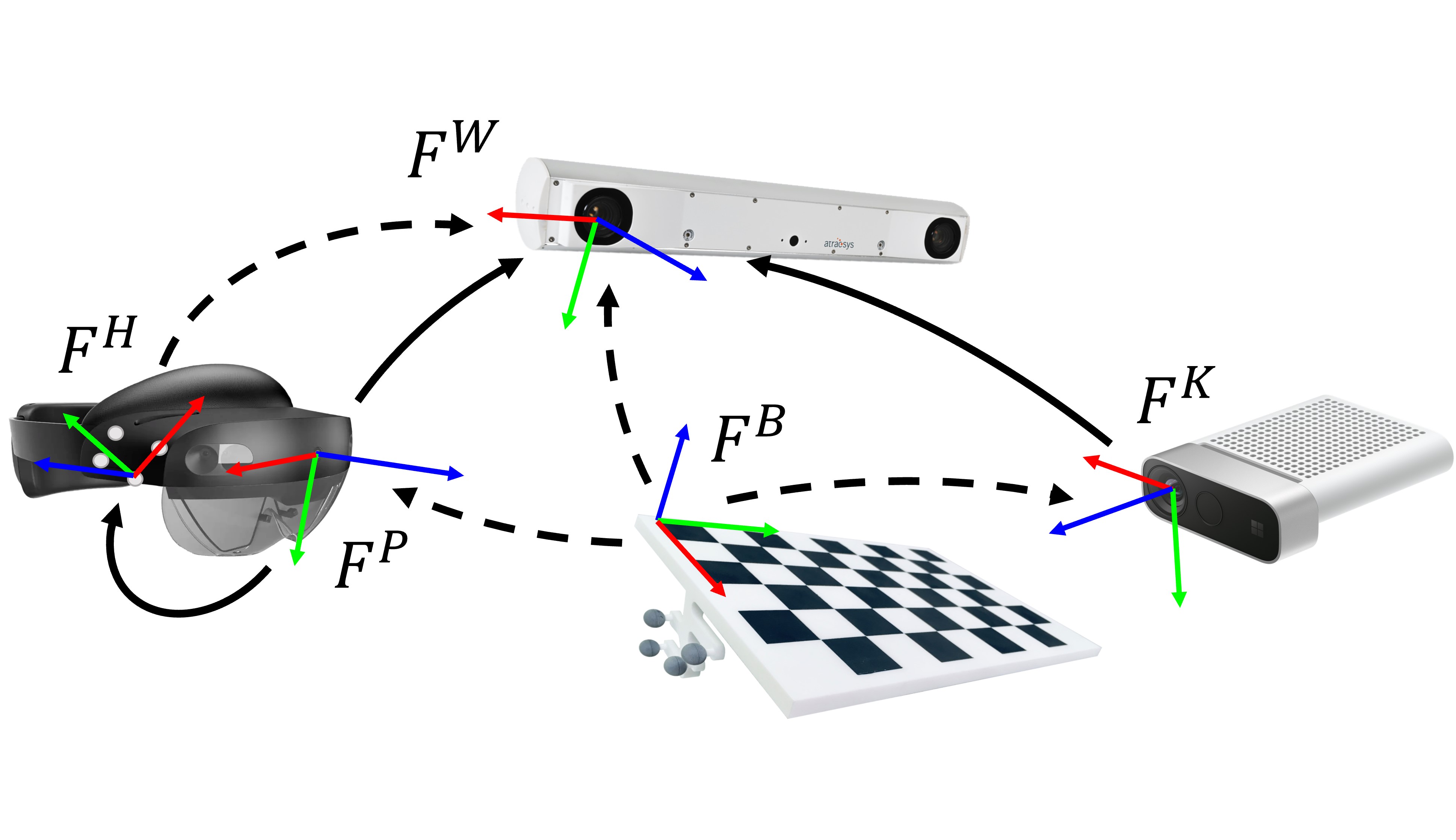}
    \caption{
    Schematic overview of the coordinate frames and transformations used for the joint extrinsic calibration and synchronization. 
    Indicated are relevant transformations between the coordinate frames of the calibration board $F^B$, the tracking system $F^W$, a stationary camera $F^K$, a \ac{hmd} camera $F^P$ and the attached \ac{ir} marker array $F^H$.
    Dashed lines indicate that the transformation is estimated via \acs{pnp}.
    }
    \label{fig:transformations}
\end{figure}

In addition to the aforementioned cameras, we track the surgical instruments and \ac{hl} devices using a FusionTrack 500 marker-based tracking system (Atracsys LLC, Puidoux, Switzerland) to obtain accurate ground truth pose annotations and to circumvent potential errors of the \ac{hl} integrated SLAM system.
As shown in \cref{fig:marker_arrays}, we place small \ac{ir} reflective hemispheres with a diameter of \SI{3}{\milli\meter} on the object surfaces to minimize appearance changes.
To calibrate the attached \ac{ir} marker arrays we acquire 3D models of all instruments and the HoloLenses using a high-fidelity 3D scanner (Artec3D, Senningerberg, Luxembourg).

Besides the instruments, we also track the anatomy via an \ac{ir} marker array attached to the sacrum.
To this end, a post-experimental CT scan was acquired, from which 3D models of the spine anatomy were created by segmentation (Mimics Medical, Materialise NV, Leuven, Belgium).
We used the method proposed by \cite{liebmann2021spinedepth} to register all 3D models to their attached marker coordinate frames.
Although the anatomy pose is not relevant to evaluate instrument pose estimation models, it enables further uses of our dataset.

\subsection{Camera Calibration and Temporal Synchronization} \label{sec:calibration}

An accurate calibration of camera extrinsic and synchronization parameters is crucial when collecting a multi-camera dataset. 
To give an intuition, a synchronization error of \SI{8}{\milli\second} between the devices will result in a position error of \SI{2}{\milli\meter} for a surgical instrument moving with a speed of \SI{0.25}{\meter/\second} relative to the camera. 
We found the synchronization via the host computer's real-time clock to be insufficient due to varying latencies of the devices.
Instead, we jointly optimize extrinsic and synchronization parameters by minimizing the average re-projection error of a moving multi-modal marker $B$ that can be recognized by the tracking system $W$, the \acp{hmd} $P_i$ and static cameras $K_j$ at the beginning of every recording.
Hereby, we directly estimate the offset between all device-internal clocks, using the tracking system $W$ as a reference.
Similarly, we define the tracking system $W$ as the world coordinate frame $F^W$ and co-register all cameras $F^P_i$ and $F^K_j$ with this reference frame using a self-designed multi-modal calibration board $F^B$ similar to the work by \cite{liebmann2021spinedepth}.
A schematic overview of the calibrated transformations is shown in \cref{fig:transformations}.
The extrinsic calibration of static and mobile cameras in our setup differs slightly due to the outside-in tracking of the \acp{hmd}, so we provide both variants in the next paragraphs.
Note that we calibrate the intrinsic parameters of all cameras in a separate step prior to the extrinsic calibration and synchronization using the method by \cite{zhang2000flexible}.
In the following, we denote a transformation from coordinate frame $F^A$ to frame $F^B$ as $T_A^B$, and omit the indices for cameras $P_i$ and $K_j$ to improve readability.

To spatio-temporally register a static camera $K$ with the reference tracking system $W$, we observe a sequence 
\begin{equation}
    \{(x_i^K, t_i^K) ~|~ 1 \leq i \leq N\}
\end{equation}
of 2D marker locations $x_i^K$ and timestamps $t_i^K$ from camera $K$, and a sequence 
\begin{equation}
    \{(T_{B,m}^{W}, t_m^W) ~|~ 1 \leq m \leq M\}
\end{equation}
of 6D marker poses $T_{B,m}^{W}$ and timestamps $t_m^W$ from the reference system $W$.
We piece-wise linearly interpolate the pose sequence, obtaining the function $f_B^W : t_m \rightarrow T_{B,m}^{W}$.
Then, the camera's extrinsic parameters $T_{W}^{K}$ and synchronization parameters $\delta t^K$ can be estimated by minimizing the re-projection error over the entire sequence
\begin{equation}
T_{W}^{K}, \delta t^K = \argmin_{\hat{T_{W}^{K}}, \hat{\delta t}} \sum_{1 \leq i \leq N} \| \pi_K ( \hat{T_{W}^{K}} f_B^W(t_i^K + \hat{\delta t}) X_i) - x_i^K \|_{2},    
\end{equation}
where $\pi_K$ is the projection onto the image plane of camera $K$ and $X_i$ are the 3D marker points in their local coordinate frame $F^B$.

Similarly, for a \ac{hmd} $P$ we observe the sequences
\begin{equation}
    \{(x_i^P, t_i^P) ~|~ 1 \leq i \leq N\}
\end{equation}
of 2D marker locations $x_i^P$ and timestamps $t_i^P$, and 
\begin{equation}
    \{(T_{B,m}^{W}, t_m^W) ~|~ 1 \leq m \leq M\}
\end{equation}
of 6D marker poses $T_{B,m}^{W}$ and timestamps $t_m^W$ from the reference system $W$.
Since each \ac{hmd} $P$ is tracked outside-in, we additionally observe a sequence
\begin{equation}
    \{(T_{H,k}^{W}, t_k^W) ~|~ 1 \leq k \leq K\}
\end{equation}
of 6D \ac{hmd} marker poses $T_{H,k}^{W}$ and timestamps $t_k^W$ from the tracking system $W$, which we piece-wise linearly interpolate to obtain the function $f_H^W : t_k \rightarrow T_{H,k}^{W}$.
Then, the temporal offset $\delta t^P$ can be estimated by minimizing the re-projection error over the entire sequence
\begin{equation}
\delta t^P = \argmin_{\hat{\delta t}} \sum_{1 \leq i \leq N} \| \pi_P(T_{H}^{P} f_H^W(t_i^P + \hat{\delta t})^{-1} f_B^W(t_i^P + \hat{\delta t}) X_i) - x_i^P \|_{2},    
\end{equation}
where $\pi_P$ is the projection onto the image plane of camera $P$, and $T_{P}^{H}$ is the transformation between the \ac{hmd}'s camera sensor $F^P$ and the attached marker array $F^H$, which is calibrated separately beforehand.
Note that the optimization objective can be generalized to mobile cameras with inside-out tracking, however, we decided to use the more accurate outside-in tracking to minimize this source of error in the evaluations.

Both objectives are optimized using LO-RANSAC \citep{chum2003locally} with an inlier threshold of $\theta = 2\textnormal{px}$ and the Levenberg-Marquardt algorithm. 
Due to the limited temporal resolution of the sequences, we locally optimize the temporal offset via a grid search with a step size of \SI{250}{\micro\second}.
Note that since the \acl{ak} supports hardware synchronization, we only optimize the time shift $\delta t^K$ of the first device and keep it fixed for all remaining ones.

\paragraph{Ground Truth Quality}
We evaluate the accuracy of the camera extrinsic calibration and synchronization by comparing the calibration board corner locations as detected in the camera images with their corresponding ground truth positions.
The average re-projection error is \num{1.82}px, which corresponds to mean errors of \SI{0.88}{\milli\meter} and \SI{0.83}{\milli\meter} along the camera's X and Y axes, respectively.
Note that these errors include both spatial and temporal calibration errors as they refer to a moving target.

\subsection{Surgery Datasets} \label{sec:datasets}

\paragraph{Surgical Wet lab}
To evaluate our approach, we record the instrumentation phase of spinal fusion surgery using the presented multi-camera acquisition setup in a surgical wet lab.
Spinal instrumentation consists of pre-drilling a screw trajectory, implantation, and removal of a pedicle screw implant.
Hereby, we use a Colibri II battery-powered drill (DePuy Synthes, Raynham, MA, USA) for pre-drilling, and a polyaxial pedicle screwdriver (Medacta SA, Castel San Pietro, Switzerland) for screw insertion.
Both instruments are subject to our marker-less pose estimation system.
Screw implantation is conducted on three human specimens between T10 and L5 vertebrae by one trained surgeon and three researchers using pre-drilled optimal screw trajectories.

The static cameras capture RGB frames with a resolution of $2048 \times 1536$ pixels and 30 \ac{fps}.
\rev{Both \acp{hmd} capture RGB frames with a resolution of $896 \times 504$px and 30 \ac{fps}, as well as depth frames in the AHAT and long-throw mode.
The effective frame rate varies due to dropped frames, especially for the AHAT depth.
During post-processing, we pair each RGB frame with the temporally closest depth frame and transform the depth map into the RGB camera frame via the calibrated extrinsics and nearest-neighbor interpolation.}

The dataset contains a total of 21 recordings with \SI{1.7}{\mega\relax} frames.
Each recording consists of a varying number of pre-drilling, screw implantation, and removal steps, in random order.
Also, the scrubs and glove colors are randomized to increase the image diversity.
We split the dataset into 17 training recordings and 4 test recordings.
From the 4 test recordings we sample \revTwo{6880} multi-view image sets with 7 camera views each, for a total of \revTwo{48160 RGB-D} frames.

In the sampling process, we ensure that the pair-wise temporal offset between RGB exposure windows within each multi-view set is at most \SI{8}{\milli\second}.
This filtering step is necessary because we can only synchronize the device-internal clocks but not the camera shutters.
In contrast to the \acl{ak}, neither the \ac{hl} nor the FusionTrack 500 support any hardware-synchronization of the \ac{pv} camera shutter, e.g. with an external trigger signal.
As such, there will be varying temporal offsets of up to \SI{16.7}{\milli\second} between pairs of captured images from multiple cameras (assuming $30$ \ac{fps}). 
These temporal offsets break the underlying assumptions of multi-view pose estimation models and may introduce additional errors depending on the dynamics in the scene.
We evaluate the effect of this temporal offset in the appendix but find no significant correlation between the temporal offset of image pairs and the accuracy of the multi-view pose estimates in our experiments.

\paragraph{Synthetic Dataset}

\begin{figure}[t]
\centering
\hspace*{0pt}
\hfill
\includegraphics[height=35mm, width=.48\linewidth, keepaspectratio]{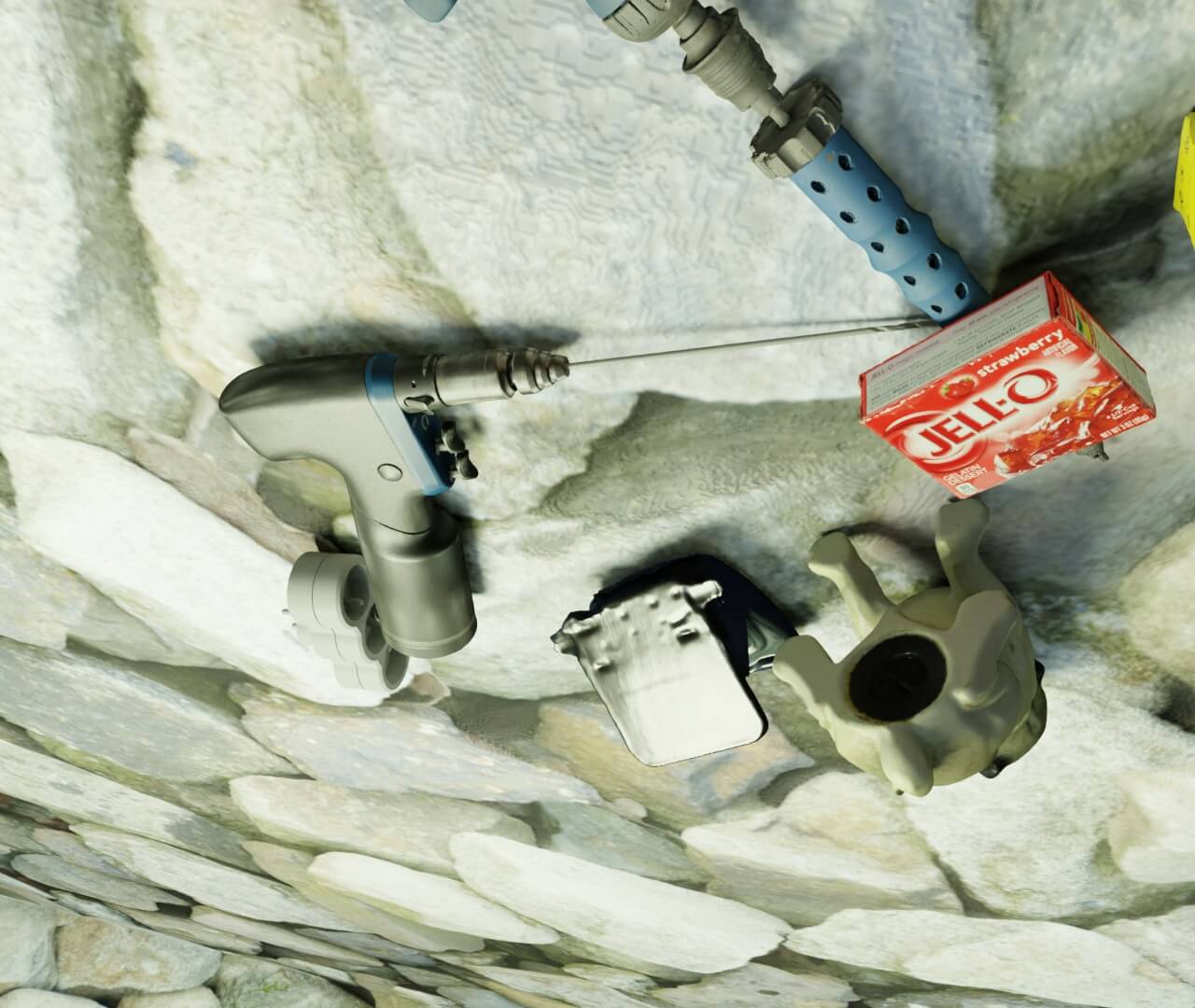}
\includegraphics[height=35mm, width=.48\linewidth, keepaspectratio]{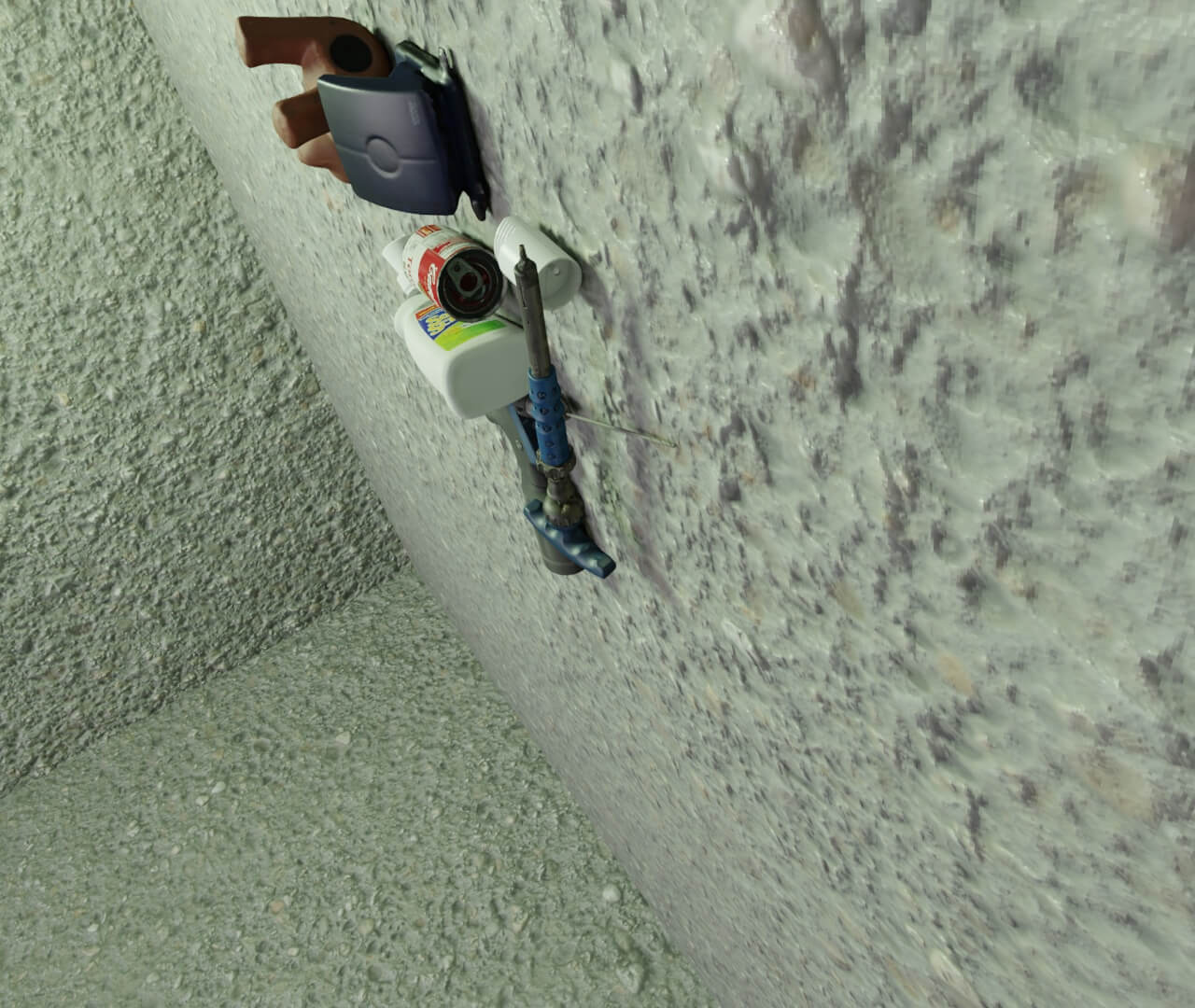}
\hfill
\hspace*{0pt} 
\\ \vspace*{.5mm}
\hspace*{0pt}
\hfill
\includegraphics[height=35mm, width=.48\linewidth, keepaspectratio]{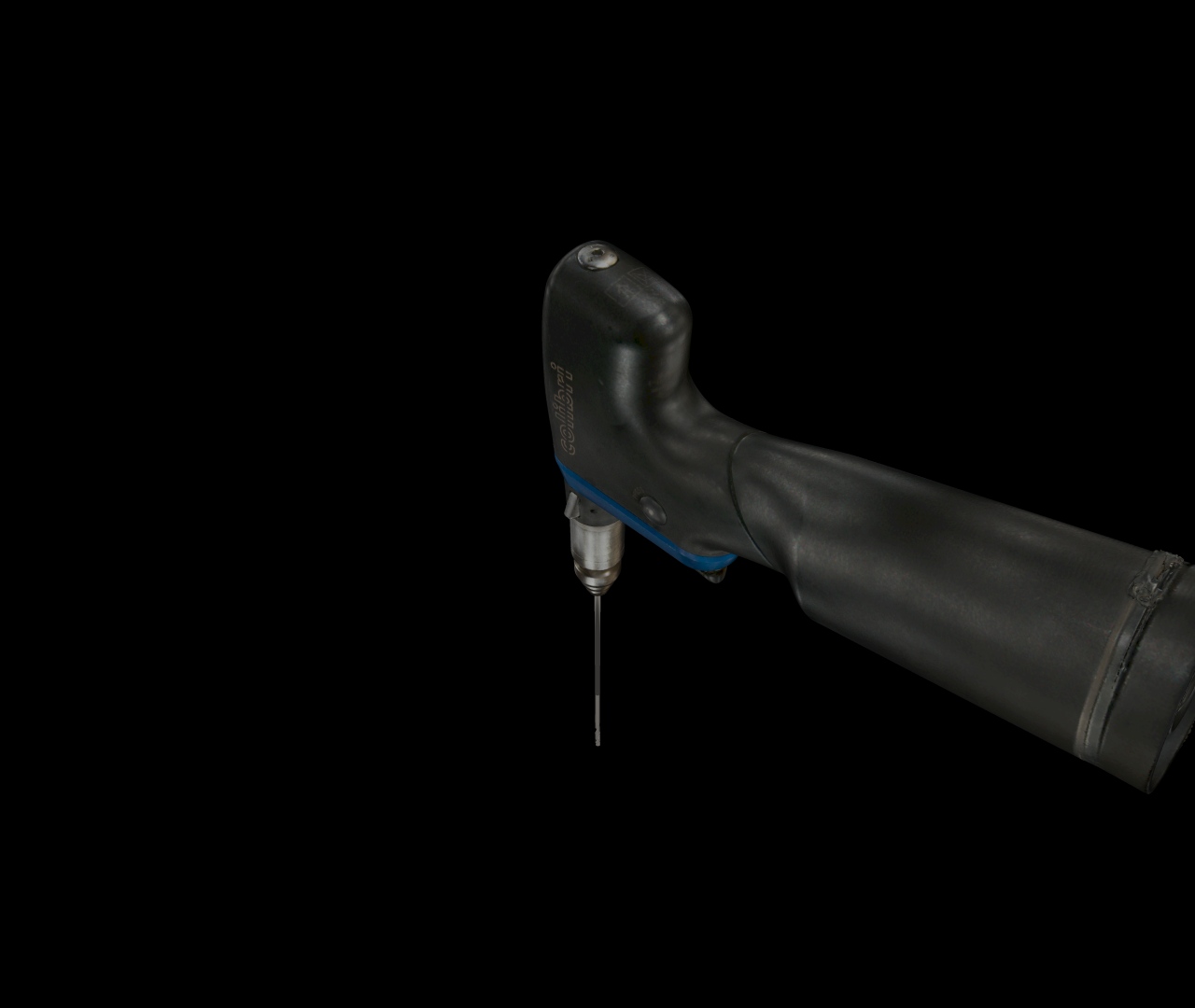}
\includegraphics[height=35mm, width=.48\linewidth, keepaspectratio]{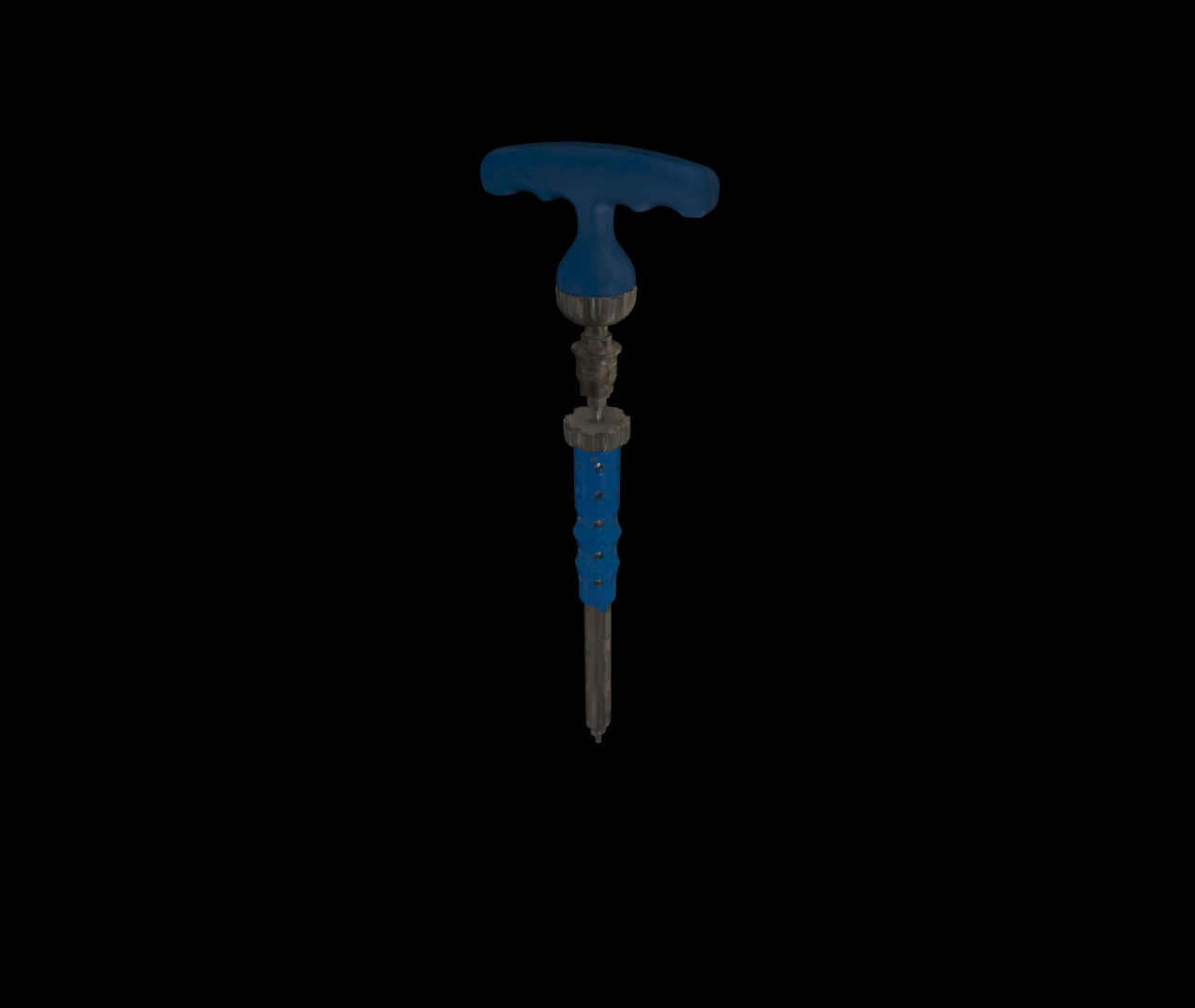}
\hfill
\hspace*{0pt}
\caption{Exemplary synthetic images generated with BlenderProc2 (top) and an OpenGL-based renderer without shading (bottom).
}
\label{fig:synthetic-frames}
\end{figure}

In addition to the real images, we render synthetic images of the instruments to support the training process \citep{movshovitz2016useful}.
We generate \SI{25}{\kilo\relax} renderings from uniformly sampled poses with a distance between \SIrange[multi-part-units=single]{0.4}{1.7}{\meter}, matching the distance range of the wet lab dataset.
Additionally, we provide \SI{38}{\kilo\relax} photo-realistic renderings generated by BlenderProc2 with the same pose sampling strategy \citep{Denninger2023}.
\rev{Exemplary renderings are shown in \cref{fig:synthetic-frames}.
We uniformly sample the light color, position, and intensity from manually defined intervals in order to obtain a neutral illumination on average. 
All synthetic frames are rendered using camera intrinsics similar to those of the Azure Kinect or HoloLens PV cameras, and show the instruments in the same articulation and without any \ac{ir} markers.
We do not include include surgical background images but show random textures from the CC0 texture library\footnote{https://ambientcg.com/} or a black background.
While backgrounds with surgical environments may look more realistic, we observed that more diverse and readily available datasets of generic textures are sufficient to train models to be invariant to the background.
}

\begin{figure}[t]
\centering
\hspace*{0pt}
\hfill
\includegraphics[height=5cm, width=0.61\linewidth, keepaspectratio]{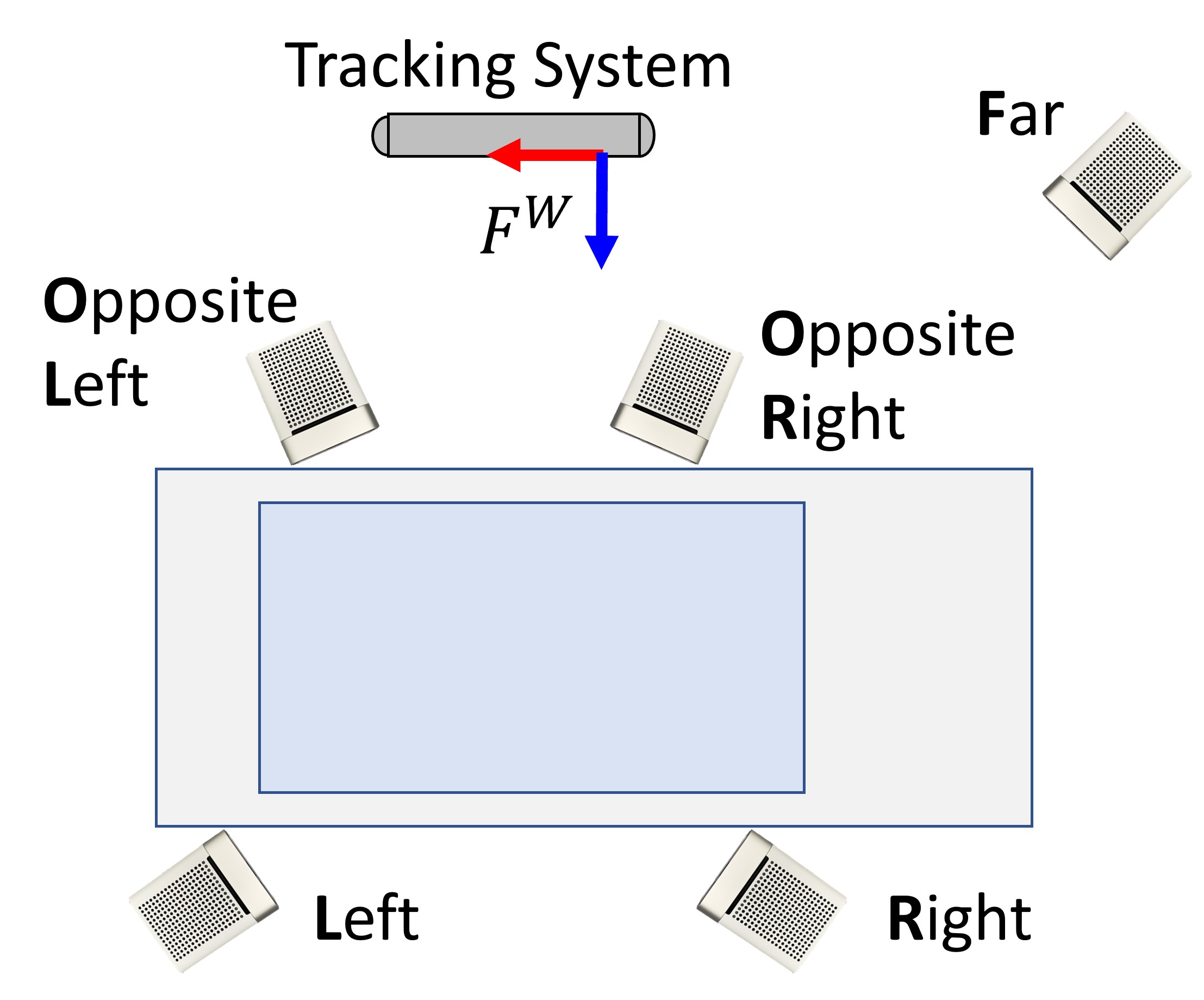}
\hfill
\adjincludegraphics[height=5cm, width=0.35\linewidth, trim={{0.2\width} {0.1\height} {0.1\width} {0.25\height}}, clip, keepaspectratio]{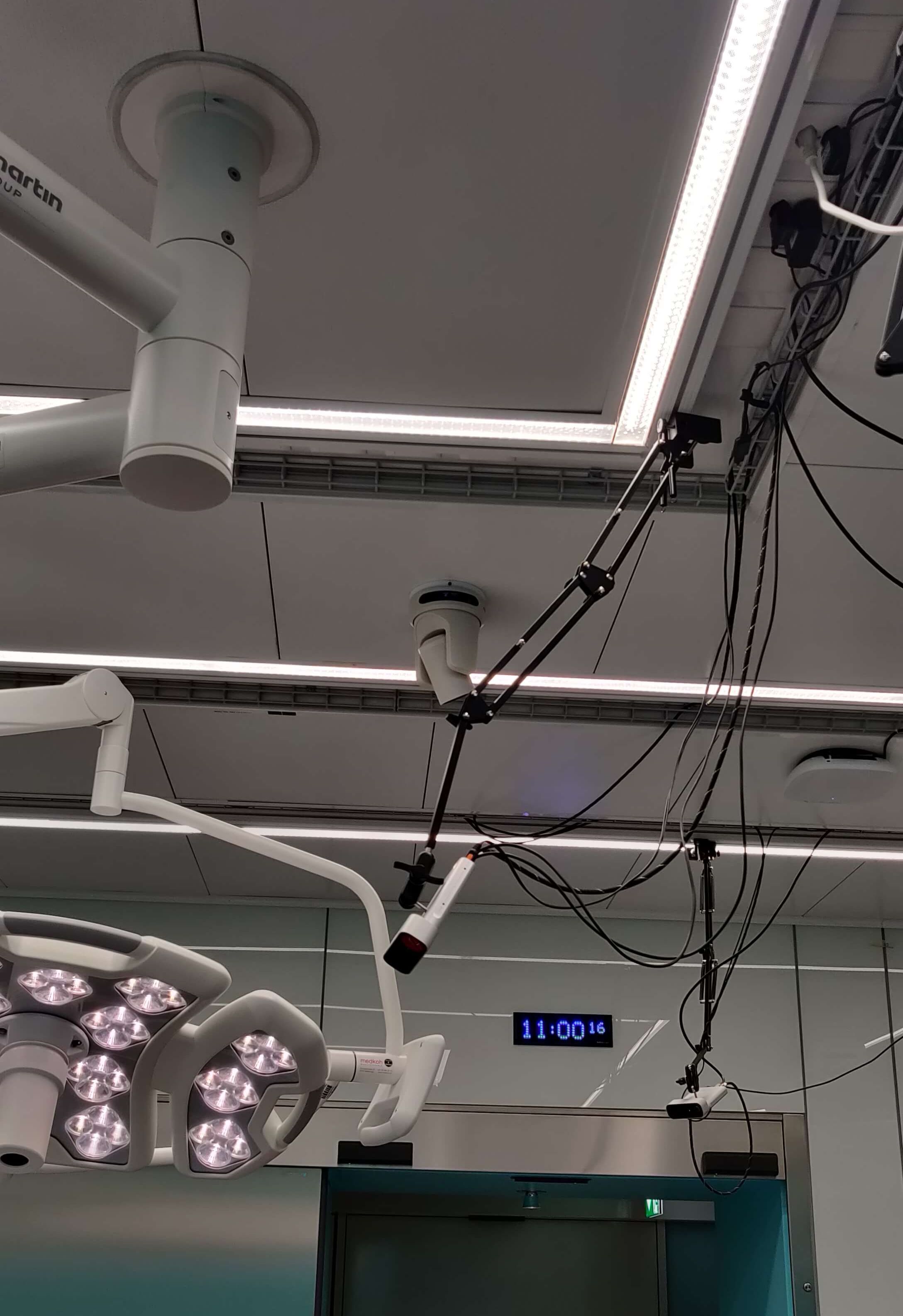}
\hspace*{0pt}
\caption{Schematic overview of the camera setup for the OR-X test set (left) and an exemplary ceiling-mounted camera in the OR-X (right).
}
\label{fig:camera_setup_orx}
\end{figure}

\begin{figure*}[t]
\centering
\hspace*{0pt}
\includegraphics[width=0.192\linewidth, keepaspectratio]{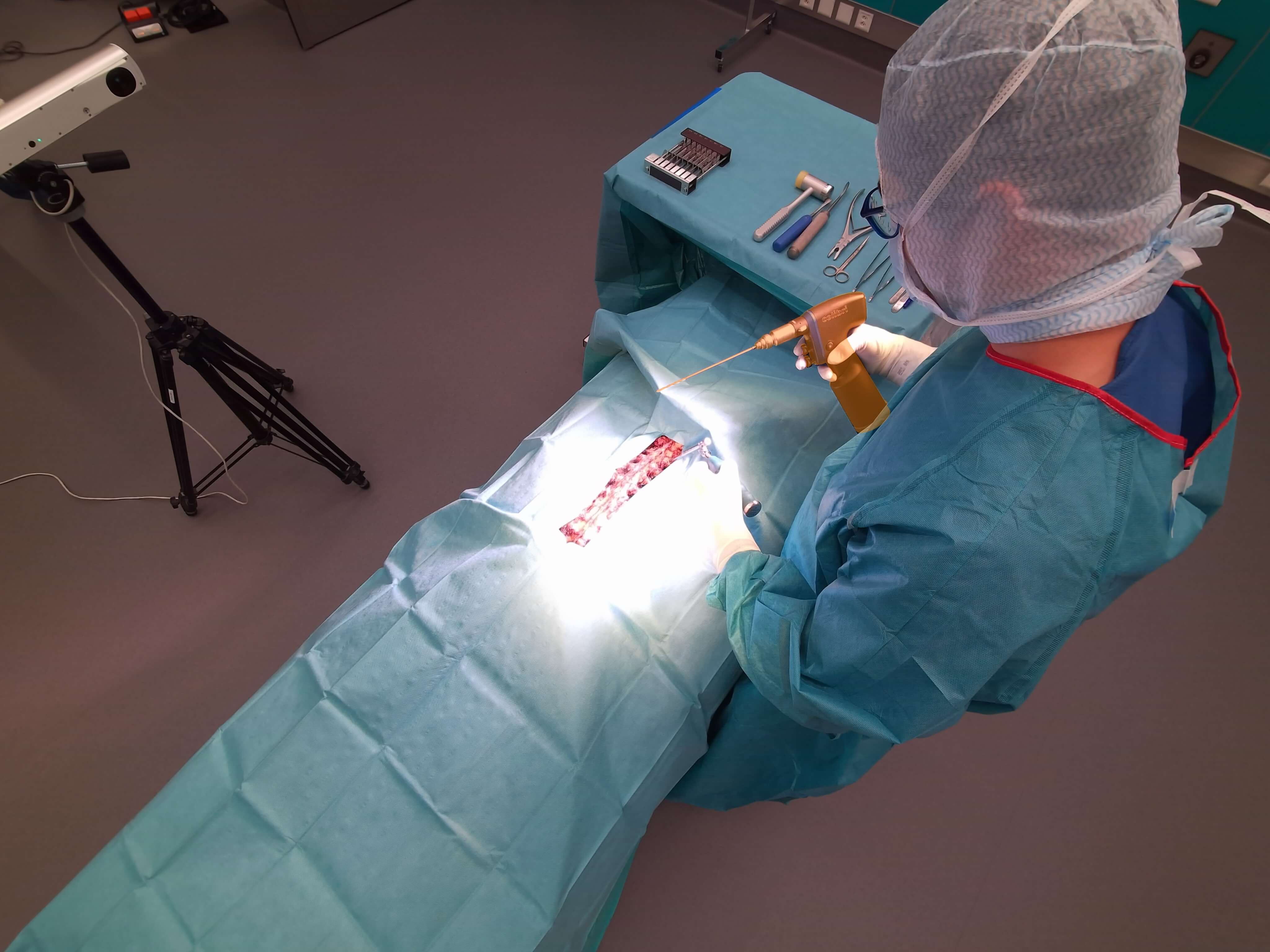}
\hfill
\includegraphics[width=0.192\linewidth, keepaspectratio]{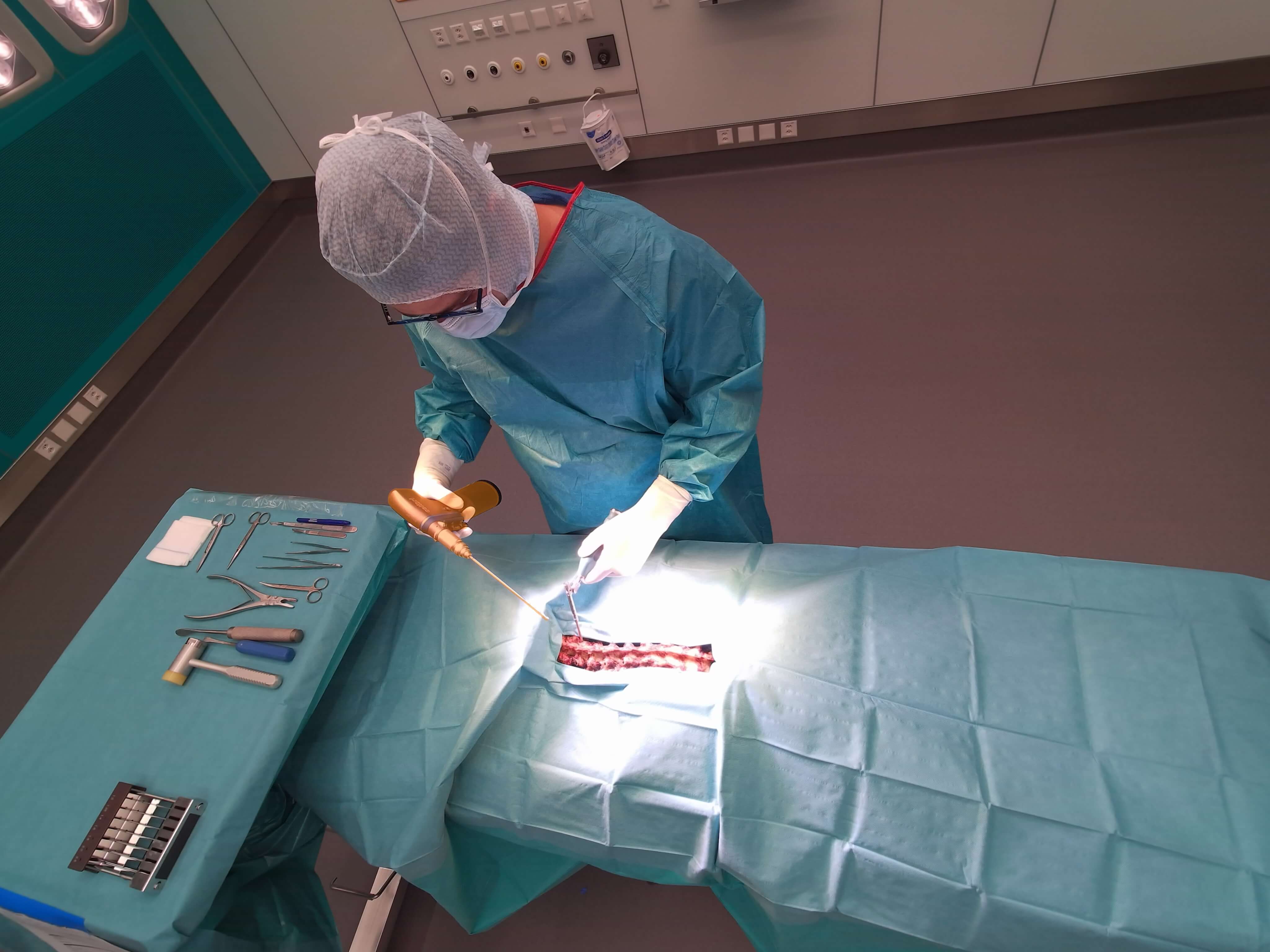}
\hfill
\includegraphics[width=0.192\linewidth, keepaspectratio]{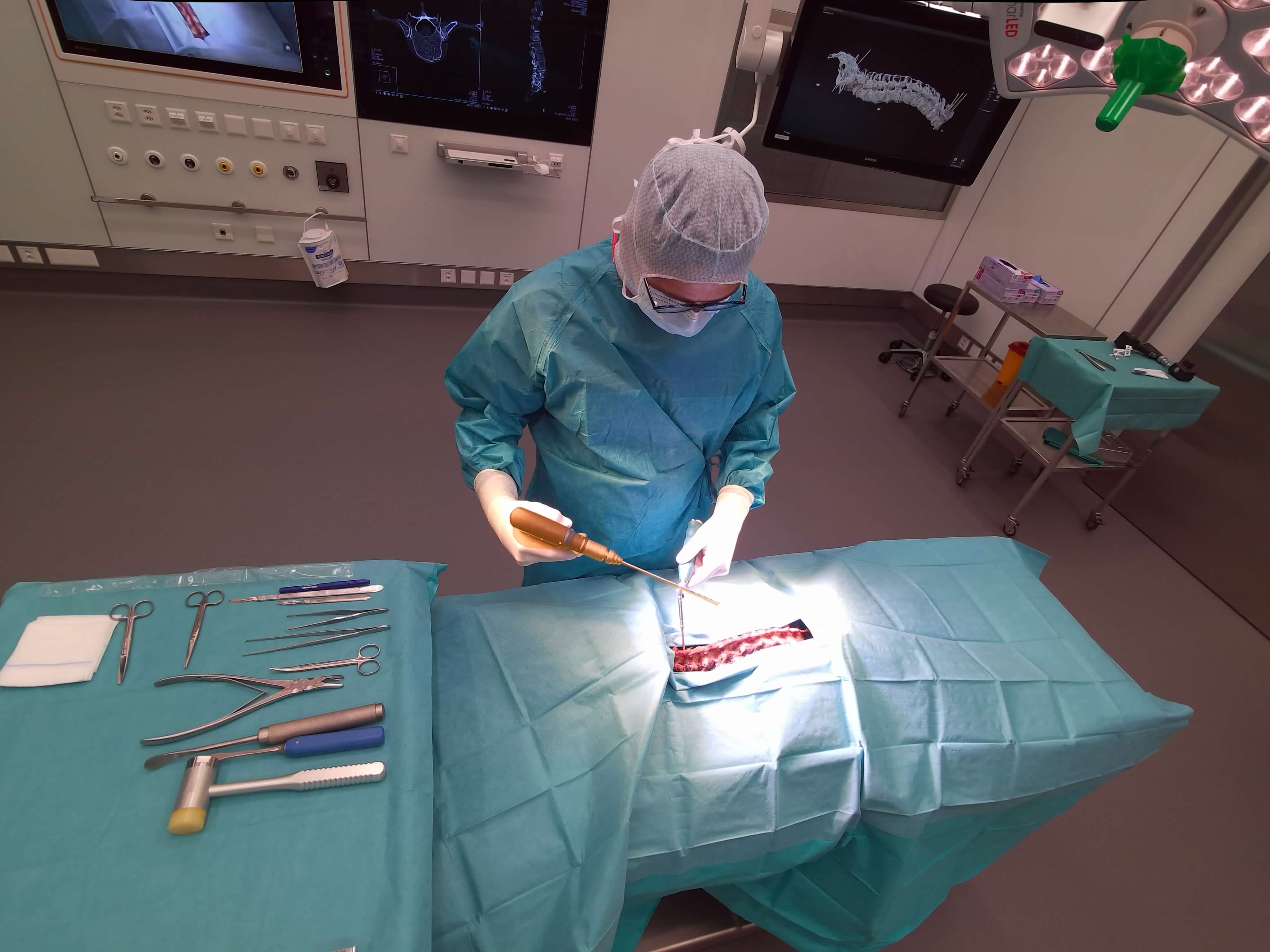}
\hfill
\includegraphics[width=0.192\linewidth, keepaspectratio]{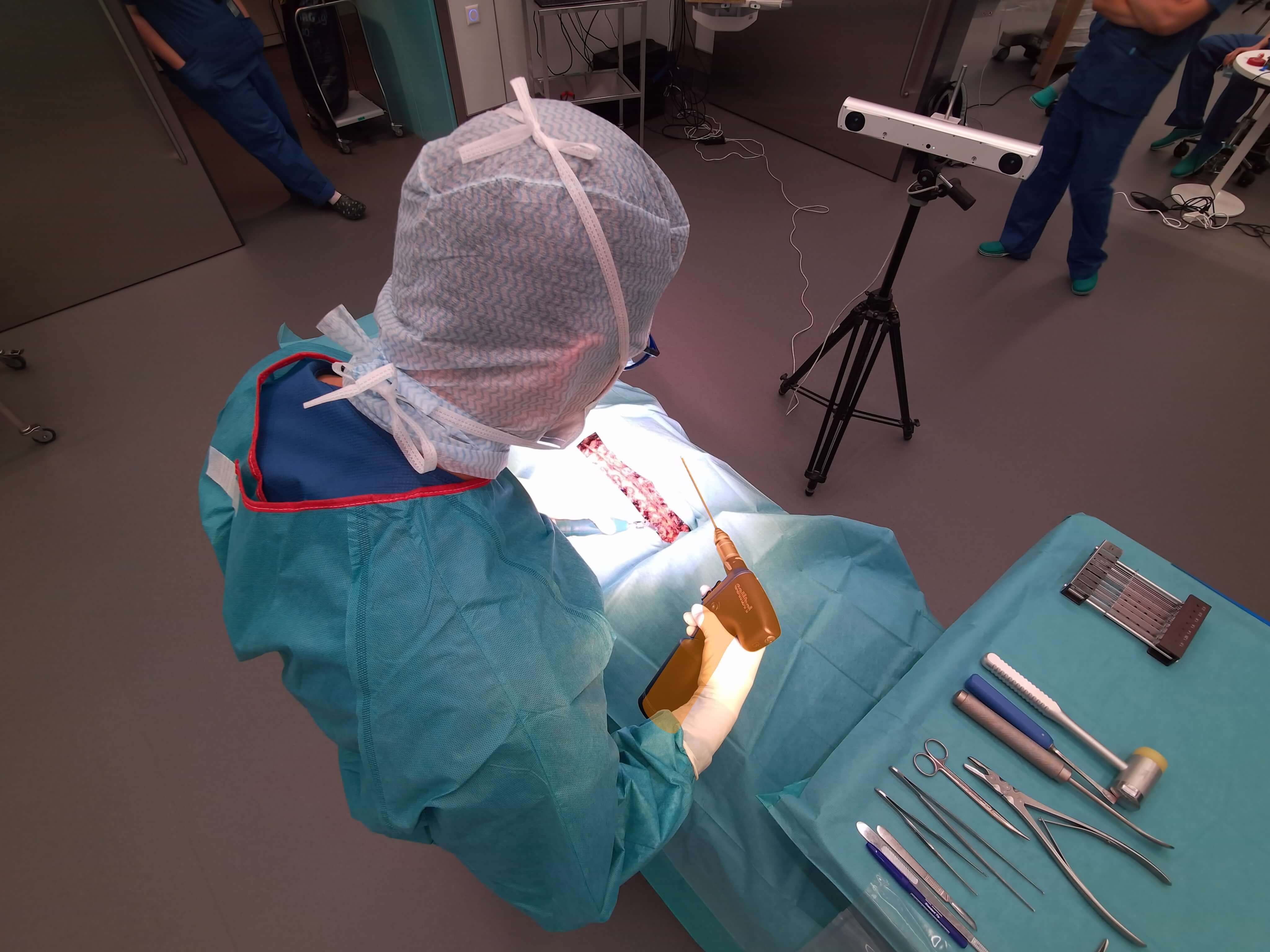}
\hfill
\includegraphics[width=0.192\linewidth, keepaspectratio]{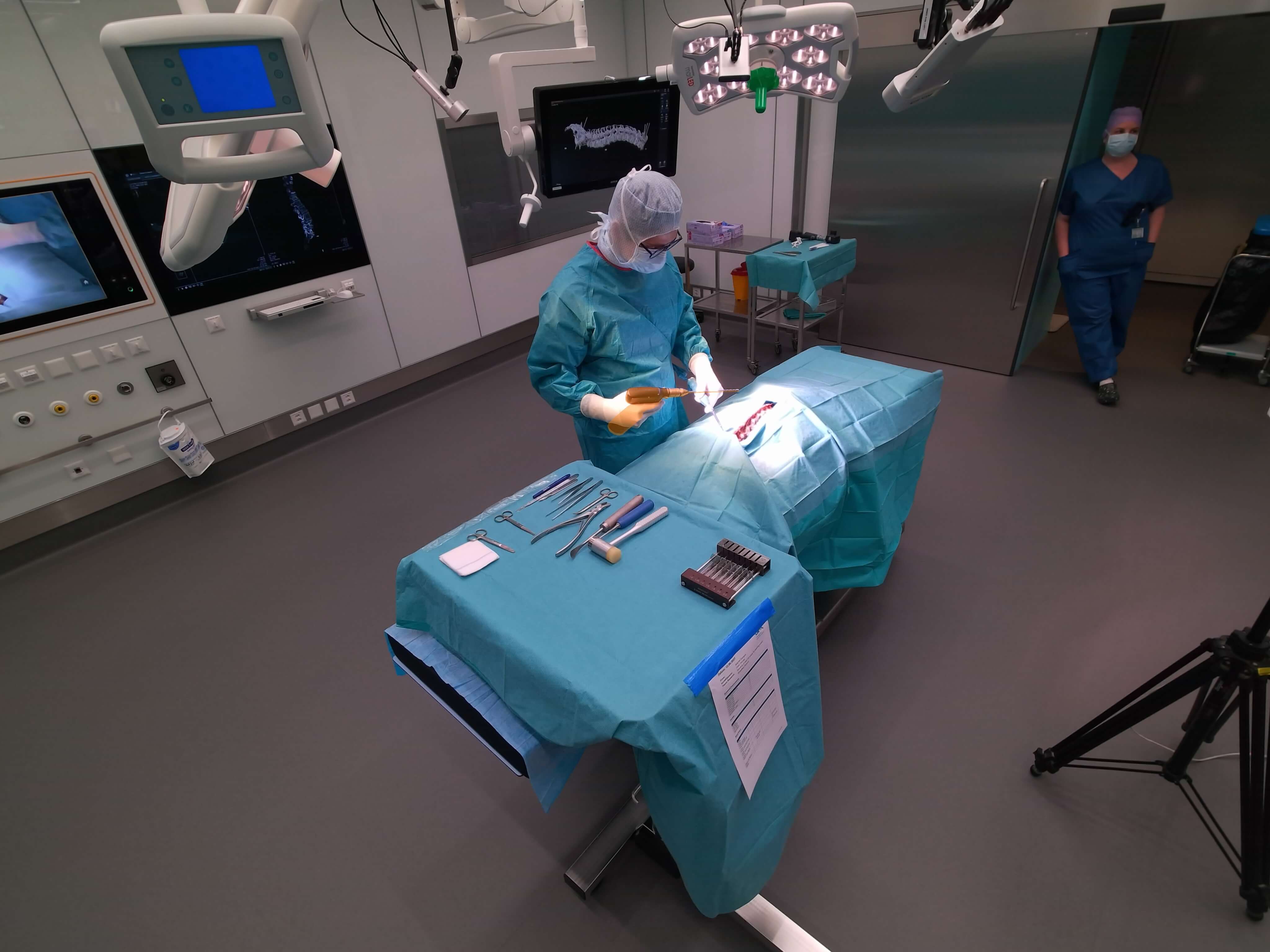}
\hspace*{0pt}
\\ \vspace*{.5mm}
\hspace*{0pt}
\includegraphics[width=0.192\linewidth, keepaspectratio]{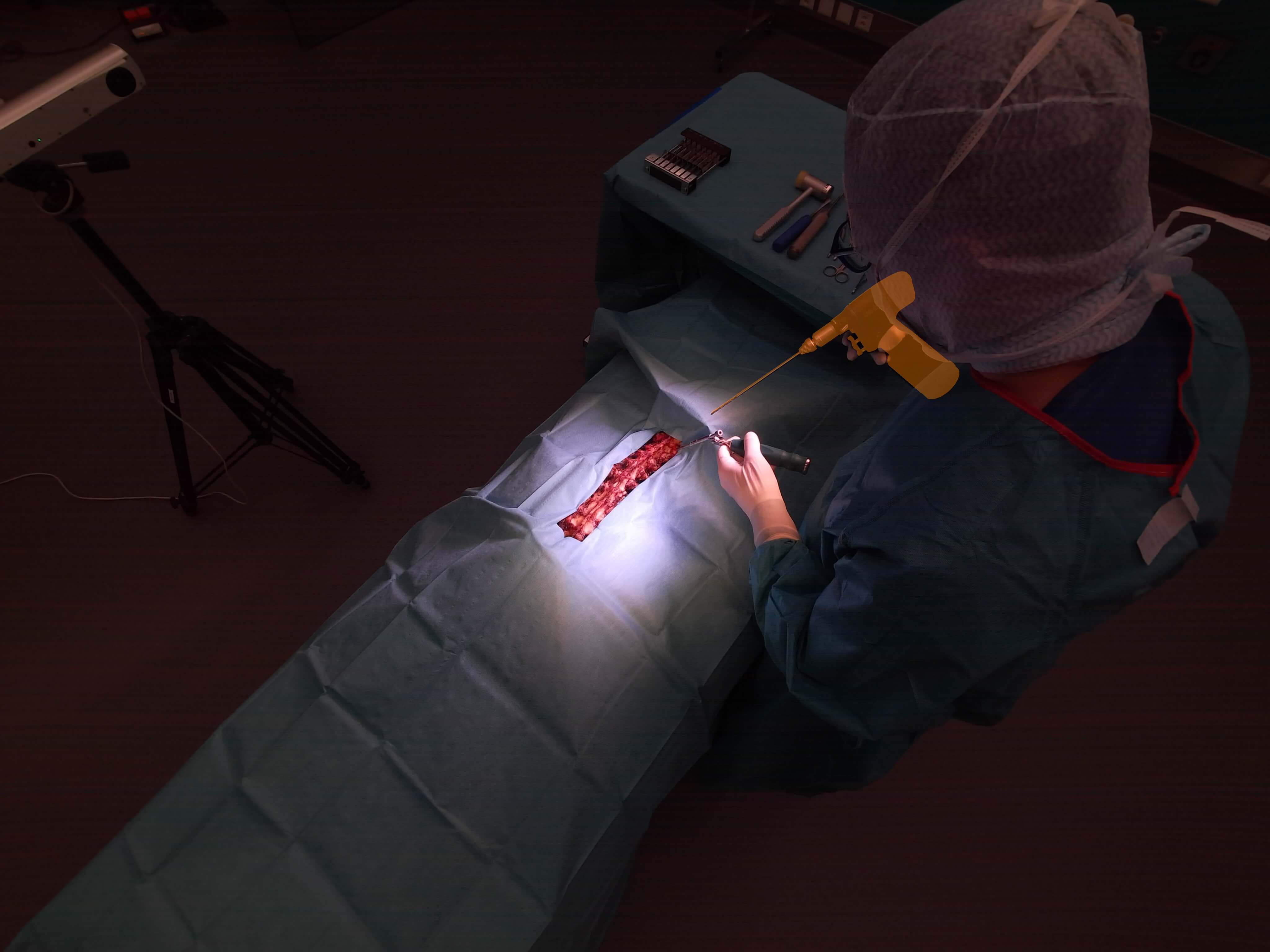}
\hfill
\includegraphics[width=0.192\linewidth, keepaspectratio]{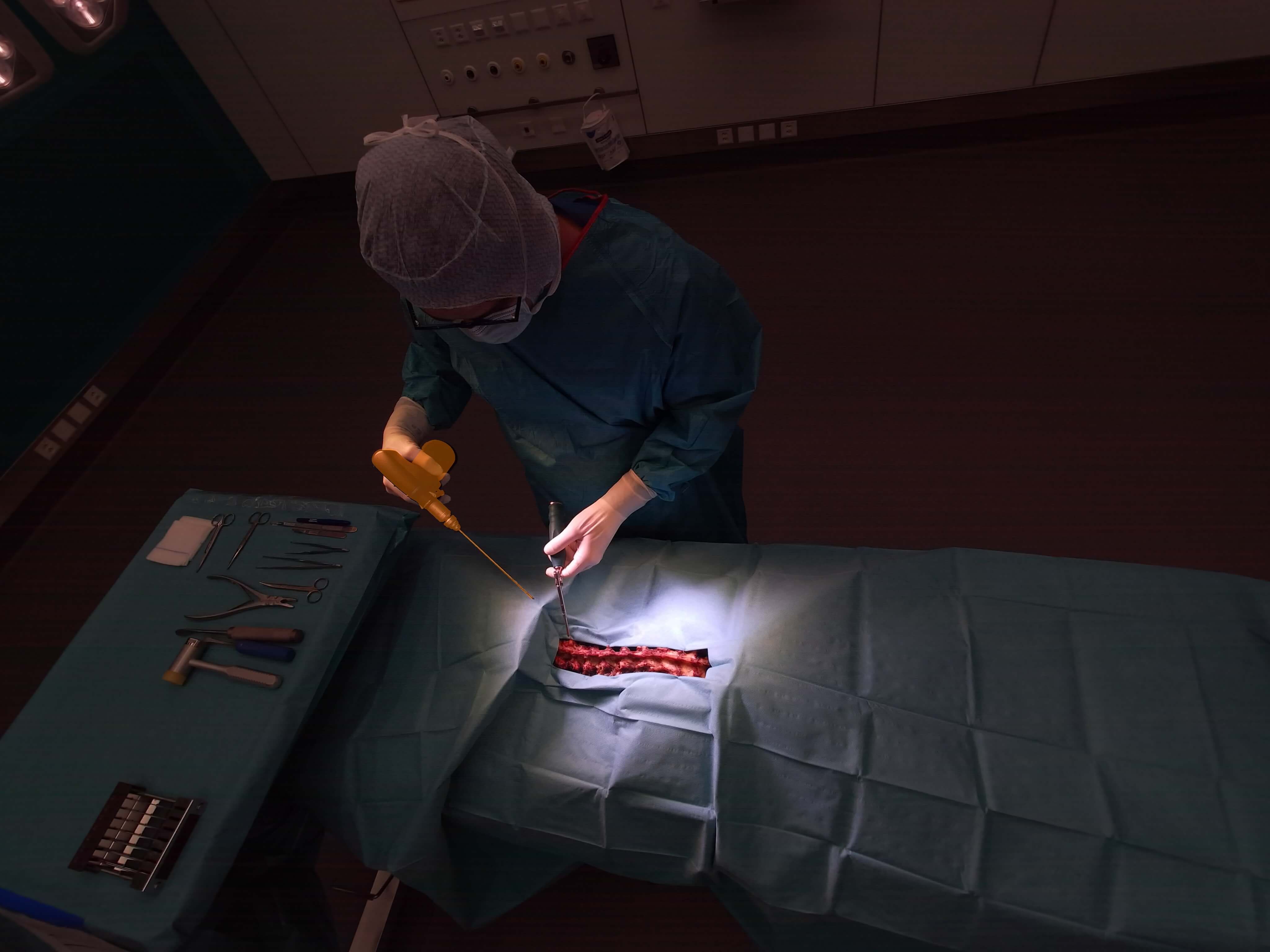}
\hfill
\includegraphics[width=0.192\linewidth, keepaspectratio]{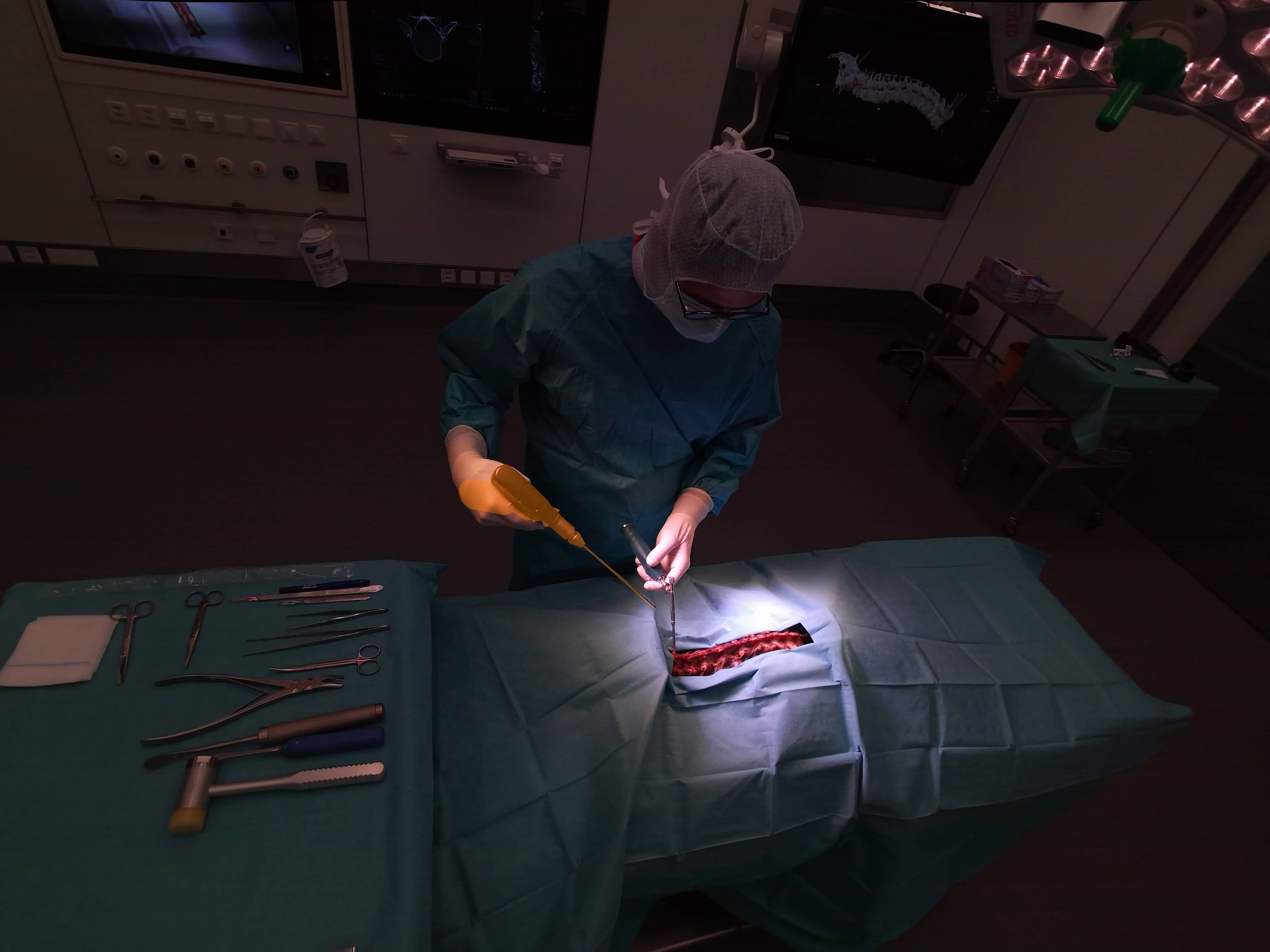}
\hfill
\includegraphics[width=0.192\linewidth, keepaspectratio]{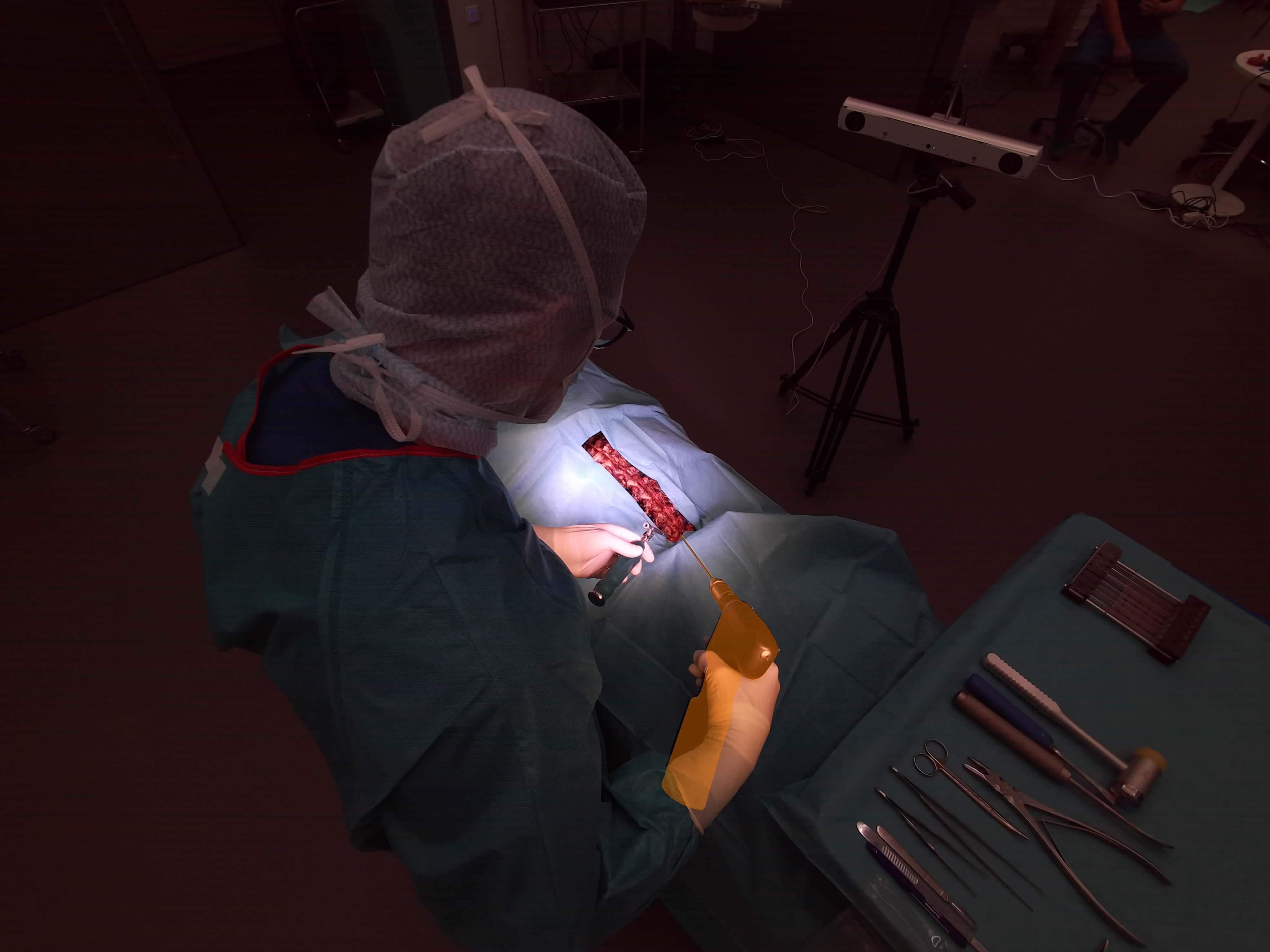}
\hfill
\includegraphics[width=0.192\linewidth, keepaspectratio]{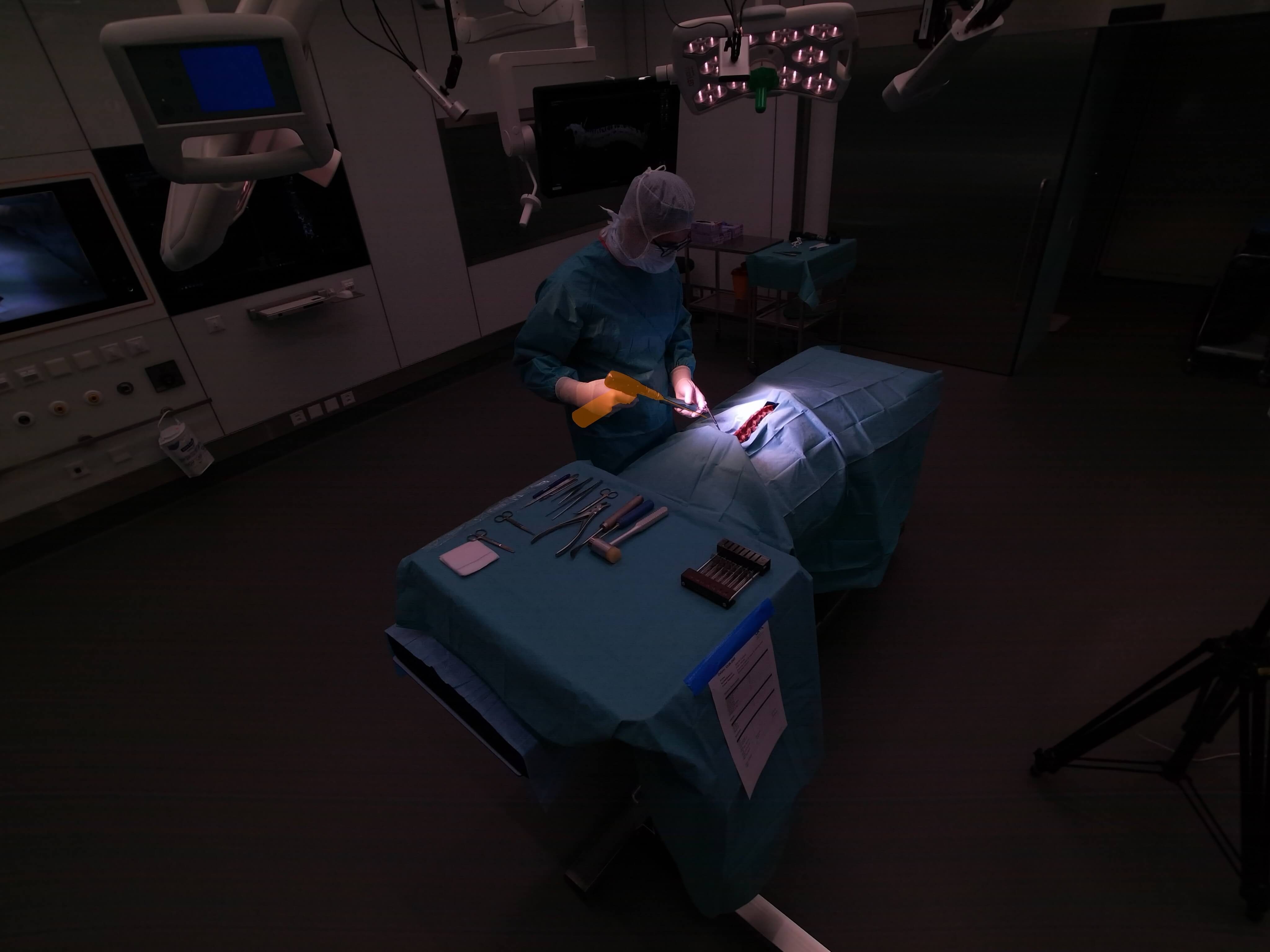}
\hspace*{0pt}
\caption{Comparison of the camera views and exposure windows used in the OR-X test set, with the multi-view pose estimates of EpiSurfEmb superimposed.
The shown cameras are (left-to-right) left (L), opposite left (OL), opposite right (OR), right (R), and far (F).
The first recording (top row) was captured with a longer exposure time that is optimal for acquiring the entire surgery room environment, but results in an overexposed surgical near field.
The second recording (bottom row) was captured with a shorter exposure time optimal for the surgical near field, but the environment is generally underexposed.
}
\label{fig:orx-example-frames}
\end{figure*}

\paragraph{OR-X Test Set}
To show that our system can be translated to a realistic environment, we additionally capture a second test set in the OR-X\footnote{https://or-x.ch/}, a real operating theatre dedicated to research use.
This test set consists of five \acl{ak} cameras attached to the ceiling around the operating table, as illustrated in \cref{fig:camera_setup_orx}.
All cameras are mounted above head height to minimize the invasiveness of our setup.
We collect two subsets totaling about 25k frames, where each subset consists of one recording of pedicle screw pre-drilling with the Colibri II
\footnote{For logistic reasons the OR-X test set does not include tracked anatomy or the pedicle screwdriver. \acp{hmd} were excluded due to their poor performance on preliminary experiments.}.
In both subsets, the drill is operated by a surgeon in training.
The calibration, synchronization, and data processing are carried out identically to the generation of the training dataset.

Besides the more realistic environment, the OR-X test set has additional and intentional differences compared to the wet lab dataset which enables the evaluation of the model's robustness and generalizability.
First, both subsets were captured with different exposure settings to obtain significantly differing and more challenging lighting conditions.
Second, camera-to-camera and camera-to-instrument distances are significantly larger compared to the wet lab setup with cameras rigidly fixed on arms attached to the ceiling.
To compensate for the increased distance range of about \SIrange[multi-part-units=single]{0.9}{2.8}{\meter}, we record this test set in 4K resolution instead of 1536p.
Third, the drilling in the OR-X test set is conducted with the help of a drill sleeve, which is not used in the wet lab dataset.
Exemplary frames can be seen in \cref{fig:orx-example-frames}.

This test dataset is used exclusively to analyze the robustness and generalizability of models to novel environments.
As such, it is not used for training or refinement in any experiment.

\subsection{Pose Estimation Baselines} \label{sec:baselines}

\begin{figure}[t]
\centering
\hspace*{0pt}
\hfill
\adjincludegraphics[height=6cm, width=0.48\linewidth, keepaspectratio, trim={{0.0\width} {0.0\height} {0.0\width} {0.0\height}}, clip]{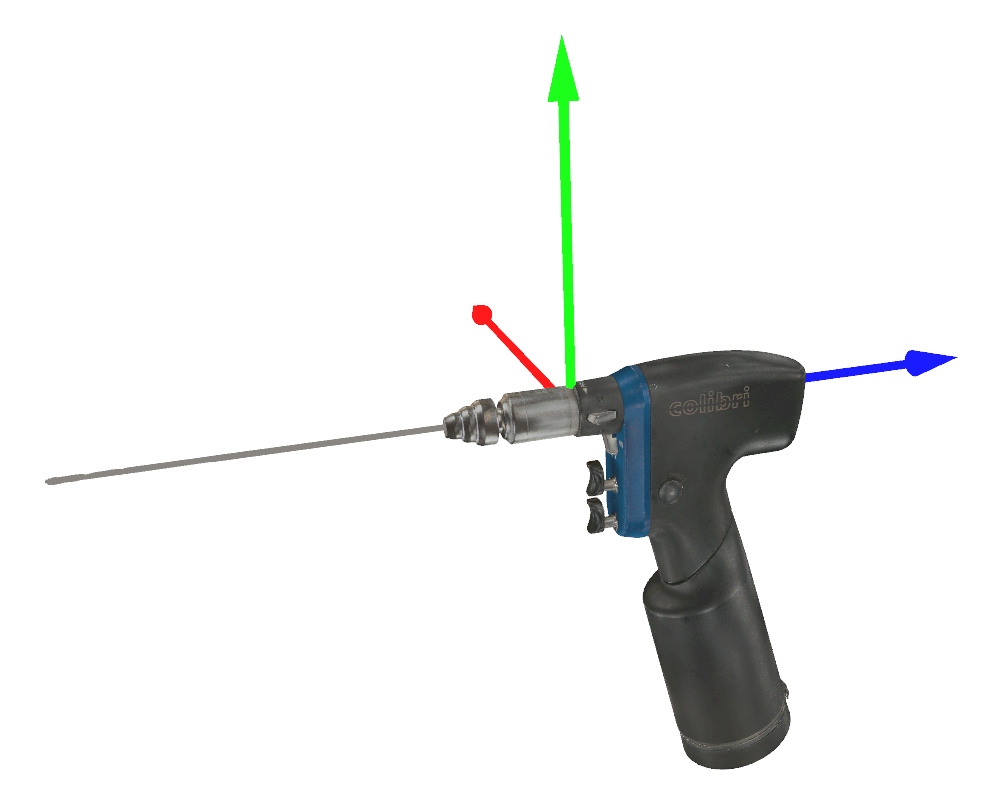}
\hfill
\adjincludegraphics[height=6cm, width=0.48\linewidth, keepaspectratio, trim={{0.0\width} {0.0\height} {0.0\width} {0.0\height}}, clip]{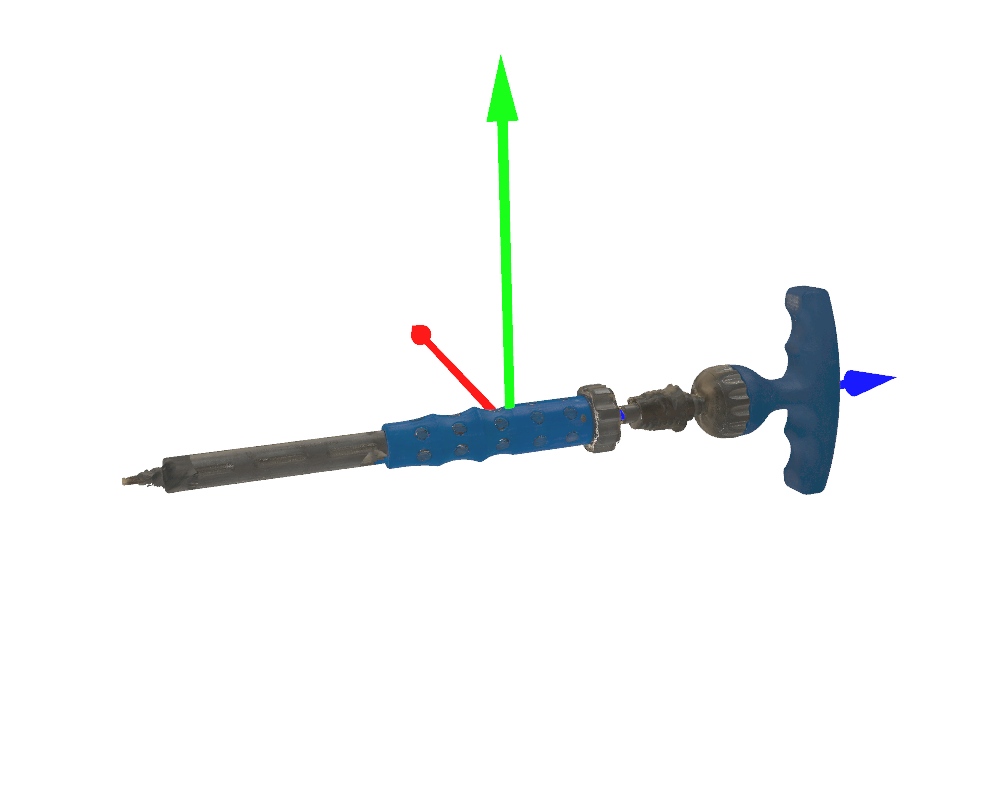}
\hfill
\hspace*{0pt}
\caption{Local coordinate frames of the tracked surgical drill (left) and screwdriver (right). 
}
\label{fig:tool-coordinate-frames}
\end{figure}

\begin{table*}[t]
\centering
\caption{
We report the ADD(-S), position, and orientation errors of the single-view RGB(-D) and multi-view RGB baselines as mean $\pm$ std. 
The orientation error $\Delta R$ is defined as the geodesic distance between the estimated and ground truth rotation.
Bold and underlined font indicates the lowest and second-lowest average error, respectively.
Single-view results are averaged over all cameras.
Both single-view baselines estimated a few ($<0.7\%$) unrealistic poses with \SI{>1}{\meter} position errors, which were excluded from the evaluation.
The camera identifiers are opposite left (OL), opposite right (OR), ceiling (C), left (L), right (R), surgeon (S), and assistant (A), as shown in \cref{fig:camera_setup}. 
* indicates hybrid camera setups with static and mobile cameras.
$\dagger$ indicates fully mobile camera configurations.
}
\begin{tabular}{lcccccc}
\hline
\multicolumn{1}{c}{\multirow{2}{*}{Model}} & \multicolumn{3}{c}{Drill} & \multicolumn{3}{c}{Screwdriver} \\
\multicolumn{1}{c}{} & ADD (mm) $\downarrow$ & $\Delta t$ (mm) $\downarrow$ & $\Delta R$ (deg) $\downarrow$ & ADD-S (mm) $\downarrow$ & $\Delta t$ (mm) $\downarrow$ & $\Delta R$ (deg) $\downarrow$
 \\ \hline
ZebraPose & $12.57 \pm 29.19$ & $11.13 \pm 32.35$ & $3.69 \pm 12.03$ & $22.58 \pm 44.15$ & $41.44 \pm 88.10$ & $15.75 \pm 21.76$ \\
ZebraPose + ICP & $48.58 \pm 62.88$ & $47.38 \pm 76.64$ & $21.05 \pm 26.23$ & $17.16 \pm 33.75$ & $37.77 \pm 82.81$ & $25.17 \pm 26.64$ \\
SurfEmb & $12.46 \pm 21.64$ & $11.04 \pm 27.16$ & $3.37 \pm 6.66$ & $12.70 \pm 25.33$ & $25.65 \pm 56.37$ & $12.98 \pm 17.23$ \\
SurfEmb RGB-D & $20.80 \pm 41.95$ & $24.59 \pm 68.40$ & $3.40 \pm 6.98$ & $11.80 \pm 22.23$ & $24.75 \pm 53.73$ & $13.05 \pm 17.40$ \\ \hline
\multicolumn{7}{l}{EpiSurfEmb (multi-view, trained on synthetic and real data)} \\
OL+OR & $2.26 \pm 1.25$ & $1.34 \pm 0.83$ & $1.08 \pm 0.69$ & $1.73 \pm 1.60$ & $3.29 \pm 3.22$ & $5.37 \pm 5.14$ \\
L+OL+OR+R & \underline{$2.02 \pm 1.16$} & $1.15 \pm 0.73$ & $1.00 \pm 0.62$ & $1.53 \pm 1.46$ & \underline{$2.91 \pm 2.81$} & $4.77 \pm 4.10$ \\
L+OL+OR+R+C & $2.02 \pm 1.22$ & \underline{$1.06 \pm 0.71$} & \underline{$0.95 \pm 0.61$} & \underline{$1.47 \pm 1.44$} & $2.95 \pm 2.82$ & $\mathbf{3.29 \pm 2.65}$ \\
L+OL+OR+R+C+S+A* & $2.14 \pm 1.22$ & $1.22 \pm 0.81$ & $1.00 \pm 0.60$ & $1.52 \pm 1.44$ & $3.11 \pm 2.83$ & $3.48 \pm 2.82$ \\
L+C & $3.56 \pm 2.99$ & $2.13 \pm 1.48$ & $1.56 \pm 1.63$ & $2.09 \pm 2.05$ & $4.59 \pm 4.36$ & $4.00 \pm 5.66$ \\
R+A* & $4.20 \pm 2.48$ & $3.35 \pm 2.59$ & $1.66 \pm 1.11$ & $2.47 \pm 1.90$ & $5.17 \pm 4.20$ & $7.44 \pm 7.35$ \\
R+S* & $6.35 \pm 7.65$ & $5.79 \pm 7.58$ & $2.49 \pm 2.18$ & $7.38 \pm 8.51$ & $15.53 \pm 15.62$ & $9.20 \pm 8.83$ \\
R+S+A* & $3.95 \pm 2.07$ & $3.13 \pm 2.12$ & $1.79 \pm 1.07$ & $2.31 \pm 1.75$ & $4.70 \pm 3.60$ & $7.58 \pm 7.51$ \\
S+A$\dagger$ & $9.50 \pm 11.28$ & $7.45 \pm 9.49$ & $3.82 \pm 5.89$ & $7.08 \pm 12.29$ & $12.60 \pm 15.65$ & $17.65 \pm 17.62$ \\ 
\hline
\multicolumn{7}{l}{EpiSurfEmb (multi-view, trained purely on synthetic data)} \\
OL+OR & $ 7.80 \pm 6.74 $ & $ 3.38 \pm 2.96 $ & $ 3.81 \pm 4.20 $ & $ 3.60 \pm 2.38 $ & $ 6.59 \pm 4.50 $ & $ 14.06 \pm 15.39 $ \\
L+OL+OR+R+C & $ 5.19 \pm 3.02 $ & $ 2.41 \pm 1.13 $ & $ 2.46 \pm 1.64 $ & $ 2.42 \pm 3.19 $ & $ 4.48 \pm 4.35 $ & $ 7.33 \pm 9.65 $ \\
\hline
\multicolumn{7}{l}{EpiSurfEmb (multi-view, trained purely on real data)} \\
OL+OR & $2.26 \pm 1.26$ & $1.42 \pm 1.01$ & $1.06 \pm 0.68$ & $1.60 \pm 1.58$ & $3.07 \pm 3.07$ & $4.91 \pm 4.96$ \\
L+OL+OR+R+C & $\mathbf{1.85 \pm 1.10}$ & $\mathbf{1.01 \pm 0.70}$ & $\mathbf{0.89 \pm 0.58}$ & $\mathbf{1.42 \pm 1.44}$ & $\mathbf{2.79 \pm 2.81}$ & \underline{$3.33 \pm 2.68$} \\ 
\hline
\end{tabular}
\label{tab:test-results}
\end{table*}

\begin{figure*}[t]
\centering
\hspace*{0pt}
\hfill
\includegraphics[height=6cm, width=.32\linewidth, keepaspectratio]{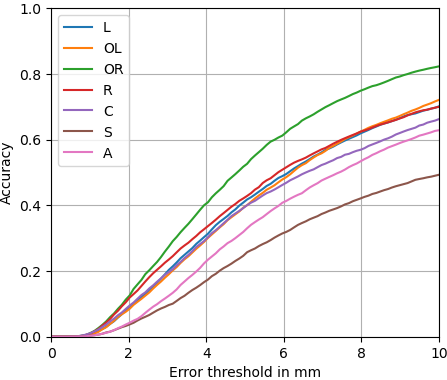}
\hfill
\includegraphics[height=6cm, width=.32\linewidth, keepaspectratio]{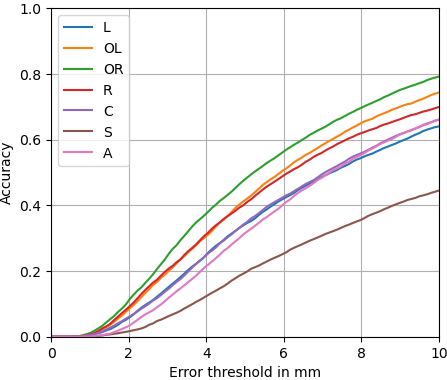}
\hfill
\includegraphics[height=6cm, width=.32\linewidth, keepaspectratio]{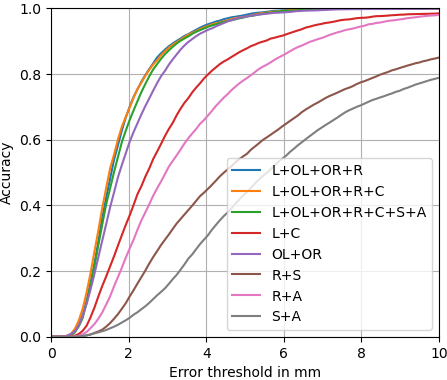}
\hfill
\hspace*{0pt}
\caption{
Accuracy-threshold curves of the ADD(-S) error for ZebraPose (left), SurfEmb (middle), and EpiSurfEmb (right), per camera. 
Results are averaged over both instruments.}
\label{fig:recall-curves}
\end{figure*}

\begin{figure*}[t]
\centering
\hspace*{0pt}
\hfill
\includegraphics[height=6cm, width=.5\linewidth, keepaspectratio]{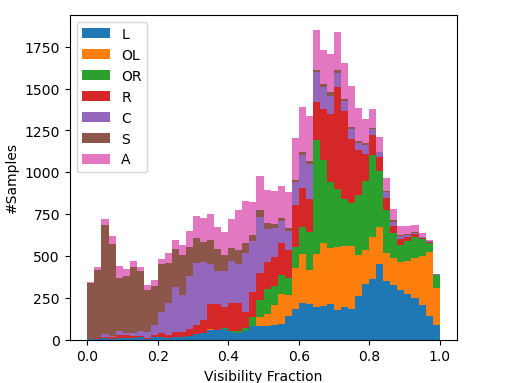}
\hfill
\includegraphics[height=6cm, width=.5\linewidth, keepaspectratio]{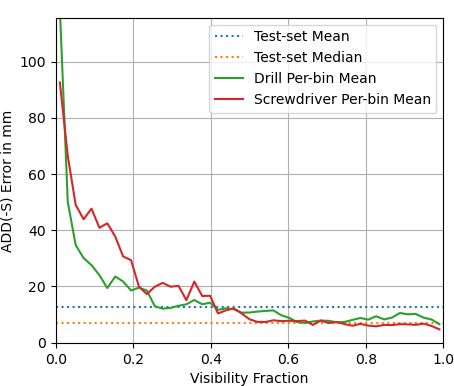}
\hfill
\hspace*{0pt}
\caption{
The left plot shows the stacked, per-camera histograms of the visibility fractions in the wet lab dataset.
The majority of frames captured with the surgeon's \ac{hmd} have a relative visibility of less than $0.2$.
On the right, we compare the relative visibility to the ADD(-S) error of the SurfEmb baseline, prior to depth refinement. 
The shown ADD(-S) error is averaged within 50 bins of equal width.
We observe that the ADD(-S) error increases exponentially with decreasing visibility. }
\label{fig:evaluation-visibility}
\end{figure*}

\begin{figure*}[t]
\centering
\hfill
\begin{minipage}[b]{.245\linewidth}
    \centering
    \includegraphics[height=3cm, width=0.49\linewidth, keepaspectratio]{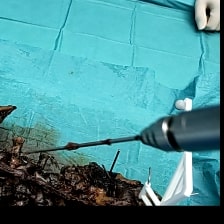}
    \includegraphics[height=3cm, width=0.49\linewidth, keepaspectratio]{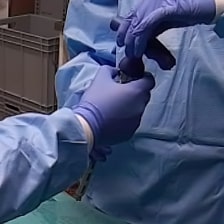} \\
    \includegraphics[height=3cm, width=0.49\linewidth, keepaspectratio]{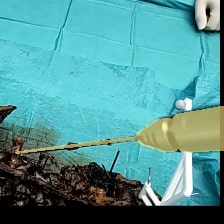}
    \includegraphics[height=3cm, width=0.49\linewidth, keepaspectratio]{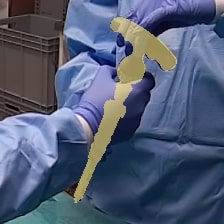} \\
    (a)
\end{minipage}%
\hfill
\begin{minipage}[b]{.36\linewidth}
    \centering
    \adjincludegraphics[height=3cm, width=\linewidth, trim={0 0 {0.4\width} 0}, clip, keepaspectratio]{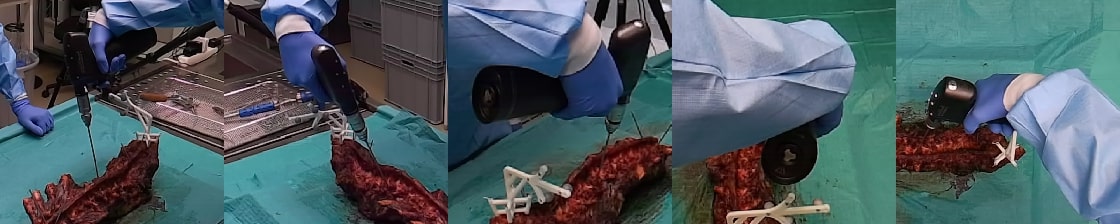} \\
    \adjincludegraphics[height=3cm, width=\linewidth, trim={0 {.34\height} {0.4\width} {.33\height}}, clip, keepaspectratio]{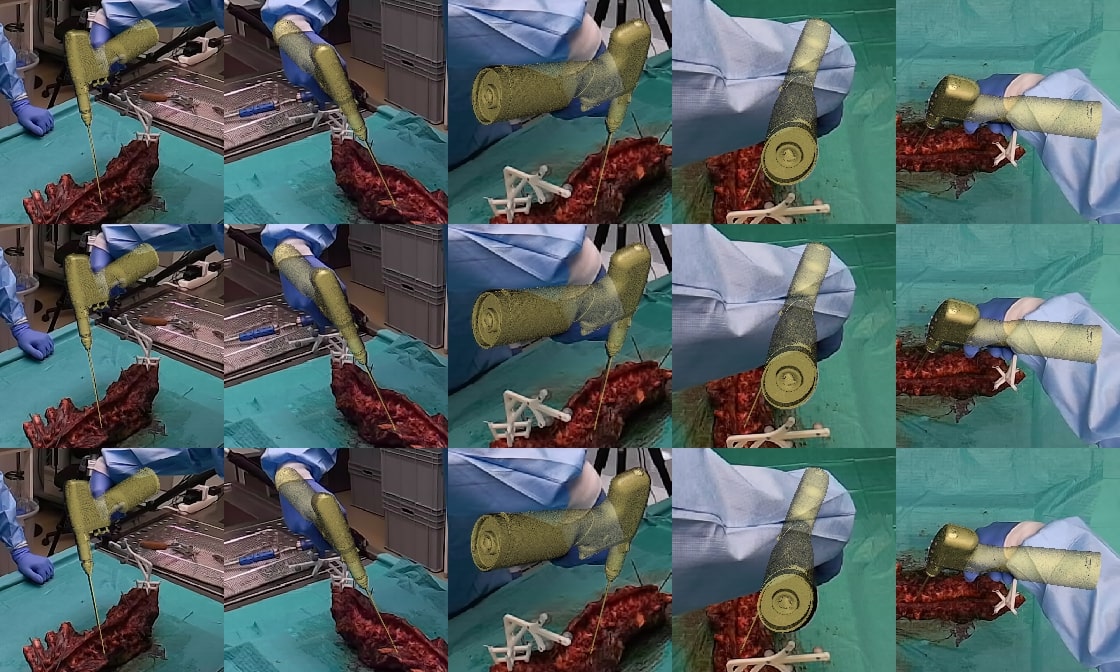} \\
    (b)
\end{minipage}%
\hfill
\begin{minipage}[b]{.36\linewidth}
    \centering
    \adjincludegraphics[height=3cm, width=\linewidth, trim={{.2\width} 0 {.2\width} 0}, clip, keepaspectratio]{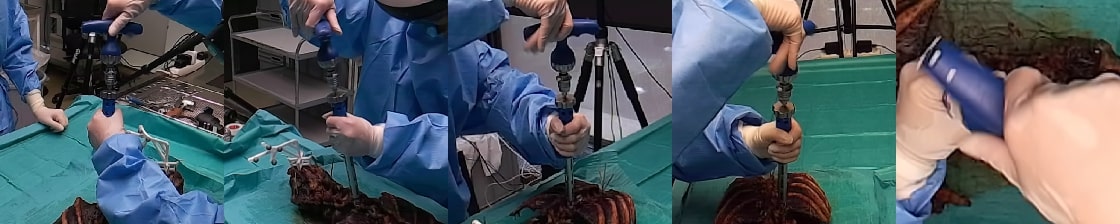} \\
    \adjincludegraphics[height=3cm, width=\linewidth, trim={{.2\width} {.34\height} {.2\width} {.33\height}}, clip, keepaspectratio]{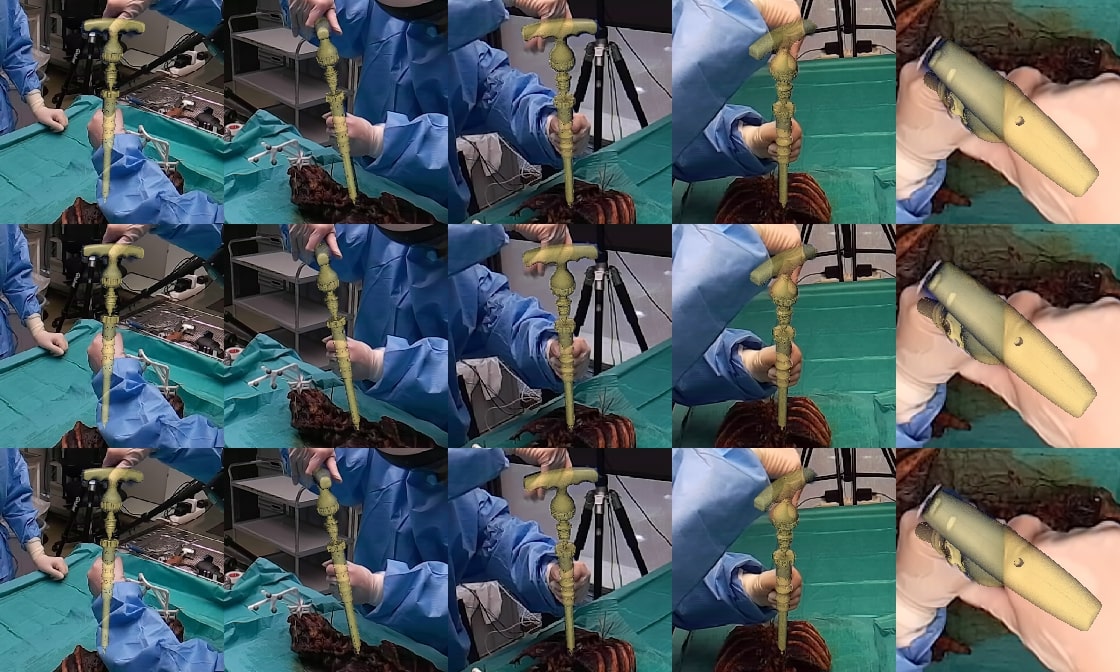} \\
    (c)
\end{minipage}%
\hfill
\caption{Qualitative results for single-view (a) and multi-view (b) \& (c). 
SurfEmb is robust to heavy truncations and occlusions, as shown in (a).
For EpiSurfEmb three of five input views are displayed. 
Input RGB patches are shown in the top row. 
The estimated poses are superimposed in the bottom row. 
}
\label{fig:qualitative-results}
\end{figure*}

\begin{figure*}[t]
\centering
\begin{minipage}[t]{.32\linewidth}
\centering
\includegraphics[height=5cm, width=\linewidth, keepaspectratio]{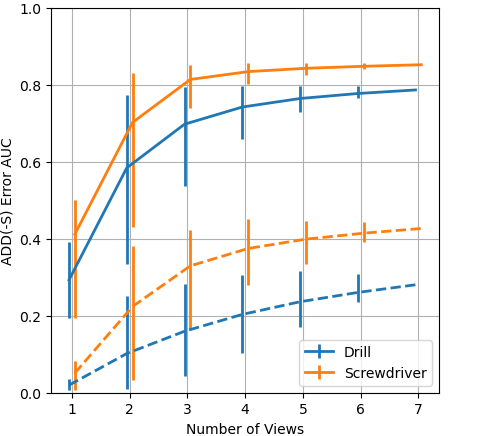}
\captionof{figure}{Influence of the number of cameras on the area under the ADD(-S) curve (AUC) on the interval of \SIrange{0}{10}{\milli\meter} (solid lines) and on the interval of \SIrange{0}{2.5}{\milli\meter} (dashed lines). 
Error bars indicate the best and worst camera configurations.}
\label{fig:ad_vs_nviews}
\end{minipage}%
\hfill
\begin{minipage}[t]{.65\linewidth}
\includegraphics[height=5cm, width=.5\linewidth, keepaspectratio]{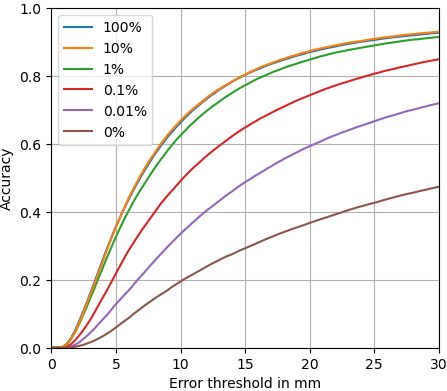}%
\hfill
\includegraphics[height=5cm, width=.5\linewidth, keepaspectratio]{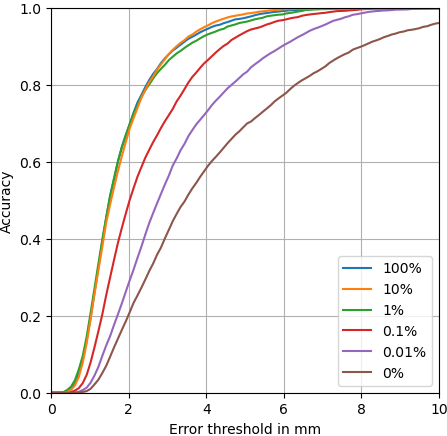}%
\hfill
\captionof{figure}{
Influence of the amount of real training data, evaluated for SurfEmb (left) and EpiSurfEmb with five static cameras (right) on the wet lab dataset. 
All models were trained with a fraction of the real training set and the complete synthetic training set.
We observe no significant performance drop with as little as 1\% of the real training samples. 
The curves for 100\% and 10\% are almost identical.
}
\label{fig:limited_data_graphs}
\end{minipage}%
\end{figure*}

\begin{table*}[t]
\centering
\caption{We report the ADD(-S), position, and orientation errors on the OR-X test set as mean $\pm$ std. 
The orientation error is defined as the geodesic distance between the estimated and ground truth rotation.
Bold and underlined font indicates the lowest and second-lowest average error, respectively.
The camera identifiers as shown in \cref{fig:camera_setup_orx} are opposite left (OL), opposite right (OR), left (L), right (R), and far (F). 
}
\begin{tabular}{lcccccc}
\hline
\multicolumn{1}{c}{\multirow{2}{*}{Model}} & \multicolumn{3}{c}{Bright Subset}                      & \multicolumn{3}{c}{Dark Subset} \\
\multicolumn{1}{c}{} & ADD (mm) $\downarrow$ & $\Delta t$ (mm) $\downarrow$ & $\Delta R$ (deg) $\downarrow$ & ADD (mm) $\downarrow$ & $\Delta t$ (mm) $\downarrow$ & $\Delta R$ (deg) $\downarrow$
 \\ \hline
\multicolumn{7}{l}{EpiSurfEmb (multi-view)} \\
OL+OR & $6.26 \pm 4.96$ & $5.24 \pm 4.61$ & $2.18 \pm 1.48$ & \underline{$7.79 \pm 3.51$} & \underline{$6.17 \pm 3.78$} & \underline{$3.13 \pm 1.73$} \\
L+OL+OR+R+F & $5.53 \pm 5.03$ & $\mathbf{4.77 \pm 4.55}$ & $1.69 \pm 1.33$ & $\mathbf{6.21 \pm 3.56}$ & $\mathbf{5.66 \pm 3.37}$ & $\mathbf{2.25 \pm 1.37}$ \\ \hline
\multicolumn{7}{l}{EpiSurfEmb (multi-view, trained purely on synthetic data)} \\
OL+OR & \underline{$5.46 \pm 5.26$} & \underline{$4.92 \pm 4.69$} & \underline{$1.60 \pm 1.44$} & $18.33 \pm 39.90$ & $14.87 \pm 35.29$ & $5.91 \pm 9.68$ \\
L+OL+OR+R+F & $\mathbf{5.20 \pm 7.33}$ & $5.12 \pm 22.84$ & $\mathbf{1.50 \pm 3.76}$ & $24.36 \pm 55.76$ & $31.38 \pm 86.64$ & $14.55 \pm 38.90$ \\ \hline
\multicolumn{7}{l}{EpiSurfEmb (multi-view, trained purely on real data)} \\
OL+OR & $7.12 \pm 5.05$ & $6.35 \pm 4.80$ & $2.62 \pm 1.66$ & $9.93 \pm 5.53$ & $8.00 \pm 4.33$ & $4.80 \pm 3.86$ \\
L+OL+OR+R+F & $6.15 \pm 4.98$ & $5.51 \pm 4.59$ & $1.99 \pm 1.37$ & $7.98 \pm 3.69$ & $7.31 \pm 3.89$ & $3.15 \pm 1.68$ \\ \hline
\end{tabular}
\label{tab:orx-results}
\end{table*}

\begin{figure}[t]
\centering
\hspace*{0pt}
\hfill
\includegraphics[height=6cm, width=\linewidth, keepaspectratio]{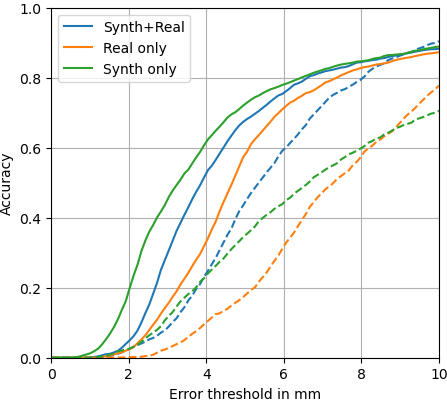}
\hfill
\hspace*{0pt}
\caption{ADD accuracy-threshold curves of EpiSurfEmb with five views on the OR-X test set.
The results on the bright subset are indicated with solid lines; results on the dark subset are shown as dashed lines.}
\label{fig:orx-graphs}
\end{figure}

We select Zebrapose \citep{su2022zebrapose} and Surfemb \citep{haugaard2022surfemb} as our single-view baselines, and EpiSurfEmb \citep{haugaard2023multi} as our multi-view baseline.
This selection is motivated by their state-of-the-art performance on the \ac{bop} \citep{hodan2018bop}. 
Other multi-view pose estimation methods like Cosypose \citep{labbe2020cosypose} and DPODv2 \citep{shugurov2021dpodv2} were discarded as they are either not designed for multi-view single-object pose estimation or do not provide a reference implementation. 
Also, in contrast to end-to-end trained pose estimation models, the selected baselines estimate 2D-3D correspondences as an intermediate representation and recover the 6\dof pose by solving an interpretable geometric optimization problem. 
The interpretability of this intermediate representation can be utilized to compute the uncertainty of the pose estimate in the future \citep{haugaard2023spyropose}, which is highly relevant to avoid presenting inaccurate information to the surgeon.

Zebrapose \citep{su2022zebrapose} iteratively divides the target object's surface into two equal parts in $N$ hierarchical steps. 
The entire surface is thus divided into $2^N$ fragments, whereby each fragment can be identified with a binary code of length $N$.
The bits of the binary code effectively describe 2D-3D correspondences with increasing granularity.
A model $f : \mathbb{R}^{H \times W \times 3} \rightarrow \mathbb{R}^{H \times W \times N+1}$ is trained to estimate the mask and per-pixel binary code of the $H \times W$ sized RGB image.
During training, the loss for each bit is gradually adjusted to shift the focus from coarse to fine correspondences.
The 6\dof pose estimate is computed using progressive-x \citep{barath2019progressive}.

Similar to Zebrapose, SurfEmb \citep{haugaard2022surfemb} estimates 2D-3D correspondences via intermediate descriptors for the 3D surface.
However, the authors propose to learn a surface embedding instead of using hand-crafted descriptors.
A key model $g : \mathbb{R}^3 \rightarrow \mathbb{R}^E$ maps 3D points on the object's surface to keys in a latent space $\mathbb{R}^E$.
A query model $f : \mathbb{R}^{H \times W \times 3} \rightarrow \mathbb{R}^{H \times W \times E+1}$ estimates the object mask as well as a per-pixel query based on the RGB input, where keys and queries live in the same latent space $\mathbb{R}^E$.
Pose hypotheses are sampled from the 2D-3D correspondence distribution via RANSAC-\acs{pnp} with an inlier threshold of $\theta = 2\textnormal{px}$ and scored based on the agreement of object mask and correspondence distributions under the pose hypothesis.
The best pose hypothesis is locally optimized based on the correspondences and optionally refined on the range map obtained from a depth sensor, if available.

EpiSurfEmb \citep{haugaard2023multi} extends SurfEmb to multi-view input. 
Given a set of input images and the relative camera poses, EpiSurfEmb estimates a 3D-3D correspondence distribution based on the per-view 2D-3D correspondence distributions obtained from SurfEmb.
Hereby, 3D points are triangulated from pairs of corresponding 2D points in two randomly selected views, taking into account epipolar constraints.
Pose hypotheses are sampled from the 3D-3D correspondence distribution via RANSAC and Kabsch's algorithm.

\section{Results}\label{sec:evaluation}

Based on the selected baseline models, we evaluated the effect of several parameters on the pose accuracy, namely the number of cameras, their spatial configuration, and the size of the real training dataset. 
All experiments are conducted on cropped image patches based on the ground truth 2D bounding box.
\rev{SurfEmb and EpiSurfEmb operate on image patches of size $224 \times 224$px, while ZebraPose operates on slightly larger patches of size $256 \times 256$px.}
In practice, these image patches can be obtained using a 2D bounding box detector or - in a tracking approach - via the estimated pose on the previous frame \citep{redmon2018yolov3, fang2021you}.
For each baseline, we train a single model to estimate the poses of both instruments.
Given that EpiSurfEmb is built upon SurfEmb, the key and query models are trained only once and then used for both single-view and multi-view assessments.

Throughout this section and concerning the parameters above, we compare three different training strategies, namely using only synthetic data, only real data, or training jointly on both synthetic and real data.
For the joint training on both data types, each model was first trained exclusively on the synthetic dataset until convergence, and then refined on both synthetic and real data of the wet lab dataset. 
All models were trained with a cyclic learning rate between \SIrange{1e-4}{1e-5}{}, which was determined via a range test as proposed by \cite{smith2018disciplined}.
We show in \cref{sec:generalizability} that the models trained jointly on synthetic and real data show the best generalization abilities.
Unless otherwise specified, models were trained using this strategy.

\paragraph{Evaluation Metrics}

Following \cite{hinterstoisser2012model} we evaluate the performance of all models using the ADD(-S) metric, which measures the average distance between corresponding object vertices under the estimated and the ground truth pose.
For symmetric objects like the screwdriver, the metric corresponds to the average distance to the closest vertex under the ground truth pose.
We additionally report the position error of the instrument origin and the orientation error, as depicted in \cref{fig:tool-coordinate-frames}.
Last, we evaluate the pose accuracy relative to the commonly used visibility factor \citep{hodan2018bop}, which is defined as the area of the modal mask relative to the area of the amodal mask.

\subsection{Camera Configurations and Pose Estimation Accuracy}

To find the optimal camera configuration, we exhaustively evaluate the baselines across all possible 127 camera configurations, ranging from 1 to 7 cameras, as depicted in \cref{fig:camera_setup}. 
The results of our single- and multi-view pose estimation baselines are summarized in \cref{tab:test-results}.

EpiSurfEmb trained on purely real data achieves the highest pose accuracy with average ADD(-S) errors of \SI[multi-part-units=single]{1.85 +- 1.10}{\milli\meter} and \SI[multi-part-units=single]{1.42 +- 1.44}{\milli\meter} for drill and screwdriver, respectively, using five static cameras.
The mean position and orientation errors for the drill are \SI{1.01}{\milli\meter} and \SI{0.89}{\degree}, while we observe errors of \SI{2.79}{\milli\meter} and \SI{3.33}{\degree} for the screwdriver.
However, as we discuss in \cref{sec:generalizability}, models trained jointly on synthetic and real data show significantly better generalization capabilities, which is required to achieve sufficient robustness for clinical applications. 
We focus on these models in the following evaluations.

On single-view RGB patches, Zebrapose and SurfEmb achieve a similar pose accuracy of about \SI{12.5}{\milli\meter}  average ADD error for the drill.
In comparison, the pose estimates for the screwdriver are less accurate, with about 2 - 4 times larger position and orientation errors.
Also, SurfEmb significantly outperforms Zebrapose on the screwdriver.
As shown in \cref{fig:recall-curves}, we further observe slight performance differences between the static cameras.
The best single view is provided by the opposite right camera, where SurfEmb achieves the lowest average ADD(-S) error of \SI{6.85}{\milli\meter}.
Pose estimates based on the surgeon’s perspective are significantly less accurate on average, which is due to the narrow \ac{fov} and the proximity to the instruments, resulting in frequent and heavy truncation.
\rev{The lower image resolution of the \acp{hmd} does not pose a significant limitation, as $89\%$ of the extracted image patches are still downscaled to match the input size required by our baseline models.}

To evaluate the influence of occlusions on the pose accuracy, we evaluate the visibility distributions for all cameras and express the mean ADD(-S) error (prior to depth refinement) as a function of the relative visibility. 
The results are shown in \cref{fig:evaluation-visibility}.
The average instrument surface visibility from the surgeon's perspective is only $19\%$, which is significantly lower than the average visibility of $63\%$ for all other cameras.
We find that a relative visibility of less than $60\%$ results in an exponential increase in ADD(-S) error, whereas greater visibility has little influence on the average pose accuracy.
The mean ADD(-S) error on frames with at least $60\%$ visibility is similar for all cameras and between \SI{5.93}{\milli\meter} for the surgeon \ac{hmd} and \SI{8.86}{\milli\meter} for the left camera.
On frames with a medium occlusion level between \SIrange{20}{60}{\%} the surgeon \ac{hmd} is slightly outperformed by the opposite right camera with ADD(-S) errors of \SI{8.88}{\milli\meter} and \SI{8.76}{\milli\meter}, respectively.
Nevertheless, SurfEmb achieves a low visible surface discrepancy even for heavily occluded or truncated instruments, as shown in \cref{fig:qualitative-results}.
We observe that approximately $90\%$ of the position errors are along the camera's depth (Z) axes.

The depth refinement step proposed by \cite{haugaard2022surfemb} does not consistently improve the pose accuracy, but can significantly degrade it.
We found that this degrading performance is mainly caused by a lack of robustness against partial occlusions, as well as limitations of the depth sensor. 
Similarly, we observe that the ICP refinement proposed for Zebrapose severely decreases the average pose accuracy.
These results are in line with our preliminary evaluations of ICP on the captured point clouds, where the average pose error was about \SI{7}{\milli\meter} even when initializing ICP with the ground truth pose.
We include representative failure cases in the appendix.

In the multi-view setting, we observe consistently low average pose errors throughout all combinations of static cameras.
The best configuration consists of all five static cameras and achieves an ADD(-S) error of \SI[multi-part-units=single]{1.75 +- 1.33}{\milli\meter}. 
The mean position and orientation errors of the drill are \SI[multi-part-units=single]{1.06 +- 0.71}{\milli\meter} and \SI[multi-part-units=single]{0.95 +- 0.75}{\degree}.
Similar to the single-view scenario, the mean position and orientation errors of the screwdriver are higher at \SI[multi-part-units=single]{2.95 +- 2.82}{\milli\meter} and \SI[multi-part-units=single]{3.29 +- 2.65}{\degree}.
The worst fully-static configuration consisting of the left and ceiling cameras still achieves an ADD(-S) error of \SI[multi-part-units=single]{2.83 +- 2.52}{\milli\meter}, which is a $60\%$ error reduction compared to the best single-view result.

As expected, the average pose error decreases with an increasing number of cameras, as shown in \cref{fig:ad_vs_nviews}.
Moreover, the best 2-view configuration consisting of the opposite left and opposite right cameras achieves an ADD(-S) error of \SI[multi-part-units=single]{1.99 +- 1.42}{\milli\meter}, which is only \SI{0.24}{\milli\meter} worse than the best configuration with five cameras.
This suggests that adding viewpoints to a pair of unoccluded and complementing viewpoints leads to negligible improvements when using EpiSurfEmb with the proposed multi-camera acquisition system.

Hybrid camera configurations including the surgeon's or assistant's \acp{hmd} perform worse than comparable static camera configurations.
Also, the addition of any \ac{hmd} to the best-performing configuration of five static cameras results in a slight performance decrease.
The reason may be a less accurate ground truth due to errors accumulating through the tracking of the \acp{hmd}.
Nevertheless, both \acp{hmd} can improve the performance of small configurations with only two static cameras.
The best hybrid 2-view configuration consists of the right static camera and the assistant's \ac{hmd}, which achieves a mean ADD(-S) error of \SI[multi-part-units=single]{3.33 +- 2.19}{\milli\meter}.

\subsection{Training Strategy}

Collecting real data with accurate annotations is time-consuming and challenging, thus being able to train models on synthetic data is clearly favorable.
We evaluate the SurfEmb and EpiSurfEmb on the wet lab test set after training with only a fraction of the wet lab training set, as well as after training on purely synthetic data.
As can be seen in \cref{fig:limited_data_graphs}, there is a negligible performance drop when performing the training with 1\% (about \SI{12}{\kilo\relax}) of the total real training samples instead of 100\%, for both single-view and multi-view settings.
Below 1\% of the total real samples, the accuracy decreases significantly.
Training without any real data increases the ADD(-S) error by \SI{3.95}{\milli\meter} on average in a multi-view setting with five static cameras.

\subsection{Generalizability} \label{sec:generalizability}

To estimate the generalizability to different environments we evaluate SurfEmb and EpiSurfEmb without any additional refinement on the OR-X test set and report the results in \cref{tab:orx-results}. 
Further qualitative results are available in the appendix.

On the bright subset, EpiSurfEmb achieves an average ADD error of \SI[multi-part-units=single]{5.53 +- 5.03}{\milli\meter} using all five static cameras, and \SI[multi-part-units=single]{6.26 +- 4.96}{\milli\meter} using only the opposite left and opposite right cameras.
As shown in \cref{fig:orx-graphs}, training solely on synthetic data results in a slightly better performance than training on both synthetic and real data, which in turn outperforms training solely on real data.
These performance differences indicate that training on real data biases the model towards specific characteristics of this dataset \citep{torralba2011unbiased, tommasi2017deeper}, such as lighting conditions, the instrument pose distributions, and instrument-to-camera distances.
Training without synthetic data further decreases the test-time performance, likely due to the lack of uniformly sampled viewpoints.
In contrast, training exclusively on synthetic data results in a similar performance of about \SI{5.20}{\milli\meter} ADD error on the wet lab test set and the OR-X bright test subset.
These results highlight the need for synthetic training data with controllable and diverse data distributions to obtain robust models.

\rev{
On the dark test subset, we observe a significant performance drop of more than \SI{10}{\milli\meter} when training solely on synthetic data. 
In contrast, the ADD and position errors of models trained on real data or a combination of real and synthetic data decrease by only \SIrange{1}{2}{\milli\meter}. 
While a performance decline is expected due to the absence of such dark and high-contrast images in both the synthetic and real training datasets, the models trained on real data generalize much better to these out-of-domain test samples. 
These results suggest that training on a combination of synthetic and real data leads to a more robust model, and should be the preferred strategy when the test environment settings, e.g. lighting conditions, camera poses, and -configurations, are unknown.
}

\section{Discussion}\label{sec:discussion}

Accurate tracking of surgical instruments can improve the safety and efficiency of surgical procedures. 
In this work, we presented a multi-camera acquisition setup consisting of both static cameras and \acp{hmd} and collected a large-scale dataset with rich annotations including instrument-, anatomy- and \ac{hmd} poses, as well as hands and eye gaze information.
We evaluated single- and multi-camera configurations using state-of-the-art pose estimation methods to find an optimal configuration serving surgical needs with respect to accuracy, occlusion, and simplicity.

Our evaluations show that monocular pose estimation methods do not satisfy the high accuracy requirements of clinical applications due to inherent depth ambiguities.
These results are in line with the findings of \cite{doughty2022hmd} and \cite{hein2021towards}, who report ADD errors of \SI{11.71}{\milli\meter} and \SI{16.73}{\milli\meter} for the same surgical drill in a similar setting.
Our preliminary experiments showed that pose refinement on depth information obtained from time-of-flight sensors results in worse pose estimates on average.
This is primarily attributed to a lacking robustness of the methods against partial occlusions, as well as a low accuracy exhibited by depth sensors.
The use of RGB cameras without depth sensors brings a desirable flexibility to camera hardware and enlarges the application fields.
We found that our selected pose estimation method is robust to strong occlusions with a pose accuracy below 10mm up to 40\% occlusion.
Although the surgeon \ac{hmd} provided one of the most beneficial perspectives on un-occluded frames, in our experiments the small \ac{fov} of the \ac{hl} \ac{pv} camera resulted in frequent and heavy truncation of the instruments, which significantly decreased the average pose accuracy.
In contrast, the perspectives of both static cameras opposite of the surgeon are consistently among the most favorable for the pose estimation task.

Using multiple views for pose estimation resulted in a 10-fold improvement in position accuracy compared to the single-view baselines, which is in line with the findings by \cite{haugaard2023multi}.
The results on the wet lab dataset show that position and orientation errors as low as \SI{1.01}{\milli\meter} and \SI{0.89}{\degree} can be achieved when the test-time camera configuration is known and the model is refined on in-domain data.
The camera configuration consisting of the two cameras opposite of the surgeon consistently outperformed all other configurations of two cameras and achieved a similar pose accuracy as the best overall configuration with five cameras.
These findings highlight that highly accurate and marker-less pose estimation is already within reach with two well-placed cameras.

However, our wet lab dataset was recorded in a controlled environment that is arguably less cluttered than a real operation room.
The benefits of using configurations comprising more than two cameras are expected to be more noticeable in cluttered environments, where cameras are frequently occluded.
As such, our results should be interpreted with an understanding that they indicate the minimum number of unoccluded views necessary to achieve a certain pose accuracy, rather than an absolute quantification of the number of cameras.
Similarly, wide-\ac{fov} \acp{hmd} will be able to show their advantages in more cluttered environments where static cameras are more prone to occlusions \citep{saito2021camera}.

The evaluations on the OR-X test set highlight that synthetic data with controllable and diverse image distributions are important to train robust models that can generalize to different camera setups, i.e. when the test-time conditions are unknown.
In our experiments, complementing our synthetic data set with only \SI{12}{\kilo\relax} real samples was sufficient to reach a similar performance as training with $100\times$ more real samples.
However, the worse pose accuracy on the OR-X test set indicates that in-domain data is still necessary to satisfy the high clinical requirements.
Improvements in the generation of synthetic images could further reduce these requirements.
For example, rendering synthetic images based on a known test-time camera configuration for model refinement may result in more accurate pose estimates than sampling from a uniform pose distribution.

Compared to the results on the surgical wet lab dataset, the evaluations on the OR-X test set show an expected decrease in pose accuracy, which can be partially attributed to the larger scale of the ceiling-mounted camera setup.
This performance decrease can be avoided by using cameras with a higher optical zoom.
Although this is not possible with Azure Kinect cameras, we retained them in our experiments to assess the potential of using depth information. 
Higher optical zooms for RGB cameras are widely available and could increase the pixel density in the surgeon's working volume.

\begin{figure}[t]
\centering
\includegraphics[height=4cm, width=\linewidth, keepaspectratio]{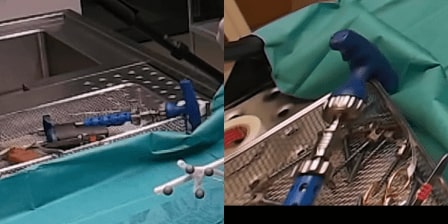}
\\ \vspace*{.5mm}
\includegraphics[height=4cm, width=\linewidth, keepaspectratio]{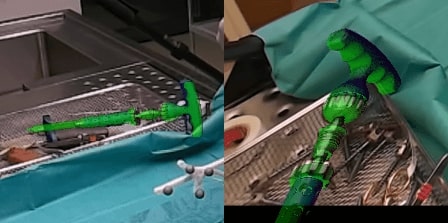}
\caption{In the shown frames the marker-based tracking failed because the instrument is located outside the working volume, while the proposed marker-less multi-camera system remains functional.
The top row shows the two input RGB patches while the bottom row highlights the estimated pose. 
The image patches were manually cropped for these frames, as no ground truth annotations are available. 
The large visual overlap indicates that the pose estimate is accurate even with only two views.}
\label{fig:outside_tracking_volume}
\end{figure}

\subsection{Limitations}

Our work has several limitations. 
First, the use of a single clinical-grade marker-based tracking system for ground truth data capture prevents a comprehensive comparison of the robustness of marker-based and marker-less methods, as no secondary source of ground truth poses is available when tracking is lost.
We found that a significant fraction of frames had to be discarded due to line-of-sight issues.
While we cannot evaluate the performance of our system on these frames due to the lack of ground truth annotations, the pose estimates of the marker-less baselines have a great visual overlap, as shown in \cref{fig:outside_tracking_volume}.
We decided against additional tracking systems despite the increased effective working volume and reduced occlusion issues, to limit interferences between the tracking system and depth sensors in the \ac{ir} spectrum.
Still, multiple tracking systems could be deployed in combination with stereo RGB cameras or sensors that utilize different wavelengths, potentially coupled with matching \ac{ir} filters.

\rev{
Second, the fiducials are visible in a large fraction of training and test images and can potentially bias the model. 
We took measures to avoid overfitting any baseline to the visible marker arrays, by utilizing small fiducials, training on synthetic images without any fiducials, and using a different spatial configuration for the marker array used in the OR-X test set.
\revTwo{Alternatively, visible fiducials can be removed from the training images via inpainting.}
On the OR-X bright subset, the baseline trained purely on synthetic data achieves similar results as the variant trained purely on real data, indicating that the potential bias is limited in our experiments.
However, this risk needs to be taken into account during training.
}

\rev{Third}, the hemispherical fiducials can introduce a triangulation error during the regression of the detected blob centers due to the asymmetric shape of their projections. 
We observed that the screwdriver's pose annotations are less accurate than the drill's pose annotations due to the more challenging placement and detection of the marker array.
Thus, the worse accuracy of the screwdriver's pose estimates might be partially caused by a less accurate ground truth.
The accuracy of the ground truth could be improved using larger spherical or disk-shaped fiducials.
However, large visible markers increase the risk of models overfitting to the marker.
Alternatively, a robot-based capture setup could alleviate the need for markers.

Forth, our dataset does not capture instrument articulations but assumes full rigidity, since marker-based tracking of all articulations is complex and impractical.
As a result, the ground truth 2D-3D correspondences extracted on the affected regions can be incorrect and may prevent models from fully utilizing the local shape and texture information.
In contrast to marker-based approaches, learning-based models can be extended to explicitly model articulations in the future.

Last, occlusion patterns in ex-vivo surgeries are less complex compared to corresponding in-vivo surgeries, due to the limited staff and instruments present.
An analysis of real surgeries and different interventions is necessary to verify our findings on realistic occlusion patterns.

\subsection{Conclusion}

Our study showcased how a dedicated computer vision setup for surgery can enhance current capabilities in surgical navigation and instrument tracking. The comprehensive and systematic evaluation will bring us one step closer to transferring such systems into everyday clinical practice.

A main finding concerning accuracy is that marker-less and millimeter-accurate pose estimation is attainable with as little as two cameras, demonstrating that marker-less tracking is becoming a feasible alternative to existing marker-based systems.
Furthermore, we show that if the test-time camera configuration is known, refinement on real in-domain data can further reduce pose errors to \SI{1.01}{\milli\meter} and \SI{0.89}{\degree} under optimal conditions.
In addition, our results show that synthetic data is important to obtain more robust models, which is particularly relevant in a largely dynamic and varying environment such as surgery.

Nevertheless, there are still surgical applications with accuracy requirements in the sub-millimeter range \citep{rampersaud2001accuracy}.
Further research is needed to improve the pose estimation accuracy and robustness, especially for minimal camera setups and mobile cameras. 
Potential improvements include the explicit modeling of articulations, the temporal integration of 2D-3D correspondences, and a pose uncertainty estimation.
Moreover, capturing the characteristics of a known test-time environment to generate similar synthetic data could further reduce the need for a time-consuming collection of annotated in-domain data.
Last, determining the occlusion patterns during real surgeries is highly relevant to finding optimal camera configurations that satisfy the clinical requirements for accuracy and robustness, while maximizing space efficiency.

We envision our setup as a prototype for robust marker-less optical 6\dof tracking systems in the future trajectory of surgery, and that our dataset accelerates further research in this direction.

\section*{Data availability}
The dataset \revTwo{is} available on our project page \hyperlink{https://jonashein.github.io/mvpsp/}{https://jonashein.github.io/mvpsp/}.

\section*{Declaration of competing interest}
The authors declare the following financial interests/personal relationships which may be considered as potential competing interests:
Mazda Farshad reports a relationship with Incremed AG and X23D AG that includes: equity or stocks.
Philipp Fürnstahl reports a relationship with X23D AG that includes: board membership and equity or stocks.

\section*{CRediT authorship contribution statement}
\textbf{Jonas Hein:} Conceptualization, Methodology, Software, Validation, Formal analysis, Investigation, Data Curation, Writing - Original Draft, Visualization.
\textbf{Nicola Cavalcanti:} Investigation.
\textbf{Daniel Suter:} Investigation.
\textbf{Lukas Zingg:} Investigation.
\textbf{Fabio Carrillo:} Writing - Review \& Editing.
\textbf{Lilian Calvet:} Writing - Review \& Editing.
\textbf{Mazda Farshad:} Resources.
\textbf{Nassir Navab:} Supervision.
\textbf{Marc Pollefeys:} Resources, Supervision, Funding acquisition.
\textbf{Philipp Fürnstahl:} Resources, Writing - Review \& Editing, Supervision, Project administration, Funding acquisition.

\section*{Declaration of generative AI and AI-assisted technologies in the writing process}
\rev{During the preparation of this work the authors used ChatGPT and DeepL Translator to check for linguistic errors and improve readability. The authors reviewed and edited the content as needed and take full responsibility for the content of the publication.}

\section*{Acknowledgements}
This work has been supported by OR-X - a Swiss national research infrastructure for translational surgery - and associated funding by the University Hospital Balgrist, as well as by the InnoSuisse \revTwo{Flagship project PROFICIENCY (No. PFFS-21-19)} \rev{and the Swiss Center for Musculoskeletal Imaging}.
The study was conducted according to the guidelines of the Declaration of Helsinki, and approved by the local ethical committee (KEK Zurich BASEC No. 2017-00874 and 2021-01196).
We like to thank \cite{haugaard2023multi} for providing a reference implementation, as well as Olivia Bossert and Frédéric Giraud for their support during the data capture.

\bibliographystyle{model2-names.bst}\biboptions{authoryear}
\bibliography{main}

\clearpage
\appendix
\section*{Appendix}
\setcounter{figure}{0}
\renewcommand\thefigure{A.\arabic{figure}}

\rev{
\paragraph{Multi-modal Data Capture on HoloLens 2}
We initially faced severe stability issues when trying to capture multiple sensors in parallel on HoloLens 2. 
We conducted a series of tests to find a combination of HoloLens OS version and active sensors that provides a feasible trade-off between stability, frame rates, and modalities captured.
We also tested different recording approaches, namely saving the sensor streams to HoloLens’ internal storage, an external SSD, or streaming them via Wifi or USB-C. 
Based on these tests we selected HoloLens OS version \textit{20348.1450.arm64fre.fe\_release\_svc\_sydney\_rel\_prod.220302-1541} and the streaming-based approach from Microsoft PSI\footnote{\url{https://github.com/microsoft/psi/}}.
Both HoloLenses were connected to the capture server via USB-C cable to eliminate potential bandwidth bottlenecks caused by a wireless transmission.
We capture PV, AHAT and long-throw depth frames in parallel. The effective frame rates were about 29fps for the PV sensor, 11fps for AHAT depth and 5fps for long-throw depth.
During post-processing, we pair each RGB frame with the temporally closest AHAT or long-throw depth frame and constrained the maximum temporal offset between RGB and depth frame to 15ms.
}

\rev{
\paragraph{HoloLens 2 Hand Pose and Eye Gaze Annotations}
Please note that the hand pose and eye gaze information are provided as-is and without any refinement as to not alter the detection rate and quality that would be available in an \ac{ar} application.
We noticed that hand poses are regularly missing, likely due to the hands being outside of the camera's field-of-view or due to mutual occlusions of hand and instrument.
Jointly estimating the hand poses using all cameras could address these issues and likely yield more accurate and complete hand pose annotations.
}

\begin{figure}[t]
\centering
\hspace*{0pt}
\hfill
\includegraphics[height=6cm, width=\linewidth, keepaspectratio]{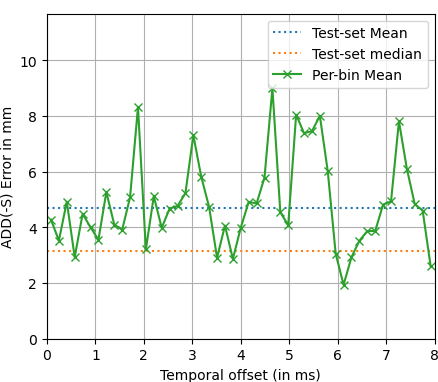}
\hfill
\hspace*{0pt}
\caption{The influence of the temporal offset on the ADD(-S) error is negligible.
We average the ADD(-S) error within 50 bins of equal width. 
Temporal offsets below \SI{50}{\micro\second} are not included.}
\label{fig:temporal_offset_vs_pose_error}
\end{figure}

\paragraph{\ac{hmd} Temporal Synchronization}
To exclude any synchronization issues as the reason for the lower pose estimation accuracy of the \acp{hmd}, we compared the ADD(-S) errors of EpiSurfEmb on all 2-view camera configurations to the corresponding temporal offset between the exposure windows of the two input RGB images.
As displayed in \cref{fig:temporal_offset_vs_pose_error}, we did not find any significant correlation between the two variables.

\begin{figure}[ht]
\centering
\hspace*{0pt}
\hfill
\includegraphics[height=5cm, width=.49\linewidth, keepaspectratio]{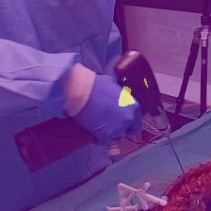}
\includegraphics[height=5cm, width=.49\linewidth, keepaspectratio]{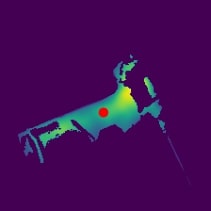}
\hfill
\hspace*{0pt}
\caption{Exemplary frame of SurfEmb's depth refinement step largely sampling from occluded pixels, in this case on the surgeon's hand.
The left image shows the input RGB patch with the selected pixels highlighted. 
The right image shows the predicted query image, where brighter colors indicate a larger norm. The red dot indicates the ray along which the pose is refined.
}
\label{fig:depth_refinement_occlusion}
\end{figure}

\begin{figure}[ht]
\centering
\hspace*{0pt}
\hfill
\includegraphics[height=5cm, width=.49\linewidth, keepaspectratio]{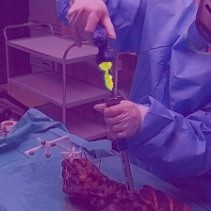}
\includegraphics[height=5cm, width=.49\linewidth, keepaspectratio]{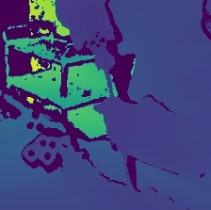}
\hfill
\hspace*{0pt}
\caption{
The sampled pixels for the depth refinement can be placed suboptimally.
The left image shows an exemplary RGB patch, where the selected pixels for the depth refinement are focused around a thin and metallic part of the screwdriver.
The right image shows the captured depth image, which does not fully capture the instrument's shape. 
Note that the center of the instrument appears thinner than it is, while some pixels erroneously contain the depth of the background.
}
\label{fig:depth_sensor_error}
\end{figure}

\begin{figure*}[th]
\centering
\hspace*{0pt}
\hfill
\includegraphics[height=6cm, width=0.49\linewidth, keepaspectratio]{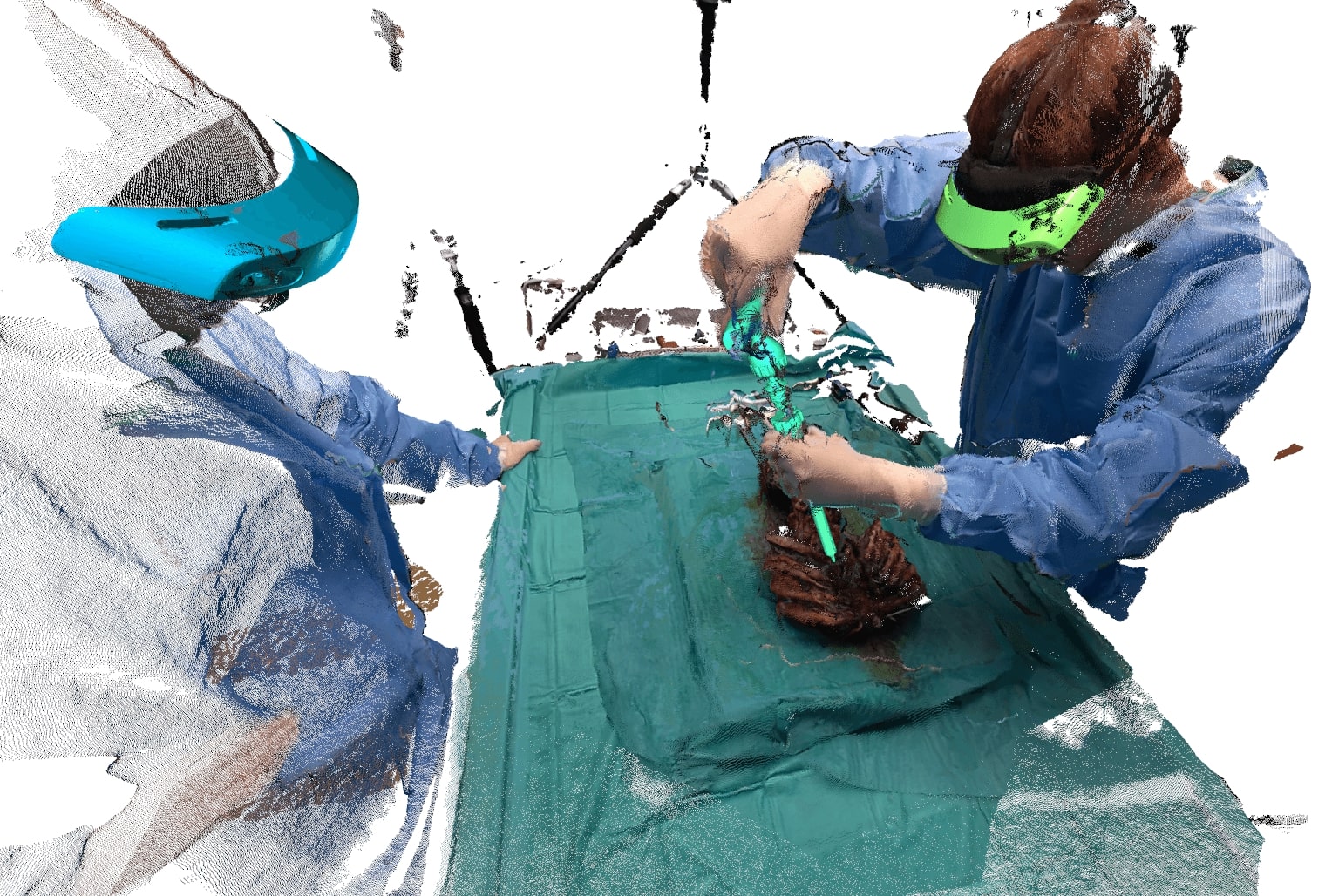}
\hfill
\includegraphics[height=6cm, width=0.49\linewidth, keepaspectratio]{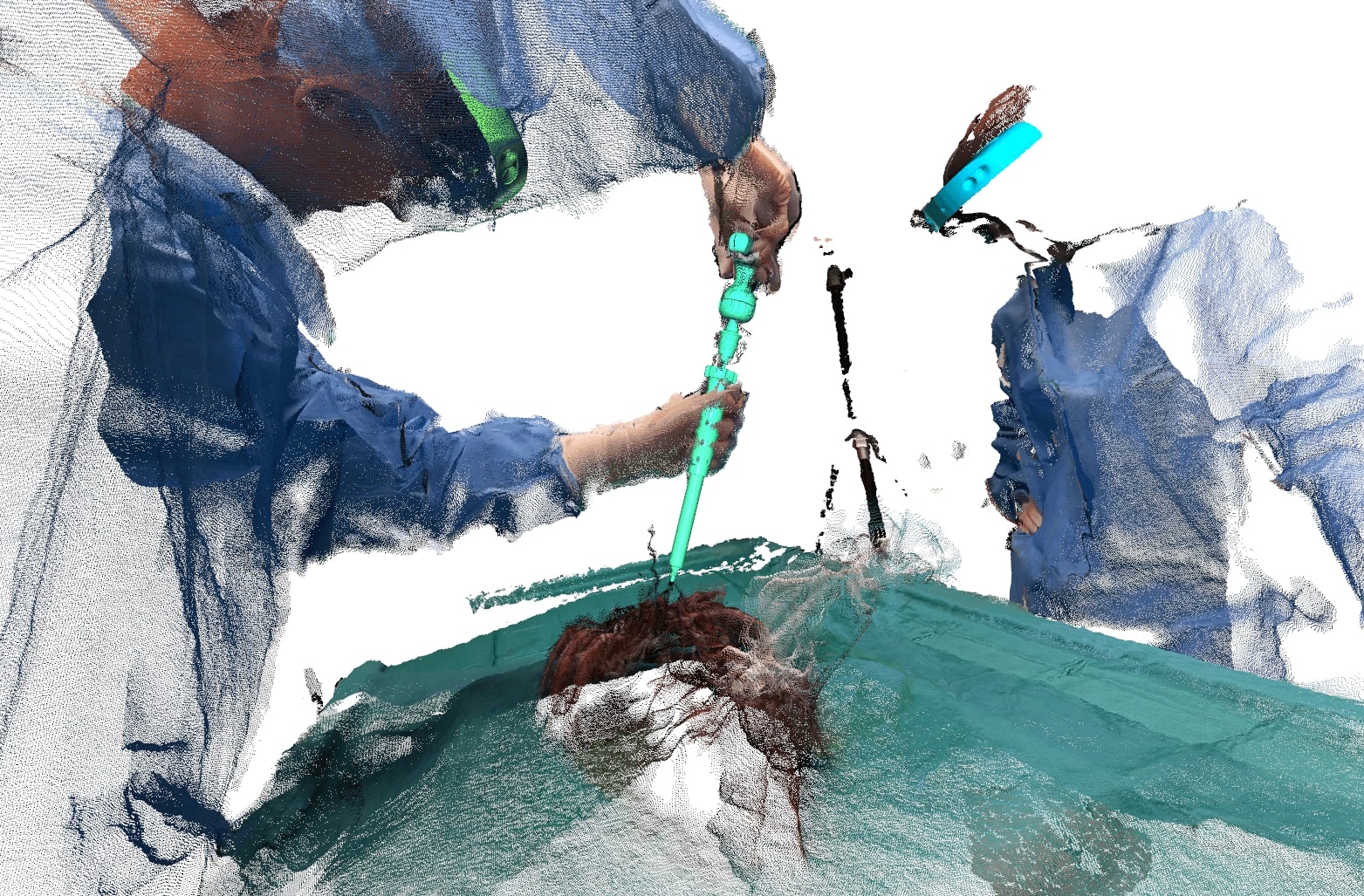}
\hfill
\hspace*{0pt}
\caption{Visualization of the colored point clouds from all cameras in the wet lab dataset. We overlay and highlight the 3D models of both tracked \acp{hmd} and the instrument. 
}
\label{fig:point_clouds}
\end{figure*}

\begin{figure*}[th]
\centering
\includegraphics[width=\linewidth, keepaspectratio]{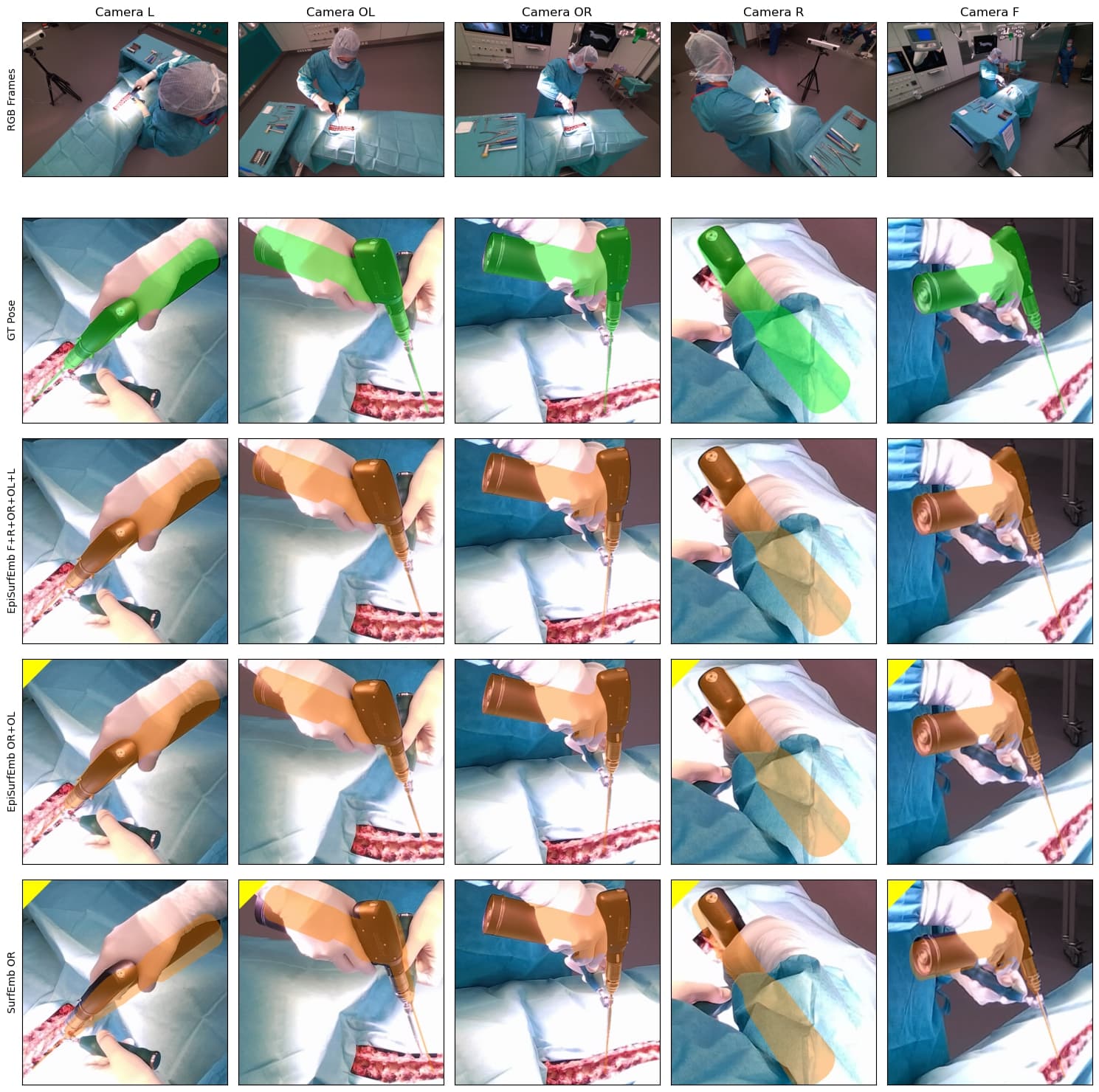}
\caption{Qualitative Comparison of the best single-view, two-view, and five-view baselines on the OR-X bright subset. 
We superimpose the ground truth pose in green in the second row and pose estimates in orange in the following rows. 
Yellow triangles in the top-left image corners indicate that the frame was not part of the input to the pose estimation method.
Note that the single-view pose estimate has a great visual overlap on the input image, but a significant error when viewed from other perspectives due the the depth ambiguity.}
\label{fig:orx_bright_qualitative_comparison}
\end{figure*}

\begin{figure*}[th]
\centering
\includegraphics[width=\linewidth, keepaspectratio]{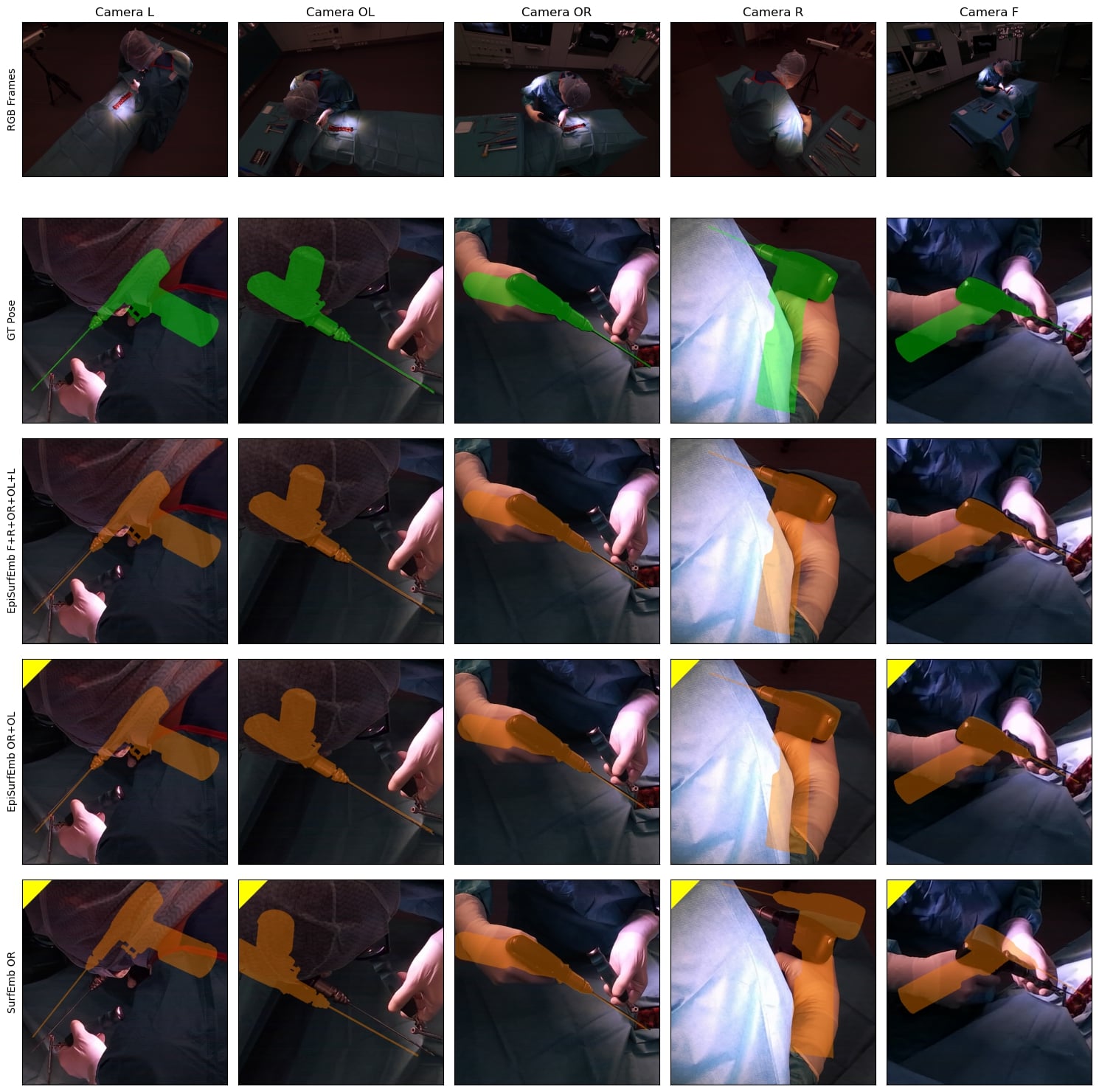}
\caption{Qualitative Comparison of the best single-view, two-view, and five-view baselines on the OR-X dark subset. We superimpose the ground truth pose in green in the second row and pose estimates in orange in the following rows. Yellow triangles in the top-left image corners indicate that the frame was not part of the input to the pose estimation method.
Note that the single-view pose estimate has a great visual overlap on the input image, but a significant error when viewed from other perspectives due the the depth ambiguity.
}
\label{fig:orx_dark_qualitative_comparison}
\end{figure*}

\paragraph{SurfEmb Depth Refinement}
SurfEmb's depth refinement step does not consistently improve the pose accuracy on our dataset.
We found that this refinement can often improve the pose accuracy significantly, but it lacks robustness to partial occlusions.
The method assumes that pixels with a high query norm (i.e. $>80\%$ of the maximum query norm in the image) are not occluded, as the model is most certain about them.
This assumption often does not hold, leading to the majority of pixels being sampled from the surgeon's hand.
In addition, the selected pixels are often focused around a single point, instead of being distributed on the instrument surface.
As a result, the method is less robust to partial occlusions.
A representative example is shown in \cref{fig:depth_refinement_occlusion}.

In some cases, the depth sensor fails to perceive thin or metallic surfaces and erroneously measures the depth of the background.
An exemplary frame is shown in \cref{fig:depth_sensor_error}.
Moreover, we observe that the depth measurements can be inaccurate depending on the normal of the reflecting surface and lighting conditions.

\end{document}